\def\year{2021}\relax
\documentclass[letterpaper]{article} 
\usepackage{aaai21}  
\usepackage{helvet} 
\usepackage{courier}  
\usepackage[hyphens]{url}  
\usepackage{graphicx} 
\usepackage{natbib}  
\usepackage{caption} 
\usepackage[utf8]{inputenc} 
\usepackage[T1]{fontenc}    
\usepackage{url}            
\usepackage{booktabs}       
\usepackage{amsfonts}       
\usepackage{nicefrac}       
\usepackage{microtype} 
\usepackage{mathrsfs}
\usepackage{amsmath}
\usepackage{multirow, hhline}
\usepackage{graphicx}
\usepackage{tikz}
\usepackage{pbox}
\usepackage{pifont}
\usepackage{lmodern}
\usepackage{textcomp}
\usepackage{amssymb}
\usepackage{rotating}
\usepackage{color}
\usepackage{colortbl} 
\usepackage{wrapfig}
\usepackage{wrapfig,lipsum,booktabs}
\usepackage{lipsum}
\usepackage{enumitem}
\usepackage{etoolbox}
\providetoggle{long}
\settoggle{long}{false}
\usepackage{adjustbox}
\usepackage{subcaption}
\usepackage{amssymb}
\usepackage{pifont}
\usepackage{graphicx}  
\usepackage{tabularx}
\usepackage{times}  
\usepackage{amssymb}
\usepackage{amsthm}
\theoremstyle{plain}
\newtheorem{proposition}{Proposition}
\newtheorem{corollary}{Corollary}
\theoremstyle{definition}
\newtheorem{definition}{Definition}
\usepackage{array}
\usepackage{mathrsfs}
\usepackage{subcaption}
\usepackage{tikz}
\usetikzlibrary{positioning}
\definecolor{processblue}{cmyk}{0.96,0,0,0}
\usetikzlibrary{babel}
\usepackage{amssymb}
\usepackage{stackengine}
\usepackage{scalerel}
\usepackage{xcolor}
\usepackage{graphicx}
\newcommand\openbigstar[1][0.7]{%
  \scalerel*{%
    \stackinset{c}{-.125pt}{c}{}{\scalebox{#1}{\color{white}{$\bigstar$}}}{%
      $\bigstar$}%
  }{\bigstar}
}

\newcommand{\comment}[1]{}

\title{5$^\bigstar$ Knowledge Graph Embeddings with Projective Transformations}
\author{
    Mojtaba Nayyeri\textsuperscript{\rm 1,}\textsuperscript{\rm 2},  
    Sahar Vahdati\textsuperscript{\rm 2}, 
    Can Aykul\textsuperscript{\rm 1},
    Jens Lehmann\textsuperscript{\rm 1,}\textsuperscript{\rm 3}
    \\
}
\affiliations{

    \textsuperscript{\rm 1}Smart Data Analytics Group, University of Bonn, Germany \\
    \textsuperscript{\rm 2}Nature-Inspired Machine Intelligence-InfAI, Dresden, Germany \\
     \textsuperscript{\rm 3}Fraunhofer IAIS, Dresden, Germany\\

    \{nayyei,aykul,jens.lehmann\}@cs.uni-bonn.de,
    vahdati@infai.org,
    jens.lehmann@iais.fraunhofer.de

}

\begin{document}
\maketitle

\begin{abstract}
Performing link prediction using knowledge graph embedding models has become a popular approach for knowledge graph completion. 
Such models employ a transformation function that maps nodes via edges into a vector space in order to measure the likelihood of the links.
While mapping the individual nodes, the structure of subgraphs is also transformed.  
Most of the embedding models designed in Euclidean geometry usually support a \emph{single} transformation type -- often translation or rotation, which is suitable for learning on graphs with small differences in neighboring subgraphs. 
However, multi-relational knowledge graphs often include multiple subgraph structures in a neighborhood (e.g.~combinations of path and loop structures), which current embedding models do not capture well.
To tackle this problem, we propose a novel KGE model ($5^\bigstar$E) in projective geometry, which supports \emph{multiple} simultaneous transformations -- specifically inversion, reflection, translation, rotation, and homothety.
The model has several favorable theoretical properties and subsumes the existing approaches.
It outperforms them on most widely used link prediction benchmarks.
\end{abstract}

\section{Introduction}
Knowledge graphs (KGs) with their graph-based knowledge representation in the form of (head,relation,tail) triples, have become a leading technology of recent years in
AI-based tasks including question answering, data integration, and recommender systems
\cite{ji2020survey}.
However, KGs are incomplete and the performance of any algorithm consuming them is affected by this problem.
Knowledge graph embeddings (KGEs) are a prominent approach used for KG completion by predicting missing links.
Every KGE model uses a transformation function to map entities (nodes) of the graph through relations in a vector space to score the plausibility of triples via a score function.
The performance of KGE models heavily relies on the design of their score function that in turn defines the type of transformation they support.  
Such transformations distinguish the extent to which a model is able to learn complex motifs and patterns formed by combinations of the nodes and edges.

\begin{figure}[ht!]
	\centering 
		\includegraphics[trim=0cm 11cm 0cm 0cm,width=9.4cm,height=5.5cm]{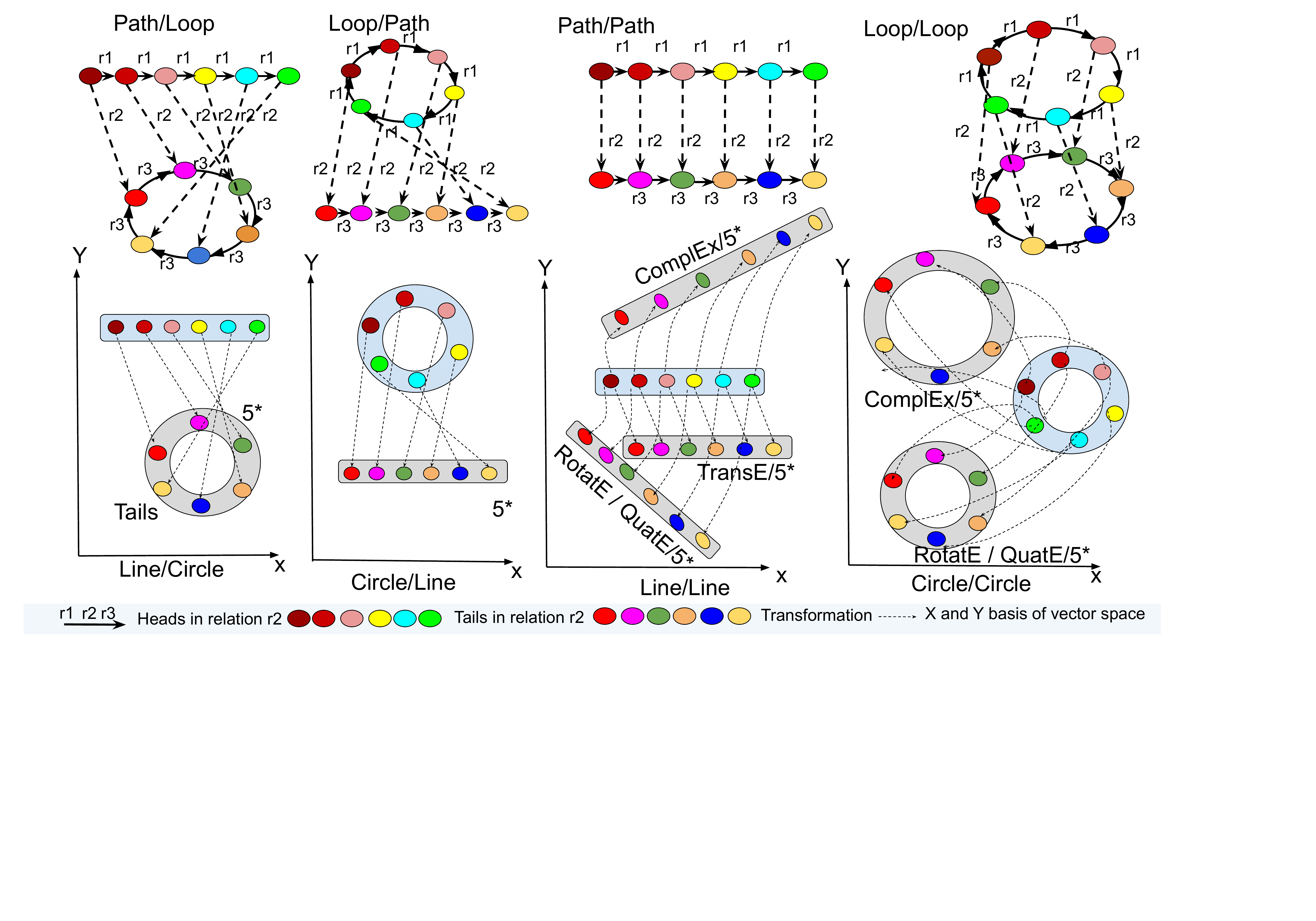}
	\caption{Graph and vector representation of path/loop.}
	\label{fig:all}%
\end{figure}

Most of the existing KGEs have been designed in Euclidean geometry and usually support a single transformation type -- often translation or rotation. 
This limits their ability in embedding KGs with complexities in subgraphs, especially when multiple structures exist in a neighborhood.
An example of this situation is the presence of a path structure for a group of nodes close to a loop structure of another group in a KG (illustrated in Figure~\ref{fig:all}).
The upper part of the figure shows examples of four different subgraphs containing combinations of path structures (a group of nodes connected via a relation) and loop structures (a group of nodes forming a loop via a relation). 
The lower part of the figure shows an exemplary visualisation of the embeddings of the entities depicted in the upper part of the figure. 
Let us focus on the left most example in the figure, i.e.~the path to loop subgraph.
In this example subgraph, a relation $r_1$ (e.g.~\textit{hypernym}) forms a path structure, a relation $r_3$ (e.g.~\textit{similar\_to}) forms a loop structure and nodes in both structures are connected via a relation $r_2$. 
A loop in the graph presentation can be preserved as a circle in a vector space, and a path as a line. 
Existing KGE models, such as TransE, RotatE, ComplEx and QuatE partially capture those structures in the embedding space.
%
The lower part of the figure shows the possible embeddings of the given subgraphs preserved by the existing models. 
Let the nodes in the path be $h_1,\ldots, h_6$, the nodes in the loop be $t_1, \ldots, t_6$ and $(h_i,r_2,t_i), i=1, \ldots, 6,$ be the triples connecting those structures. 
When we consider the TransE model specifically, the embeddings of the tails $t_1, \ldots, t_6$ cannot be transferred to the shape of a circle in the embedding space.  
This is because the following equations need to (approximately) hold
according to the TransE score functions:
$
    \mathbf{t}_1 + \mathbf{r}_3 \approx \mathbf{t}_2,\,
    \mathbf{t}_2 + \mathbf{r}_3 \approx  \mathbf{t}_3,\,
    \dots, \,
    \mathbf{t}_6 + \mathbf{r}_3 \approx  \mathbf{t}_1 $.
%
%
%
However, this results in $\mathbf{r}_3 = 0$. 
Therefore, $\mathbf{t}_1= \mathbf{t}_2= \ldots = \mathbf{t}_6$ i.e.~all entities are embedded into the same point in the embedding space rather than a circle (with positive radius). 
Similar derivations 
apply to more recent and complex models. 
Those limitations are due to the limited set of transformations supported by those models, which do not go beyond translation, rotation and homothethy operations. 
Therefore, they cannot map line structures to circle structures and vice versa.  
This type of limitation stems from the underlying geometry.
While major existing models cover at most two transformation types, we propose a model based on projective geometry that provides a uniform way for \emph{simultaneously} representing five \emph{transformation types} namely translation, rotation, homothety, inversion, and reflection.
The combination of such transformation types results in various \emph{transformation functions} (parabolic, circular, elliptic, hyperbolic, and loxodromic) subsumed by projective transformations. 
As a consequence, the embeddings can preserve the structure in the previous example and many other scenarios in which different structures exist in a neighbourhood of the KG.

Our core contribution is a new five-star embedding model, i.e.~a model that simultaneously employs five transformation types and consequently can preserve various-shaped structures in the embedding space. 
The model subsumes several existing state-of-the-art KGE models 
, i.e.~their score functions can be expressed as special cases of $5^\bigstar$E. 
Overall, our model, dubbed $5^\bigstar$E, is 
(a) capable of preserving a wider range of structures than existing models (including the path and loop combinations in Figure~\ref{fig:all}),
(b) fully expressive (as defined in~\cite{wang2018multi}), 
(c) subsumes the KGE models DistMult, RotatE, pRotatE, TransE, and ComplEx 
(d) allows to learn composition, inverse, reflexive and symmetric relation patterns. 
Our evaluation shows that $5^\bigstar$E outperforms existing models on standard benchmarks.    
\section{Preliminaries and Background}

\subsection{Knowledge Graph Embeddings} 
A KG is a multi-relational directed graph $ \mathcal{KG} =(\mathcal{E}, \mathcal{R}, \mathcal{T})$ where $\mathcal{E}, \mathcal{R}$ are the set of nodes (entities) and edges (relations between entities) respectively. 
The set $\mathcal{T} = \{(h,r,t)\} \subseteq \mathcal{E} \times \mathcal{R} \times \mathcal{E}$ contains all triples as \emph{(head, relation, tail)}, e.g.~(\emph{smartPhone, hypernym, iPhone)}. 
In order to apply learning methods on KGs, a KGE learns vector representations of entities ($\mathcal{E}$) and relations ($\mathcal{R}$). 
A vector representation denoted by ($\mathbf{h, r, t} $) is learned by the model per triple $(h,r,t)$, where $\mathbf{h,t} \in \mathbb{V}^{d_e}$, $\mathbf{r} \in \mathbb{V}^{d_r}$ ($\mathbb{V}^{d}$ is a $d$-dimensional vector space). 
TransE \cite{bordes2013transe} considers $\mathbb{V}=\mathbb{R}$ while ComplEx \cite{complex2016trouillon} and RotatE use $\mathbb{V} = \mathbb{C}$ (complex space) and QuatE \cite{zhang2019quaternion} considers $\mathbb{V}=\mathbb{H}$ (quaternion space). 
In this paper, we choose a projective space to embed the graph i.e.~$\mathbb{V} = \mathbb{CP}^1$ (a complex projective line which is introduced later). 
Most KGE models are defined via a relation-specific transformation function $g_r: \mathbb{V}^{d_e}\rightarrow \mathbb{V}^{d_e}$ 
which maps head entities to tail entities, i.e.~$g_r(\mathbf{h}) = \mathbf{t}$. 
On top of such a transformation function, the score function $f: \mathbb{V}^{d_e} \times \mathbb{V}^{d_r} \times \mathbb{V}^{d_e} \rightarrow \mathbb{R}$ is defined to measure the plausibility for triples: $f(\mathbf{h}, \mathbf{r}, \mathbf{t}) = p(g_r(\mathbf{h}), \mathbf{t})$.
Generally, the formulation of any score function can be either $p(g_r(\mathbf{h}), \mathbf{t}) = -\|g_r(\mathbf{h})- \mathbf{t}\|$ or $p(g_r(\mathbf{h}), \mathbf{t}) = \langle g_r(\mathbf{h}), \mathbf{t} \rangle$.


\subsection{Projective Geometry}
\label{sec:projGe}
Projective geometry uses \emph{homogeneous coordinates} which represent $N$-dimensional coordinates with $N+1$ numbers (i.e.~use one additional parameter). 
For example, a point in 2D Cartesian coordinates, $[X,Y]$ becomes $[x, y, k]$ in homogeneous coordinates where $X = x/k, Y = y/k$ (in 1-dimensional real numbers, $[X]$ becomes $[x, y]$ where $X=x/y$). 
The key elements of projective geometry are as follows:

A \textbf{Projective Line} is a space in which a projective geometry is defined. 
A projective geometry requires a point at infinity to satisfy the axiom of ``two parallel lines intersect in infinity''. 
Therefore, an extended line $\mathbb{P}^1(\mathbb{K})$ ($\mathbb{K}$ is a real line) is realized with $\mathbb{K}$ and a point at infinity (which topologically is a circle).
More concretely, the projective line is a set $\{[x,1] \in \mathbb{P}^1(\mathbb{K}) | x \in \mathbb{K}\}$ with an additional member $[1:0]$ denoting the point at infinity. 
When $\mathbb{K} = \mathbb{C}$, the projective line is complex (complex projective line denoted by $\mathbb{CP}^1$). 

The \textbf{Riemann Sphere} (illustrated in Figure~\ref{fig:tranfun}) is an extended complex plane with a point at infinity.
Precisely, it is built on a plane of complex numbers wrapped around a Sphere where poles denote $0$ and $\infty$. 
In projective geometry, a complex projective line is a Riemann Sphere which used as a tool for projective transformations.

A \textbf{Projective Transformation} is the mapping of the Riemann Sphere to itself. 
Let $[x:y]$ be the homogeneous coordinates of a point in $\mathbb{CP}^1$. 
A projective transformation in $\mathbb{CP}^1$ is expressed by a matrix multiplication \cite{richter2011perspectives,salomon2007transformations}  as $\tau: \mathbb{CP}^1 \rightarrow \mathbb{CP}^1$:
\begin{equation}
\label{eq:PT}
 \tau([x,y]) = \Im \begin{bmatrix}    
x\\ 
y\\
\end{bmatrix},
\,\, \Im = \begin{bmatrix}
a&b\\ 
c&d\\
\end{bmatrix},
\end{equation}
where the matrix $\Im$ must be invertible ($det(\Im) \neq 0$). 
By identifying $\mathbb{CP}^1$ with $\hat{\mathbb{C}} = \mathbb{C} \cup \{\infty\},$ a projective transformation is represented by a fractional expression through a sequence of homogenization, transformation, and dehomogenization as 
\begin{equation}
\label{eq:mob}
    x \rightarrow \begin{bmatrix}
    x\\ 
    1\\
\end{bmatrix} \rightarrow
\begin{bmatrix}
a&b\\ 
c&d\\
\end{bmatrix} \, \begin{bmatrix}
x\\ 
1\\
\end{bmatrix} \rightarrow
\begin{bmatrix}
a x + b\\ 
c x + d\\
\end{bmatrix}\rightarrow
\frac{a x + b}{c x + d},
\end{equation}
where the mapping $\vartheta: \hat{\mathbb{C}} \rightarrow \hat{\mathbb{C}}$ is defined as
\begin{equation}
\label{eq:Mobius}
    \vartheta(x) = \frac{a x + b}{c x + d}, \, a d - b c \neq 0.
\end{equation}
The resulted mapping in Equation~\ref{eq:Mobius} describes all  
\emph{M\"obius} transformations. 

The \textbf{M\"obius Group} is the set of all M\"obius transformations which is a projective linear group $PGL(2, \mathbb{C})$, i.e.~the group of all $2\times2$ invertible matrices with the operation of matrix multiplication on a projective space. 
This group is the automorphism group $Aut(\hat{\mathbb{C})}$ of the Riemann Sphere $\hat{\mathbb{C}}$ or equivalently $\mathbb{CP}^1.$ 


\begin{figure*}[t!]
	\centering 
	\begin{subfigure}[b]{0.19\textwidth}
		\includegraphics[width=\linewidth]{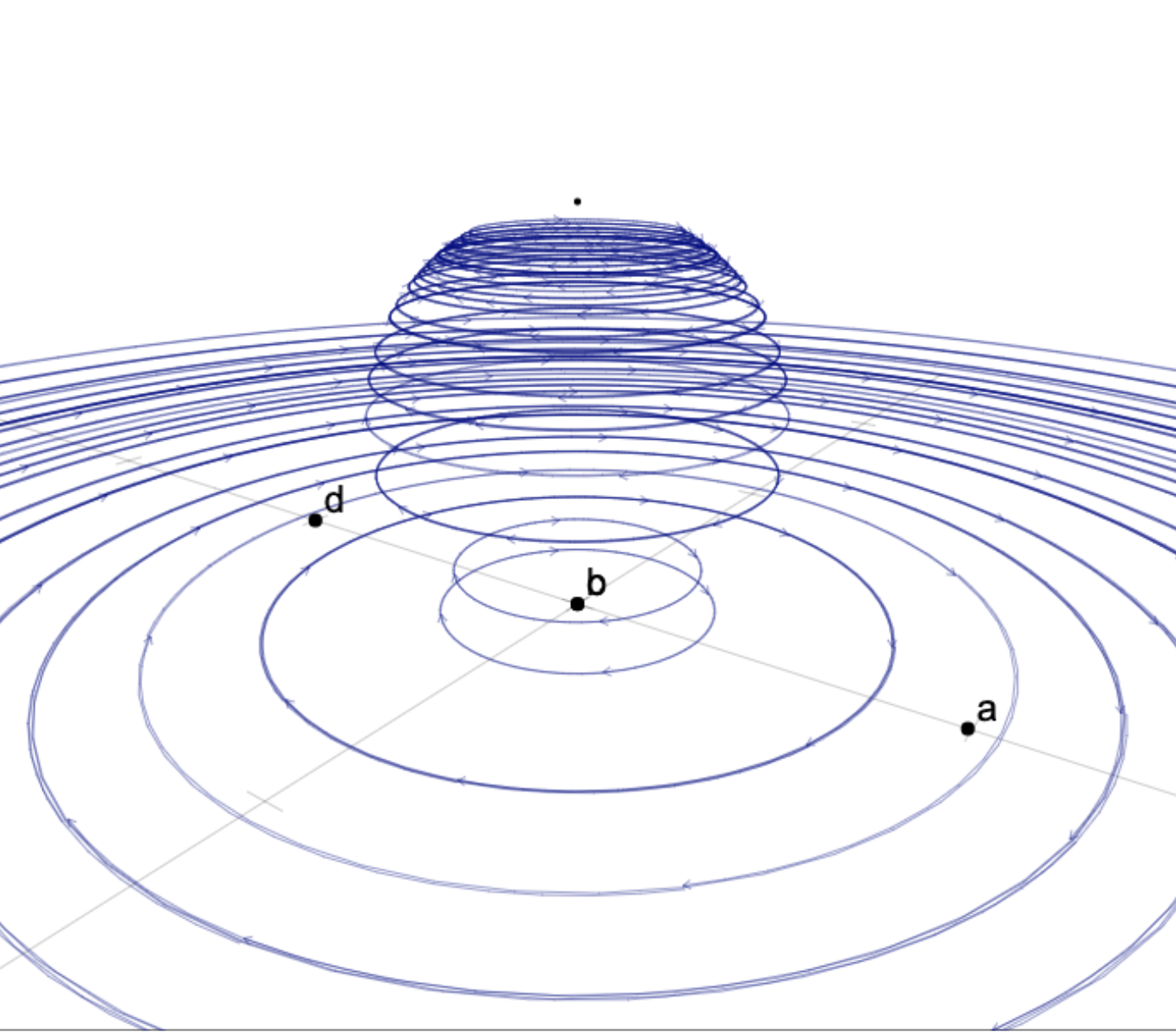}
		\label{fig:fb:noInj}
	\end{subfigure}%
	\begin{subfigure}[b]{0.19\textwidth}
		\includegraphics[width=\linewidth]{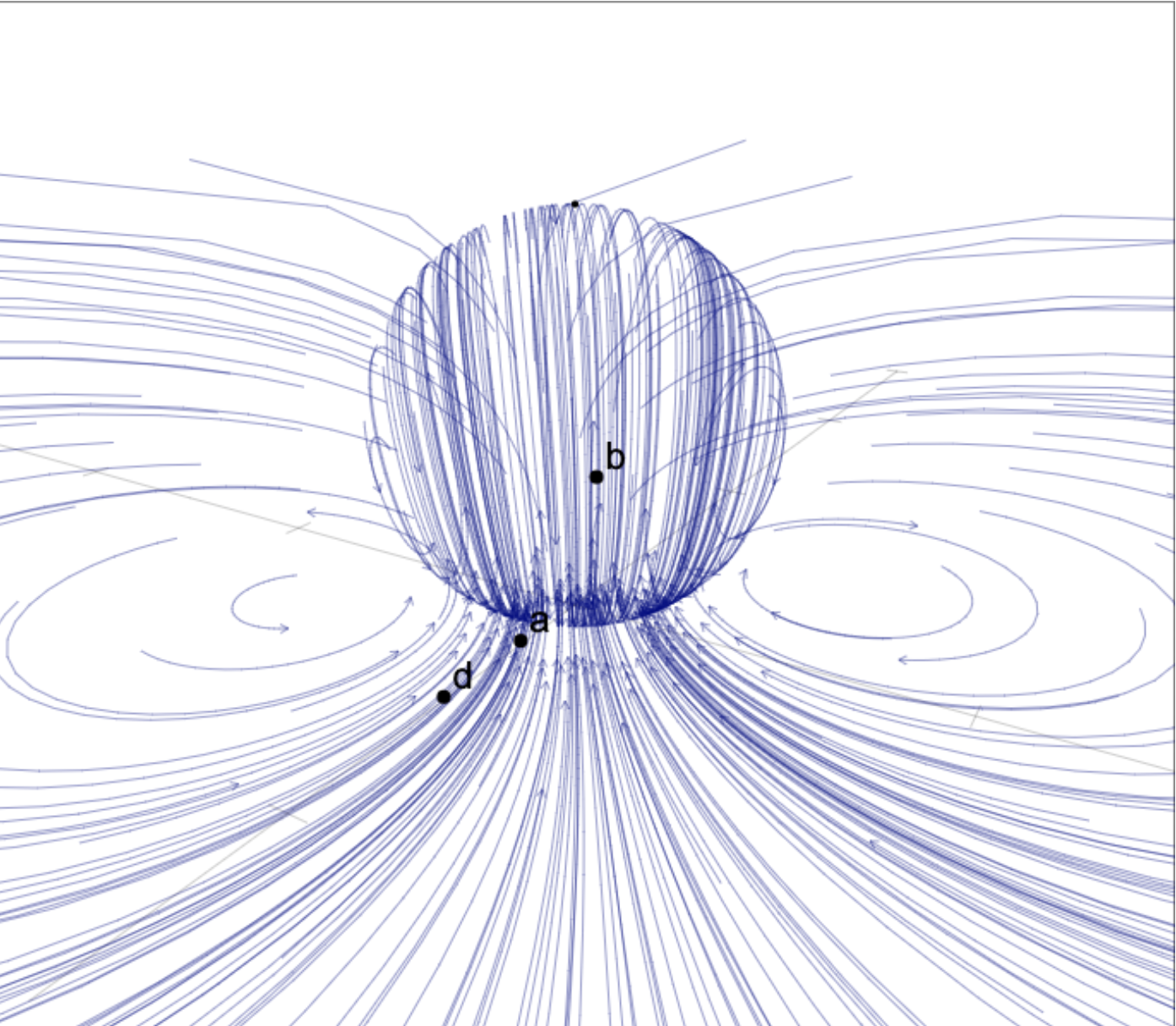}
		\label{fig:4c:eq}
	\end{subfigure}%
	\begin{subfigure}[b]{0.19\textwidth}
		\includegraphics[width=\linewidth]{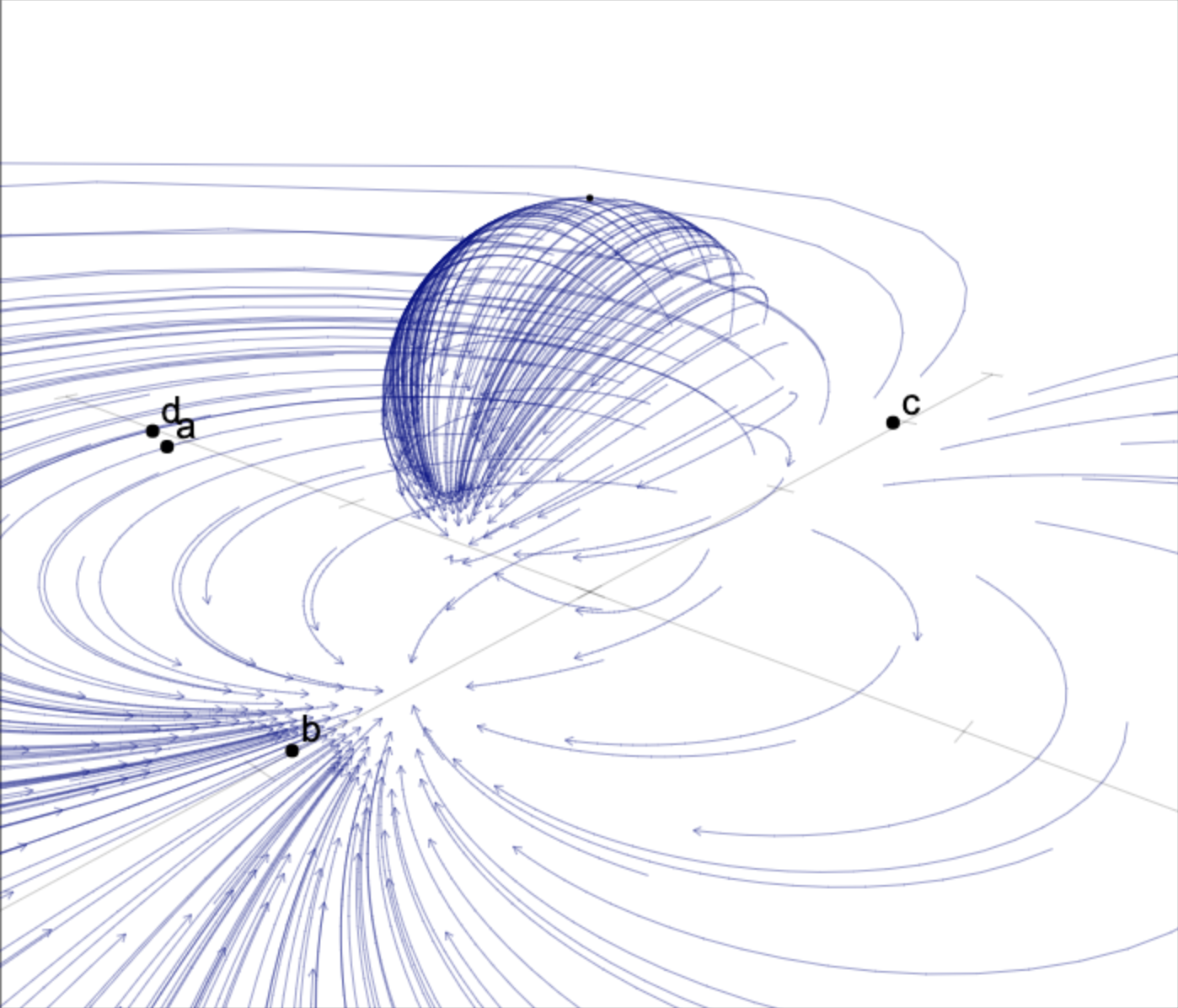}
		\label{fig:4d:sym}
	\end{subfigure}%
	\begin{subfigure}[b]{.19\textwidth}
		\includegraphics[width=\linewidth]{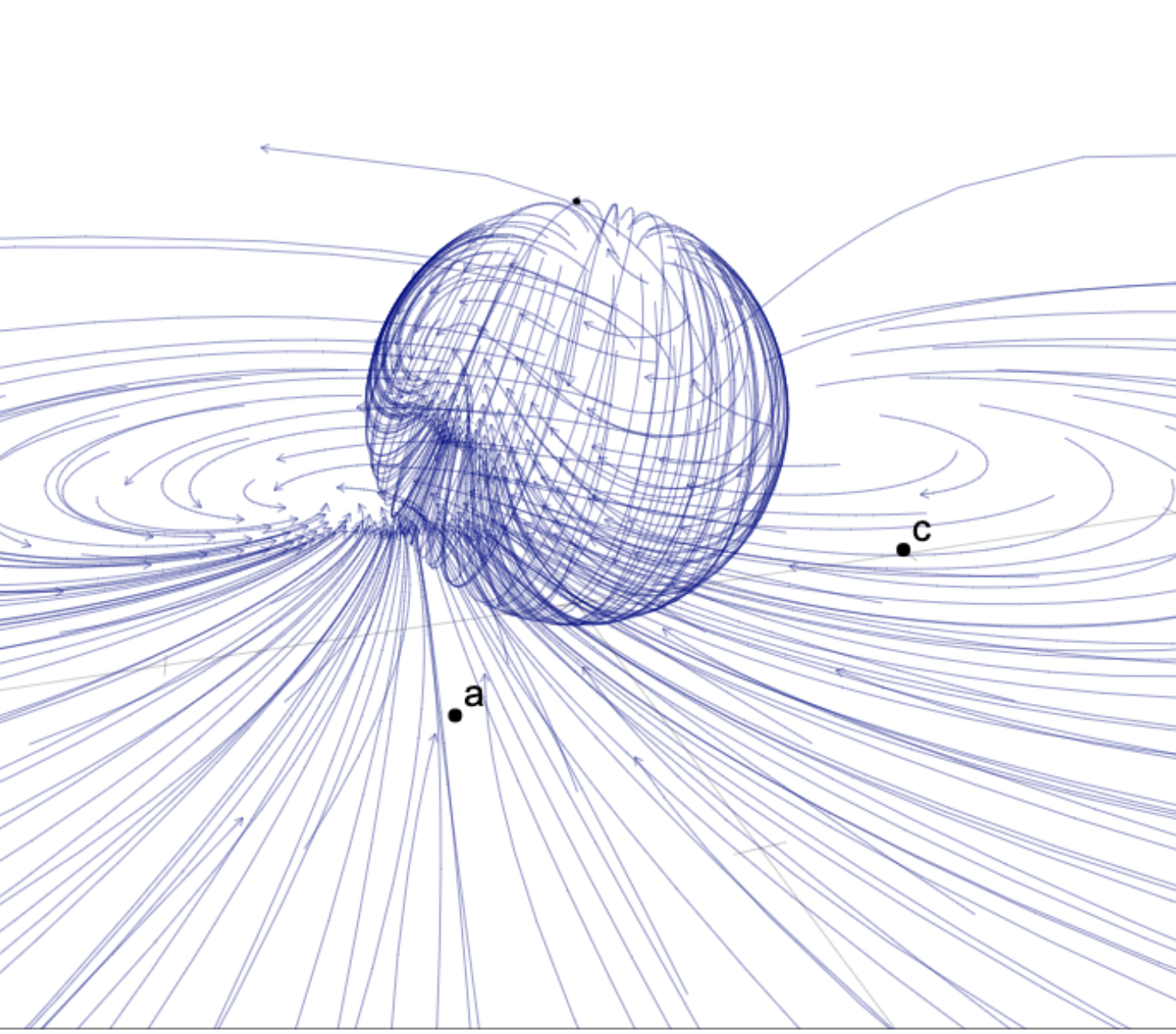}
		\label{fig:4a:noInj}%
	\end{subfigure}%
	\begin{subfigure}[b]{.19\textwidth}
		\includegraphics[width=\linewidth]{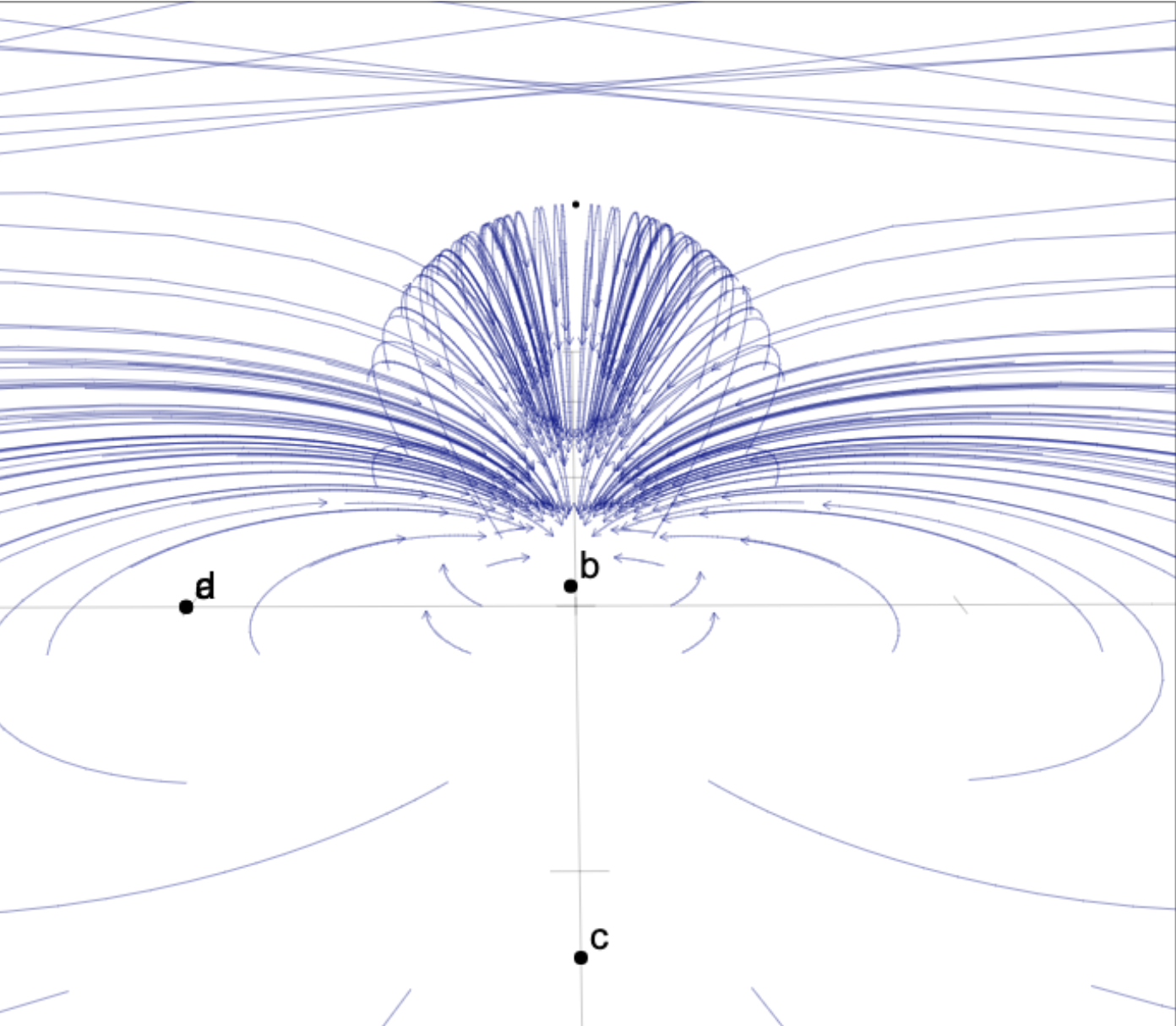}
		\label{fig:4a:noInj}%
	\end{subfigure}%
	\caption{Transformation functions illustrated \cite{visu} from left to right as circular, elliptic, hyperbolic, loxodromic, and parabolic. 
	It shows a Riemann Sphere on a complex plane with one or two fix points for each function.}
	\label{fig:tranfun}
\end{figure*}

\subsection{Variants of M\"obius Transformations}
Every M\"obius transformation has at most two fixed points $\gamma_1, \gamma_2$ on the Riemann Sphere obtained by solving $\vartheta(\gamma) = \gamma,$ \cite{richter2011perspectives} which gives 
        $\gamma_{1,2} = \frac{(a-d) \pm \sqrt{\Delta}}{2c}$. 
Depending on the number of fixed points, M\"obius transformations form parabolic or circular (one fixed point), elliptic as well as hyperbolic, and loxodromic (two fixed points) transformation functions illustrated in Figure~\ref{fig:tranfun}. 
A M\"obius transformation is performed on a grid by (a) a stereographic projection from complex plane to Riemann Sphere, (b) moving the Sphere, (c) stereographic projection from Sphere to plane. 
Each transformation has a constant $k = e^{\alpha + i \beta}$ which determines \emph{sparsity/density} of the transformation. 
$\beta$ is an expansion factor indicating the extent to which the fixed point $\gamma_1$ is repulsive ($\gamma_2$ is attractive). 
$\alpha$ is a rotation factor, determining the degree to which a transformation rotates the plane counter-clockwise around $\gamma_1$ (clockwise around $\gamma_2).$

\section{Related Work}

KGE models can be classified according to their embedding space. We discuss KGEs in Euclidean space and then describe related work for other geometries. 

\textbf{Euclidean Embedding Models}
A large number of KGE models such as TransE~\cite{bordes2013transe} and its variants~\cite{ji2015knowledge,lin2015learning,wang2014knowledge} as well as RotatE~\cite{sun2019rotate} are designed using translational or rotational (Hadamard product) score functions in Euclidean space. 
The score and loss functions of these models optimize the embedding vectors in a way that maximise the plausibility of triples, which is measured by the distance between rotated/translated head and tail vectors.
Some embedding models such as DisMult~\cite{yang2014embeddingDistmult}, ComplEx~\cite{complex2016trouillon}, QuatE~\cite{zhang2019quaternion}, and RESCAL~\cite{nickel2011three}, including our proposed model, are designed based on element-wise multiplication of transformed head and tail. 
In this case, the plausibility of triples is measured based on the angle of transformed head and tail.
A third category of KGE models are those designed on top of Neural Networks (NN) as score function such as ConvE~\cite{dettmers2018convolutional} and NTN~\cite{socher2013reasoning}.

\textbf{Non-Euclidean Embedding Models}
The aforementioned KGE models are limited to Euclidean space, which limits their ability to embed complex structures. 
Some recent efforts ~\cite{weber2018curvature,chami2020} investigated other spaces for embeddings of structures - often simpler structures than KGs.
For example, the hyperbolic space has been extensively studied in scale-free networks.
In recent work, learning continuous hierarchies from unstructured similarity scores using the Lorentz model was investigated~\cite{nickel2018learning}.
In~\cite{balazevic2019multi}, an embedding model dubbed MuRP is proposed that embeds multi-relational KGs on a Poincaré ball~\cite{ji2016knowledge}. 
MuRP only focuses on resolving the problem of embedding on KGs with multiple simultaneous hierarchies.
Overall, while the advantages of projective geometry are eminent in a wide variety of application domains, including computer vision and robotics, to our knowledge no investigation has focused on it within the context of knowledge graph embeddings.
\section{Method}
\label{sec:method}

Our method 5$^\bigstar$E inherits the five main pillars of projective transformation, namely translation, rotation, homothety, inversion and reflection.   
The transformations are performed in the following steps:
(1) \emph{element-wise stereographic projection} to map the head entity from a complex plane into a point on a Riemann Sphere; 
(2) \emph{relation-specific transformation} to move the Riemann Sphere into a new position and/or direction; 
(3) \emph{stereographic projection} to project the mapped head from the Riemann Sphere to a complex plane (1-3 in Equations~\ref{eq:mobTra} and \ref{eq:projTra}), 
(4) \emph{selection of complex inner product} between the transformed head and the tail (Equation \ref{eq:score}). 
\subsection{Model Formulation}

\textbf{Embedding on a Complex Projective Line} 
Given a triple $(h,r,t)$, the head and tail entities $h,t \in \mathcal{E}$ are embedded into a $d$ dimensional complex projective line i.e.~$\mathbf{h,t} \in \mathbb{CP}^d$. 
A relation $r \in \mathcal{R}$ is embedded into a $d$ dimensional vector $\mathbf{r}$  where each element is a $2 \times 2$  matrix. 
$\mathbf{r}$ contains four complex vectors $\mathbf{r_a}, \mathbf{r_b}, \mathbf{r_c}$ and $\mathbf{r_d} \in \mathbb{C}^d$. 
With $\mathbf{r}_{ai}, \mathbf{r}_{bi}, \mathbf{r}_{ci}, \mathbf{r}_{di}, \mathbf{h}_i, \mathbf{t}_i$, we refer to the $i$th element of $\mathbf{r}_{a}, \mathbf{r}_{b}, \mathbf{r}_{c}, \mathbf{r}_{d}, \mathbf{h}, \mathbf{t}$ respectively. 

\textbf{Relation-specific Transformation}
Based on preliminaries a projective transformation on a complex projective line has an equivalent transformation on the Riemann Sphere. 
Therefore, we use both of these perspectives in our  model formulation. 

\emph{M\"obius Representation of Transformation:} We use a relation-specific M\"obius transformation to map the head entity ($\mathbf{h}_{ri}$) from a source 
to a target complex plane ($\hat{\mathbb{C}}$).
    The transformation is performed using stereographic projection and transformation ($\vartheta$) on/from the Riemann Sphere. 
    To do so, we compute $\mathbf{h}_{ri}$ to specify the element-wise transformation:
    \begin{equation}
    \label{eq:mobTra}
    \begin{split}
        &\mathbf{h}_{ri} = g_{ri}(\mathbf{h}_i) = \vartheta(\mathbf{h}_i,\mathbf{r}_i) =  \frac{\mathbf{r}_{ai} \mathbf{h}_i + \mathbf{r}_{bi}}{\mathbf{r}_{ci} \mathbf{h}_i + \mathbf{r}_{di}}, \\ \, \, &\mathbf{r}_{ai}\mathbf{r}_{di} - \mathbf{r}_{bi}\mathbf{r}_{ci} \neq 0, i=1,\ldots,d. 
    \end{split}
    \end{equation}
This results in the relation-specific transformed head entity $\mathbf{h}_r = [\mathbf{h}_{r1}, \ldots, \mathbf{h}_{rd}].$

\emph{Projective Representation of Transformation:} Using homogeneous coordinates, we can also represent Equation~\ref{eq:mobTra} as a projective transformation:
    \begin{equation}
    \label{eq:projTra}
    \mathbf{h}_{ri} \doteq [g_r(\mathbf{h}_i),1]^T = \Im_{ri} [\mathbf{h}_i , 1]^T, 
    i = 1,\ldots,d,
    \end{equation}
    where $\doteq$ shows dehomogenization, $\Im_{ri} = \begin{bmatrix}
    \mathbf{r}_{ai} \,& \mathbf{r}_{bi}\\
    \mathbf{r}_{ci} \,& \mathbf{r}_{di}
    \end{bmatrix}$ and $\det{\Im_{ri}} \neq 0$  i.e.~$\Im_{ri}s$ are invertible.
    The matrix representation of Equation \ref{eq:projTra} is 
    $\mathbf{h}_{r} = \mathbf{R_r} [\mathbf{h}:\mathbf{1}],$
    where $\mathbf{R_r} = diag(\Im_{r1} \ldots, \Im_{rd})$ and $\mathbf{1}$ is a vector with all the elements being 1.

\textbf{Score Function} The correctness of triples in a KG is the similarity $\langle \mathbf{h}_{r}, \mathbf{t} \rangle$ between the relation-specific transformed head $\mathbf{h}_r$ and tail $\mathbf{t}$.
The model aims to minimize the angle between $\mathbf{h}_r$ and tail $\mathbf{t}$, i.e.~their product ($\langle \mathbf{h}_{r}, \mathbf{t} \rangle$) is maximized for positive triples.
For sampled negative triples, it is conversely minimized.
Overall, the score function for $5^\bigstar$E is
\begin{align}
\label{eq:score}
    f(h,r,t) = Re(\langle \mathbf{h}_{r}, \bar{\mathbf{t}}\rangle),
\end{align}
where $Re(x)$ is the real part of the complex number $x$. 

\subsection{Theoretical Analysis}
\label{sec:theo}
We first show that $5^\bigstar$E covers the five transformations.
We then discuss the capability of $5^\bigstar$E in preserving graph structures. 
We also prove $5^\bigstar$E is fully expressive and subsumes various state-of-the-art KGE models.

\textbf{M\"obius -- Composition of Transformations}
The M\"obius transformation in Equation~\ref{eq:mobTra} is a composition of a series of five subsequent transformations $\vartheta_1, \vartheta_2 \text{(two transformations in one)}, \vartheta_3$ and $\vartheta_4$ as shown in \cite{kisil2012geometry}:
    $\mathbf{h}_{ri} = \vartheta(\mathbf{h}_{i}, \mathbf{r}_i) = \vartheta_4 \circ \vartheta_3 \circ \vartheta_2 \circ \vartheta_1 (\mathbf{h}_{i}, \mathbf{r}_i)$,
where $\vartheta_1(\mathbf{x}, \mathbf{r}_i) = \mathbf{x} + \frac{\mathbf{r}_{di}}{\mathbf{r}_{ci}}$ (translation by $\frac{\mathbf{r}_{di}}{\mathbf{r}_{ci}}$), $\vartheta_2(\mathbf{x}) = \frac{1}{\mathbf{x}}$ (inversion and reflection w.r.t.~real axis), $\vartheta_3(\mathbf{x}, \mathbf{r}_i) = \frac{\mathbf{r}_{bi}\mathbf{r}_{ci} - \mathbf{r}_{ai}\mathbf{r}_{di}}{\mathbf{r}_{ci}^2} \mathbf{x}$ (homothety and rotation) and $\vartheta_4(\mathbf{x}, \mathbf{r}_i) = \mathbf{x} + \frac{\mathbf{r}_{ai}}{\mathbf{r}_{ci}}$ (translation
by $\frac{\mathbf{r}_{ai}}{\mathbf{r}_{ci}})$.
This shows that $5^\bigstar$E is capable of performing 5 transformations simultaneously. 

\textbf{Capturing Structures in a Neighborhood}
5$^\bigstar$E inherits various important properties of projective transformation as well as M\"obius transformations.
Because the projective linear group $PGL(2,\mathbb{C})$ is isomorphic to the M\"obius group, i.e., $ PGL(2,\mathbb{C}) \cong Aut(\mathbb{\hat{C}})$~\cite{kisil2012geometry}, the properties which are mentioned for Equation~\ref{eq:projTra} are also valid for Equation~\ref{eq:mobTra}. 
We investigate the inherited properties of 5$^\bigstar$E on \textit{clustering similar nodes of a neighborhood} and \textit{Capturing Sub-graph Structures}.
\emph{Clustering.}
The similarity of nodes in a KG is local, i.e.~nodes within a close neighborhood are more likely to be semantically similar~\cite{faerman2018lasagne,hamilton2017representation} than nodes at a higher distance. 
A projective transformation is a bijective conformal mapping, i.e.~it preserves angle locally but not necessarily the length.
It also preserves orientation after mapping~\cite{kisil2012geometry}. 
Therefore, 5$^\bigstar$E is capable of capturing similarity by preserving angle locally via a relation-specific transformation. 

Furthermore, the map $\pi: GL(2,\mathbb{C}) \rightarrow Aut(\mathbb{\hat{C}})$ is a group homomorphism, where $GL(2,\mathbb{C)}$ is a generalized linear group, which transfers the matrix $\Im$ into a M\"obius transformation $\vartheta$.
If $\det{\Im} = 1,$ then $\pi: SL(2,\mathbb{C}) \rightarrow Aut(\mathbb{\hat{C}})$ becomes limited to only perform a mapping from the special linear group $SL(2,\mathbb{C})$ to a M\"obius group that preserves volume and orientation.

In the context of KGs, after a relation-specific transformation (Equation~\ref{eq:projTra} or equivalently Equation~\ref{eq:mobTra}) of nodes in the head position to nodes in tail position, the relative distance of nodes can be preserved. 
From this ability, we expect that 5$^\bigstar$E is able to propagate the structural similarity from one group of nodes to another. 

\emph{Capturing Sub-graph Structures.}
Going beyond $SL(2, \mathbb{C})$ by changing the determinant to $\det\Im \neq 1$, the volume and orientation of the graph sub-structures are changed after transformation. 
Therefore, 5$^\bigstar$E is more flexible than current KGEs 
as those are not able to change volume and orientation of subgraphs.
This is visible in Figure~\ref{fig:all} when the graph includes a group of nodes in a path structure besides another group of nodes with a loop structure. 
In the vector space, other KGEs encounter a problem in preserving this type of graph structure due to the limited transformation abilities (not supporting inversion and reflection), whereas they work fine for homogeneous structures (e.g.~only lines or only circles). 
In contrast to this, 5$^\bigstar$E is capable of transforming heterogeneous structures 
due to the characteristics of a projective transformation~\cite{kisil2012geometry}.



\iftoggle{long}{
\begin{proposition}[cite??]
The M\"obius transformation on the Riemann Sphere has an Euler characteristic of 2.
\label{prop:pro4}
\end{proposition}

\textit{Discussion.} 
Euler characteristic is a number describing the shape of a topological space. Using polyhedral denoted by $\mathbb{P}$, the Euler characteristic is computed by $\mathbf{X}(\mathbb{P}) = V - E + H, $ where $V, E, H$ respectively denote the number of vertices, edges and faces of the polyhedral. 
The Euler characteristic is also well-defined for general surfaces by polygonization of the surfaces. 
From a topological perspective, the Euler Characteristic of a M\"obius transformation is equal to the Euler characteristic of a polyhederal with arbitrary vertices and edges. 
Therefore, it can be expected that any polyhedral with various number of nodes, edges and faces can be embedded by a capable KGE in a partitioned Riemann Sphere (note that any partition of the Sphere has an Euler characteristic of 2). 
}



\begin{figure*}[ht!]
	\centering 
	\begin{subfigure}[b]{0.2\textwidth}
		\includegraphics[width=\linewidth]{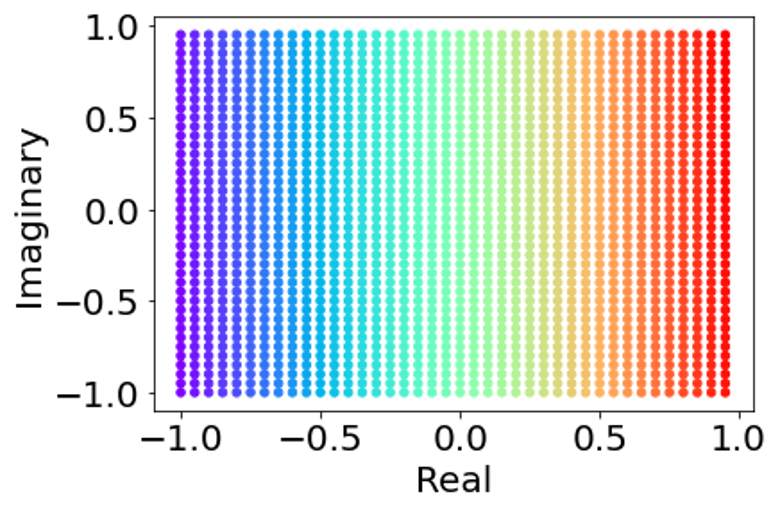}
		\caption{{\scriptsize Original Grid}}
		\label{fig:fb:noInj}
	\end{subfigure}%
	\begin{subfigure}[b]{0.2\textwidth}
		\includegraphics[width=\linewidth]{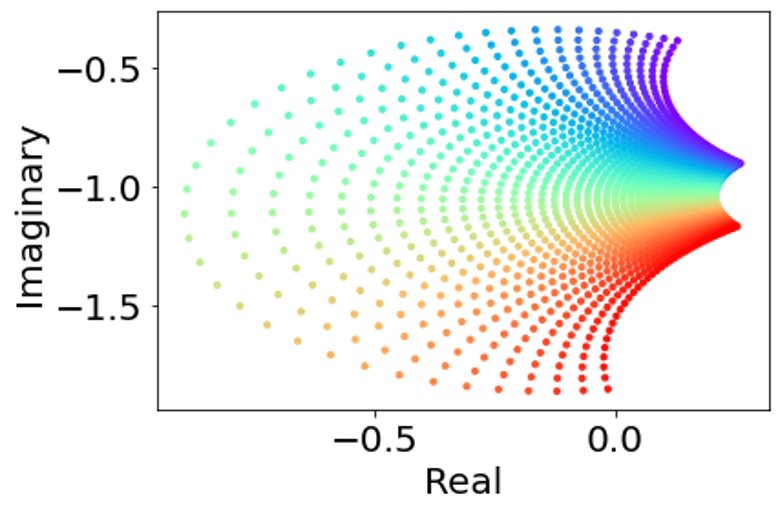}
		\caption{{\scriptsize hasPart relation}}
		\label{fig:fb:noInj}
	\end{subfigure}%
	\begin{subfigure}[b]{0.2\textwidth}
		\includegraphics[width=\linewidth]{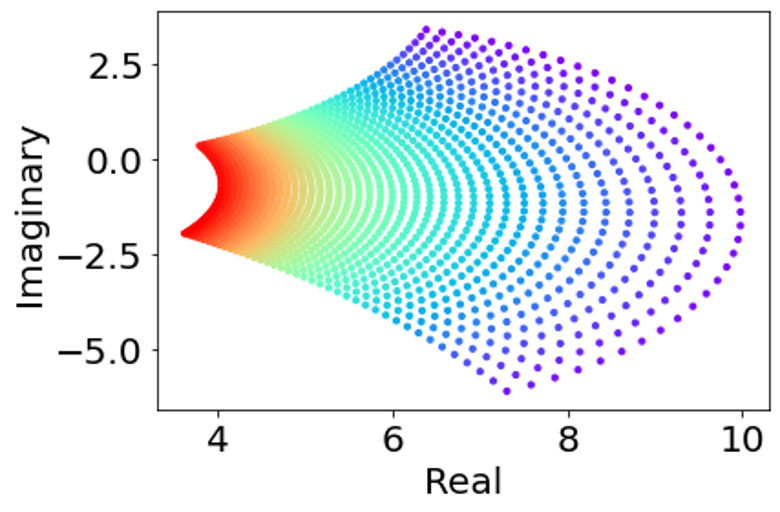}
		\caption{{\scriptsize partOf relation}}
		\label{fig:4c:eq}
	\end{subfigure}%
	\begin{subfigure}[b]{0.2\textwidth}
		\includegraphics[width=\linewidth]{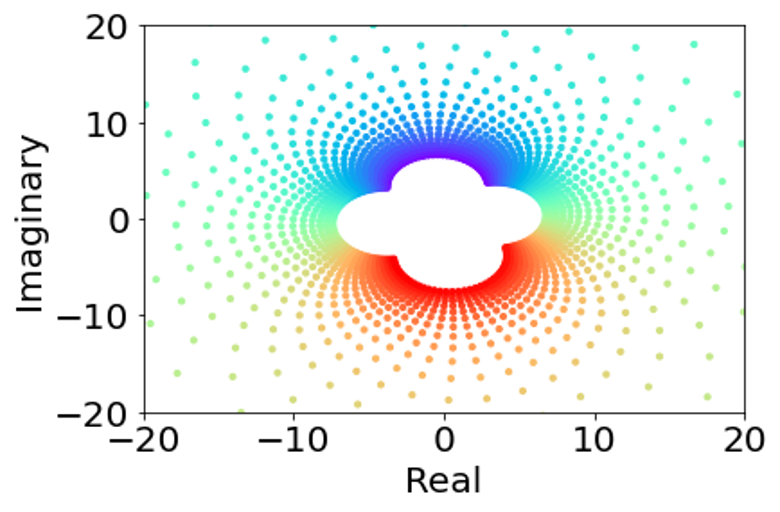}
		\caption{{\scriptsize hypernym}}
		\label{fig:4d:sym}
	\end{subfigure}%
	\begin{subfigure}[b]{.2\textwidth}
		\includegraphics[width=\linewidth]{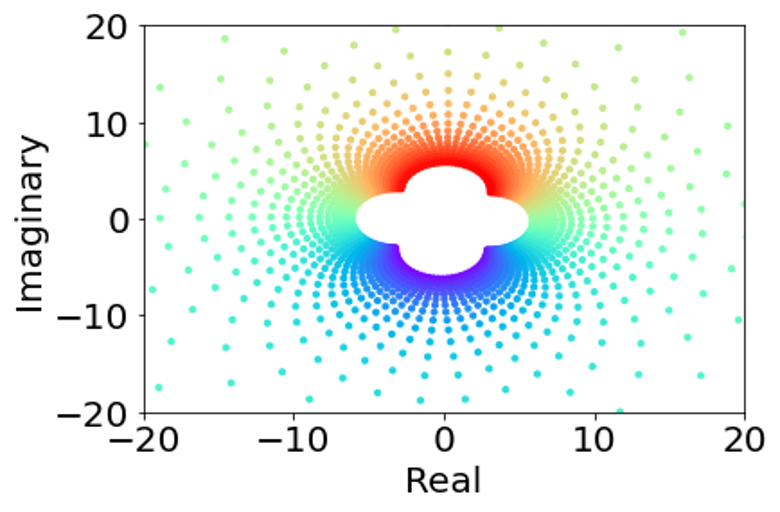}
		\caption{\scriptsize {hyponym}}
		\label{fig:4a:noInj}%
	\end{subfigure}%
	\caption{Learned 5$^\bigstar$E embeddings for a selected relations in WordNet. (b)-(e) show how the lines in (a) are transformed for a particular dimension (the 12th dimension in this case) of the embedding of the mentioned relations. 
	}
	\label{fig:grid}
\end{figure*}

\begin{figure*}[ht!]
\centering
	\begin{subfigure}[b]{0.20\textwidth}
		\includegraphics[width=\linewidth]{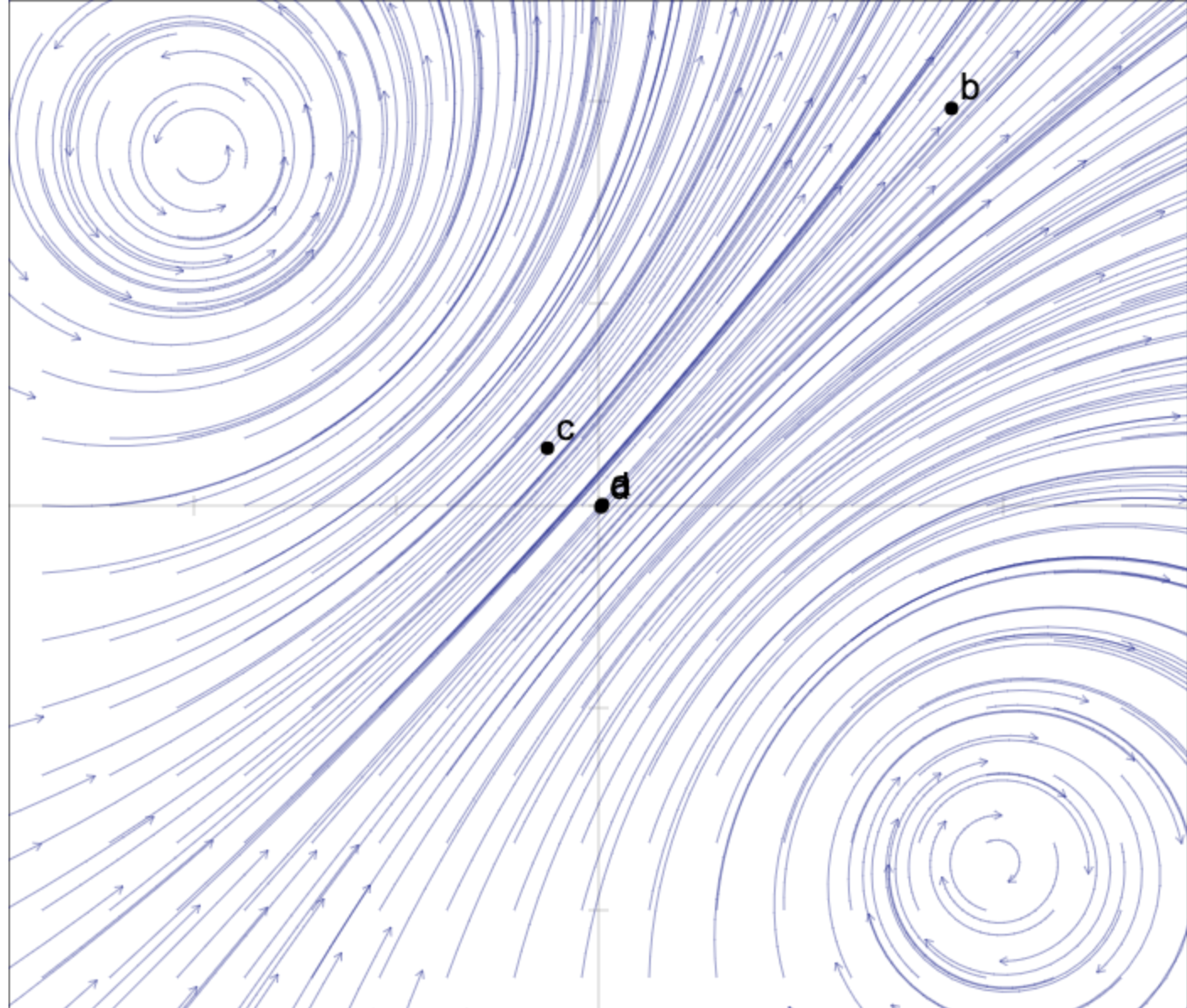}
		\caption{{\scriptsize 12-dim-hypernym}}
		\label{fig:fb:noInj}
	\end{subfigure}\hspace{5mm}
	\begin{subfigure}[b]{0.20\textwidth}
		\includegraphics[width=\linewidth]{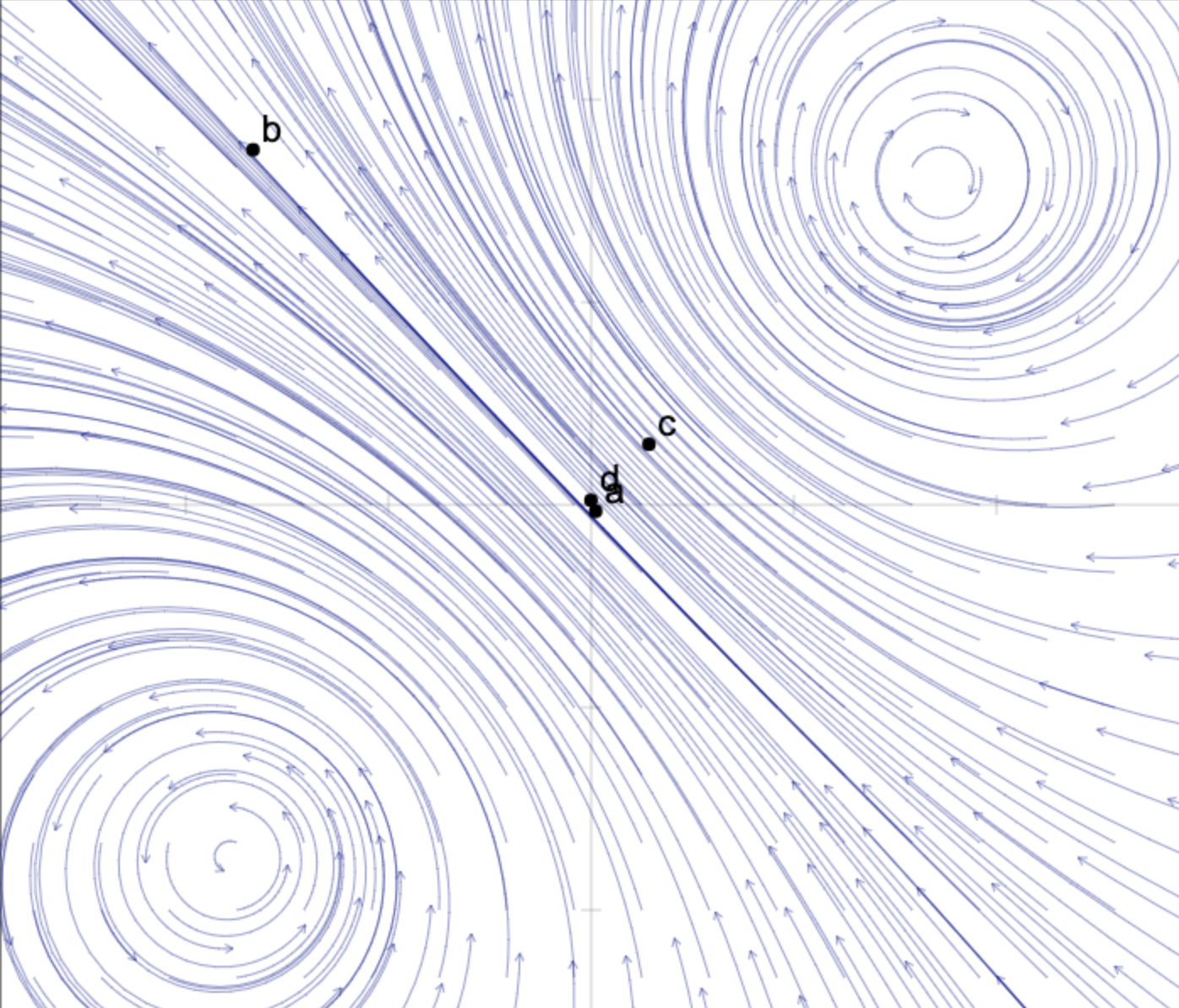}
		\caption{{\scriptsize 12-dim-hyponym}}
		\label{fig:fb:noInj}
	\end{subfigure}\hspace{5mm}
	\begin{subfigure}[b]{0.20\textwidth}
		\includegraphics[width=\linewidth]{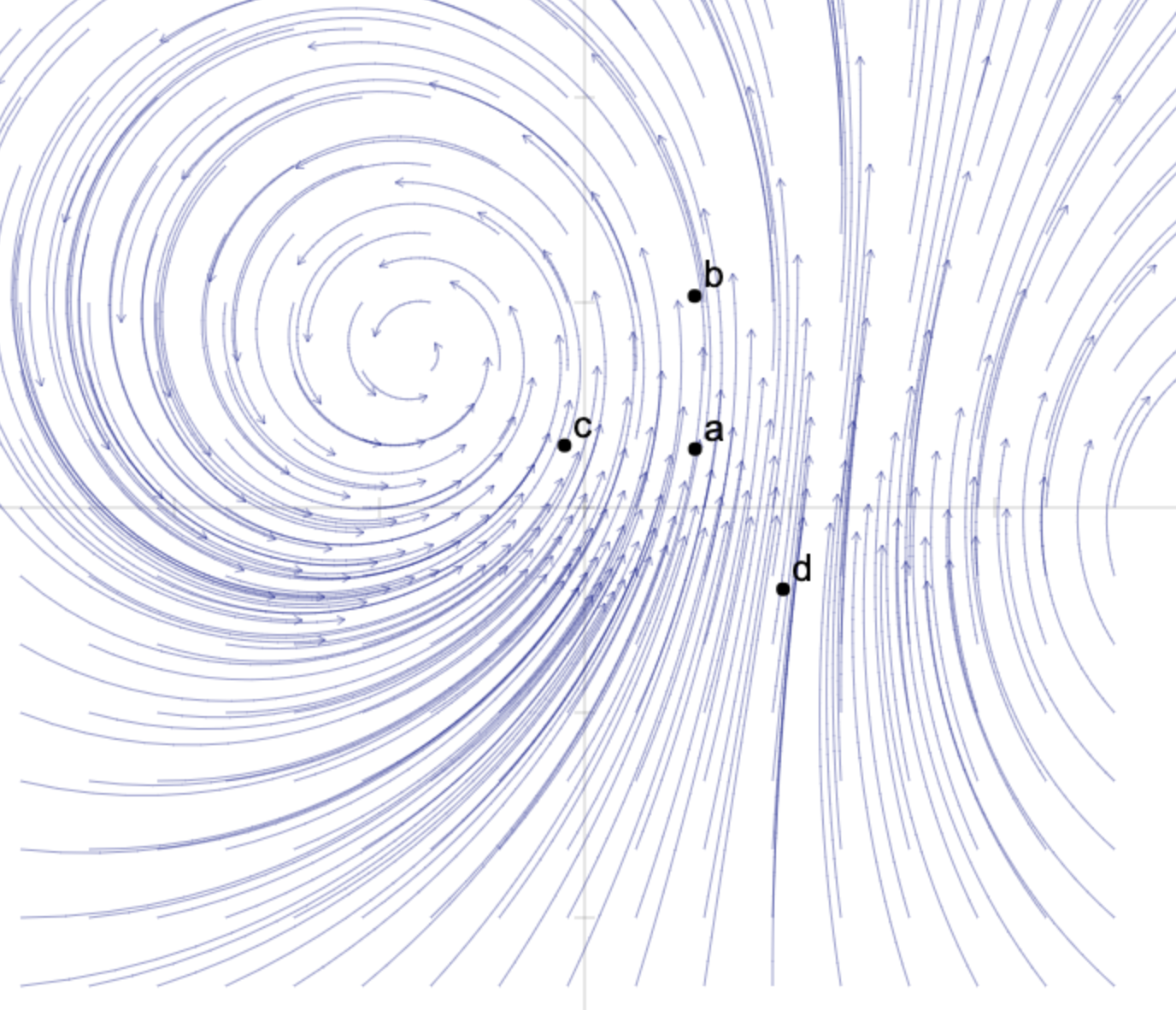}
		\caption{{\scriptsize 39-dim-haspart}}
		\label{fig:fb:noInj}
	\end{subfigure}\hspace{5mm}
	\begin{subfigure}[b]{0.20\textwidth}
		\includegraphics[width=\linewidth]{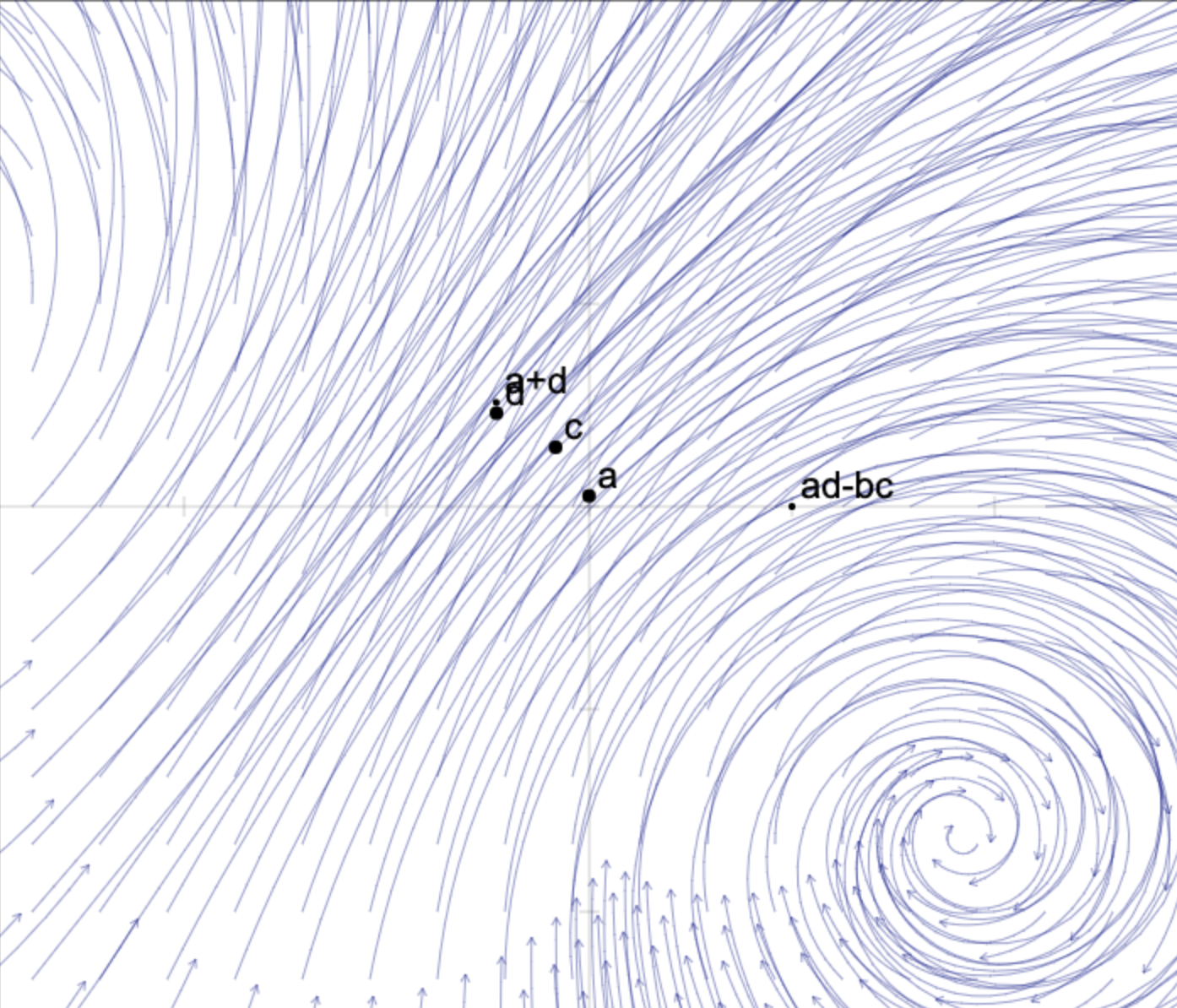}
		\caption{{\scriptsize 39-dim-partof}}
		\label{fig:fb:noInj}
	\end{subfigure}
	\caption{Embeddings for different relations usign the same dimension.}
	\label{fig:sameD}
\end{figure*}

\textbf{Subsumption of Other Models} 
We show that 5$^\bigstar$E subsumes other models and inherits their favorable characteristics in learning various graph patterns.
\begin{definition}
A model $M_1$ subsumes $M_2$ when any scoring over triples of a KG measured by model $M_2$ can also be obtained by $M_1$~\cite{wang2018multi}.
\label{def:exre}
\end{definition}
\vspace{-2mm}
\begin{proposition}
$5^\bigstar$E with variants of its score function subsumes DistMult, pRotatE, RotatE, TransE and ComplEx. 
Specifically, $5^\bigstar$E subsumes DistMult, ComplEx and pRotatE with $f(h,r,t) = Re(\langle \mathbf{h}_{r}, \bar{\mathbf{t}}\rangle)$ and
subsumes RotatE and TransE with score function $f(h,r,t) = -\| \mathbf{h}_r - \mathbf{t}\|$ (changed inner product to distance).
\label{prop:prop1}
\end{proposition}

\begin{definition}[from~\cite{kazemi2018simple}]
A model $M$ is \emph{fully expressive} if there exist assignments to the embeddings of the entities and relations, that accurately separate correct triples for any given ground truth.
\label{def:exre}
\end{definition}

\begin{corollary}
The $5^\bigstar$E model is fully expressive. 
\label{prop:col1}
\end{corollary}


\textbf{Inference of Patterns}
For relations which exhibit patterns in the form of $\textit{premise} \rightarrow \textit{conclusion}$, where \textit{premise} can be a conjunction of several triples, a model is said to be able to infer those if the implication holds for the score function, i.e.~if the score of all triples in the premise is positive then the score for the conclusion must be positive.
5$^\bigstar$E is able to infer reflexive, symmetric, inverse relation patterns as well as composition.


\begin{proposition}
Let $r_1, r_2, r_3 \in \mathcal{R}$ be relations and $r_3$ (e.g.~\textit{UncleOf}) be a composition of $r_1 (\textit{e.g.~BrotherOf})$ and $r_2 (\textit{e.g.~FatherOf}).$ 
$5^\bigstar$E infers composition with $\Im_{r_1} \Im_{r_2} = \Im_{r_3}.$ 
\label{prop:col12}
\end{proposition}

\begin{proposition}
Let $r_1 \!\in\! \mathcal{R}$ be the inverse of $r_2 \!\in\! \mathcal{R}$.
$5^\bigstar$E infers this pattern with $\Im_{r_1} = \Im_{r_2^{-1}}.$%
\label{prop:col13}%
\end{proposition} 

\begin{proposition}
Let $r \in \mathcal{R}$ be symmetric. $5^\bigstar$E infers the symmetric pattern if $\Im_{r} = \Im_{r}^{-1}.$ 
\label{prop:col14}
\end{proposition}

\begin{table*}[ht!]
\centering
\begin{tabular}{lllllllll}
 \toprule 
   \multirow{1}{*}{Model} & \multicolumn{4}{c}{FB15k237}         & \multicolumn{4}{c}{WN18RR}      \\ \cline{1-1} \cline{2-9} 
                          & MRR & Hits@1 & Hits@3 & Hits@10  & MRR & Hits@1 & Hits@3 & Hits@10 \\ \cline{2-9} 
                 TransE   &0.29&   -   &    -  &  0.47 &0.23& -     & -     & 0.50  \\ \cline{2-9} 
                 RotatE   &0.34& 0.24 &  0.38&  0.53 &0.48 &0.43  &0.49 &0.57 \\   \cline{2-9}
                TuckEr    &0.36&0.27  & 0.39 &  0.54&0.47&0.44&0.48  & 0.53  \\ \cline{2-9} 
                ComplEx   &0.36&  0.27& \underline{0.40} &  \underline{0.56}.  & 0.49 & 0.44  & 0.50  & 0.58   \\ \cline{2-9} 
                QuatE     &\underline{0.37} &  0.27&  \underline{0.40}&  \underline{0.56}&0.48& 0.44 & 0.50 & 0.57  \\ \cline{2-9} 
                ConvE      &0.33&  0.24& 0.36 & 0.50  &0.43 & 0.40  & 0.44  &0.52    \\ \cline{2-9}
                MuRP      & 0.34 & 0.24  & 0.37 &  0.52 &  0.48 &  0.44 &   0.50 &   0.57 \\  
\hline
     \bottomrule 
                $5^\bigstar$E \scriptsize{d = 500} & \underline{0.37} & \underline{0.28} & \underline{0.40} & \underline{0.56} &  \underline{0.50} & \underline{0.45}& \underline{0.51}& \underline{0.59} \\ 
                $5^\bigstar$E \scriptsize{d = 100} & 0.35& 0.26& 0.38& 0.53 & 0.47 & 0.41 &  0.50 &  0.58  \\ 
\hline
\end{tabular}
\caption{Link prediction results on d FB15k-237, and WN18RR. The Results of TransE, QuatE, RotatE, and ConvE are taken from~\cite{zhang2019quaternion}, TuckER from~\cite{tucker2019balavzevic} and MuRP from~\cite{balazevic2019multi}, and ComplEx has been experimented.}
\label{table:result_table_wnfb}
\end{table*}

\iftoggle{long}{
\textit{Discussion.}
According to Corollary \ref{prop:col14}, $5^\bigstar$E infers the symmetric pattern when the model learns $(h,r,t)$ as a positive triple, then enforced with the symmetric constrain of $\Im_{r} = \Im_{r}^{-1},$ the score of ($t,r,h$) is also considered as positive.
Now let embedding dimension equals to $d$. 
There are $2^d-1$ different unique representations for entities which are connected with a symmetric relation $r$. 
In order to show this possibility, let us consider the simplest variant of transformation by assuming the length of $|\mathbf{r}_{ai}|$ to be 1, and $\mathbf{r}_{bi} = \mathbf{r}_{ci} = 0,$ and $\mathbf{r}_{di} = 1$. 
The resulting transformation is a rotation with $\theta_{r_{ai}}$ degrees, $i.e.~\mathbf{r}_{ai} = e^{\theta_{r_{ai}}}.$ 
Since $r$ is a symmetric relation, we have $\theta_{r_{ai}} = 0, \pi.$ 
Therefore, either $\mathbf{h}_i = \mathbf{t}_i, $ or $\mathbf{h}_i = -\mathbf{t}_i$ are the two possible cases. 
For a given head and relation where $r$ symmetric, there are at least $2^d - 1$ tail entity embeddings satisfying the symmetric pattern constraint.
}



\begin{proposition} 
Let $r \in \mathcal{R}$ be a reflexive relation.
In dimension $d$, $5^\bigstar$E infers reflexive patterns with $O(2^d)$ distinct representations of entities if the fixed points are non-identical.
\label{prop:col15}
\end{proposition} 
TransE only infers composition and inverse patterns. 
RotatE is capable of inferring more patterns but is not fully expressive. 
ComplEx infers these patterns and is fully expressive. 
However, it has less flexibility than our model in learning complex structures due to using only rotation and homothety.
Therefore, it is only capable of preserving homogeneous structures (see Figure~\ref{fig:llcc}). 

\section{Experiments and Results}
\label{sec:exp}

\begin{table*}[h!]
\centering
\begin{tabular}{lllllllll}
 \toprule 
   \multirow{1}{*}{Model} & \multicolumn{4}{c}{NELL-995-h100}         & \multicolumn{4}{c}{NELL-995-h75}      \\ \cline{1-1} \cline{2-9} 
                          & MRR & Hits@1 & Hits@3 & Hits@10  & MRR & Hits@1 & Hits@3 & Hits@10 \\ \cline{2-9} 
MuRE & 0.36 & 0.27 & 0.40 & 0.53 & 0.36 & 0.27 & 0.40 & \underline{0.53}  \\  \cline{2-9} 
MuRP & 0.36 & 0.27 & 0.40 & 0.53 & 0.36 & 0.28 & 0.40 & 0.52  \\
 \hline 
 ComplEx  & 0.35 & 0.27 & 0.40 & 0.52 & 0.35 & 0.27 & 0.39 & 0.51\\
  \hline 
 QuatE  & 0.35 & 0.26 & 0.40 & 0.53 & 0.36 & 0.27 & 0.41 & 0.52 \\
\hline
\hline
$5^\bigstar$E \scriptsize{d = 200} & \underline{0.37}  & \underline{0.28} & \underline{0.42} & \underline{0.54}& \underline{0.37} & \underline{0.28} &  \underline{0.41} &  \underline{0.53}  \\
\hline
$5^\bigstar$E \scriptsize{d = 100} & 0.36  & \underline{0.28} & 0.40 & \underline{0.53} & 0.36 & 0.27 &  0.39 &  \underline{0.53}  \\ 
 \toprule 
   \multirow{1}{*}{Model} & \multicolumn{4}{c}{NELL-995-h50}  & \multicolumn{4}{c}{NELL-995-h25}      \\  \cline{1-1} \cline{2-9} 
   \toprule 
                     & MRR & Hits@1 & Hits@3 & Hits@10 & MRR & Hits@1 & Hits@3 & Hits@10 \\  \cline{2-9} 
MuRE & 0.37 & 0.28 & 0.42 & \underline{0.54} & 0.37 & 0.29 & 0.40 &  0.52  \\  \cline{2-9} 
MuRP  & 0.37 & 0.28 & 0.42 & 0.54 & 0.36 & 0.28 & 0.40 & 0.51  \\
 \hline 
 ComplEx & 0.37 & 0.29 & 0.41 & 0.52 & 0.37 & 0.30 & 0.40 & 0.51\\
  \hline 
 QuatE  & 0.36 & 0.27 & 0.40 & 0.53 & 0.36 & 0.28 & 0.40 & 0.51\\
 \hline 
 \hline
$5^\bigstar$E  \scriptsize{d = 200} & \underline{0.38} & \underline{0.30} & \underline{0.43} & \underline{0.54} & \underline{0.39} & \underline{0.31} & \underline{0.43} & \underline{0.53}\\ 
\hline
$5^\bigstar$E  \scriptsize{d = 100} & 0.38 & 0.29 & \underline{0.43} & \underline{0.54} & 0.37 & 0.30 & 0.41 & 0.52\\ 
     \bottomrule 
\end{tabular}
\caption{Link prediction results on KGs with various percentages of hierarchical relations including NELL-995-h25 (25\% hierarchical relation) and NELL-995-h50 (50\%) as well as NELL-995-h75 (75\%) and NELL-995-h100 (100\%). The results for ComplEx, QuatE, and $5^\bigstar$E  are from own experiments - all others are taken from their original works.}
\label{table:result_table_nell}
\end{table*}

\textbf{Experimental Setup}
Following the best practices of evaluations for embedding models, we consider the most-used metrics (Mean) Reciprocal Rank (MRR) and Hits@n (n = 1, 3, 10).
We evaluated our model on four widely used benchmark datasets namely FB15k-237~\cite{toutanova2015observed}, WN18RR~\cite{dettmers2018convolutional} 
, and NELL (four different versions as NELL-995-h25, NELL-995-h50, NELL-995-h75 and NELL-995-h100) ~\cite{xiong2017deeppath,balazevic2019multi}. 
The FB15k-237 and WN18RR datasets both include several relational patterns such as composition (e.g. $awardnominee/\dots/nominatedfor$), symmetry (e.g $derivationally\_related\_form$ in WN18RR), and anti-symmetry (e.g $has\_part$ in WN18RR). 
The WN18RR dataset includes hierarchical relations such as $hypernym$ and $has\_part$, which are typical examples for shaping a path structure, and relations such as $also\_see$, $similar\_to$, which are candidates for loop structures. 
The different variants of the NELL dataset include several relations that contain loops ($hassibling$, $competeswith$, $synonymfor$) as well as relations forming hierarchical paths ($subpartof$).

We compare the best performing models namely \emph{TransE}~\cite{bordes2013transe}, \emph{RotatE}~\cite{sun2019rotate}, \emph{TuckEr}~\cite{tucker2019balavzevic}, \emph{ComplEx}~\cite{complex2016trouillon}, \emph{QuatE}~\cite{zhang2019quaternion}, 
\emph{MuRP}~\cite{balazevic2019multi},
\emph{ConvE}~\cite{dettmers2018convolutional} and 
\emph{SimplE}~\cite{kazemi2018simple}.
Our model is implemented in Pytorch\footnote{\url{https://pytorch.org/}} and the code is available online\footnote{\url{https://bit.ly/2NXplO1}}.
Similar to QuatE and ComplEx, we developed our model on top of a standard framework~\cite{lacroix2018canonical}, applied 1-N scoring loss with N3 regularization, and added reverse counterparts of each triple to the train set.

\begin{figure*}[ht!]
	\centering 
	\begin{subfigure}[b]{0.18\textwidth}
		\includegraphics[width=\linewidth]{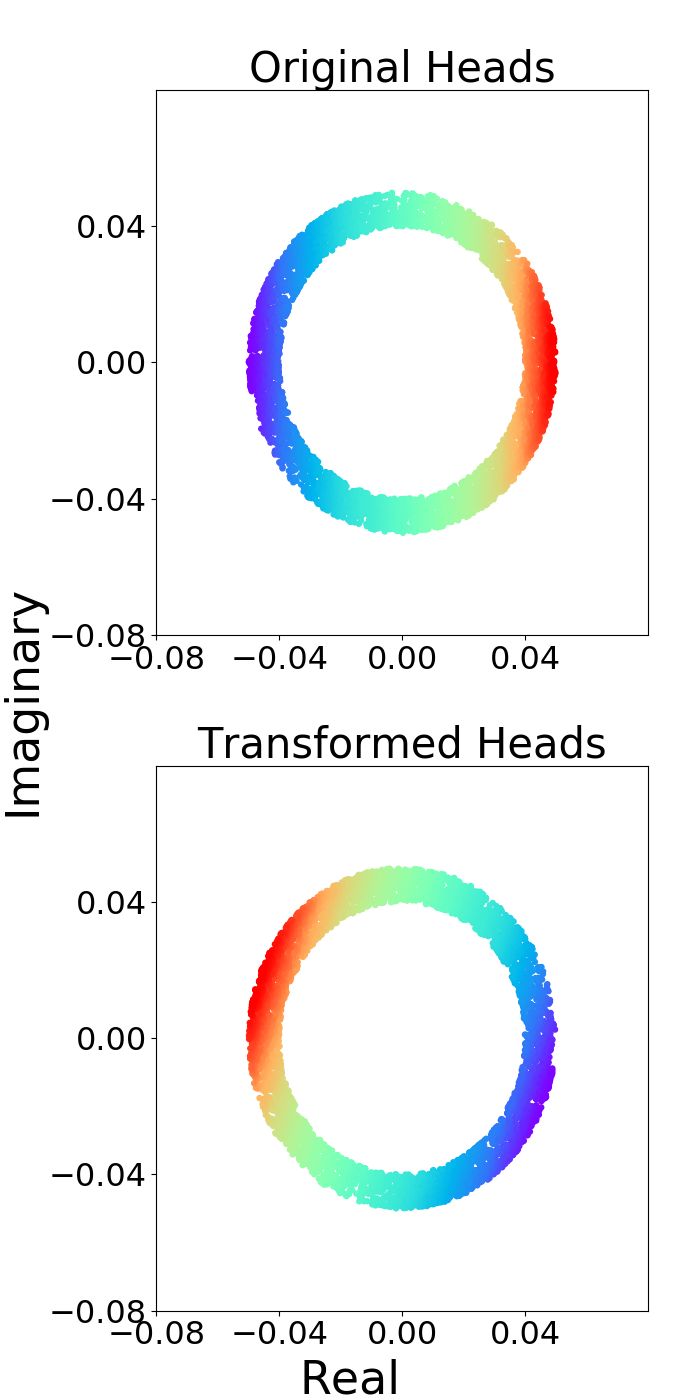}
		\caption{{\scriptsize Circle-to-circle}}
		\label{fig:fb:noInj}
	\end{subfigure}%
	\begin{subfigure}[b]{0.18\textwidth}
		\includegraphics[width=\linewidth]{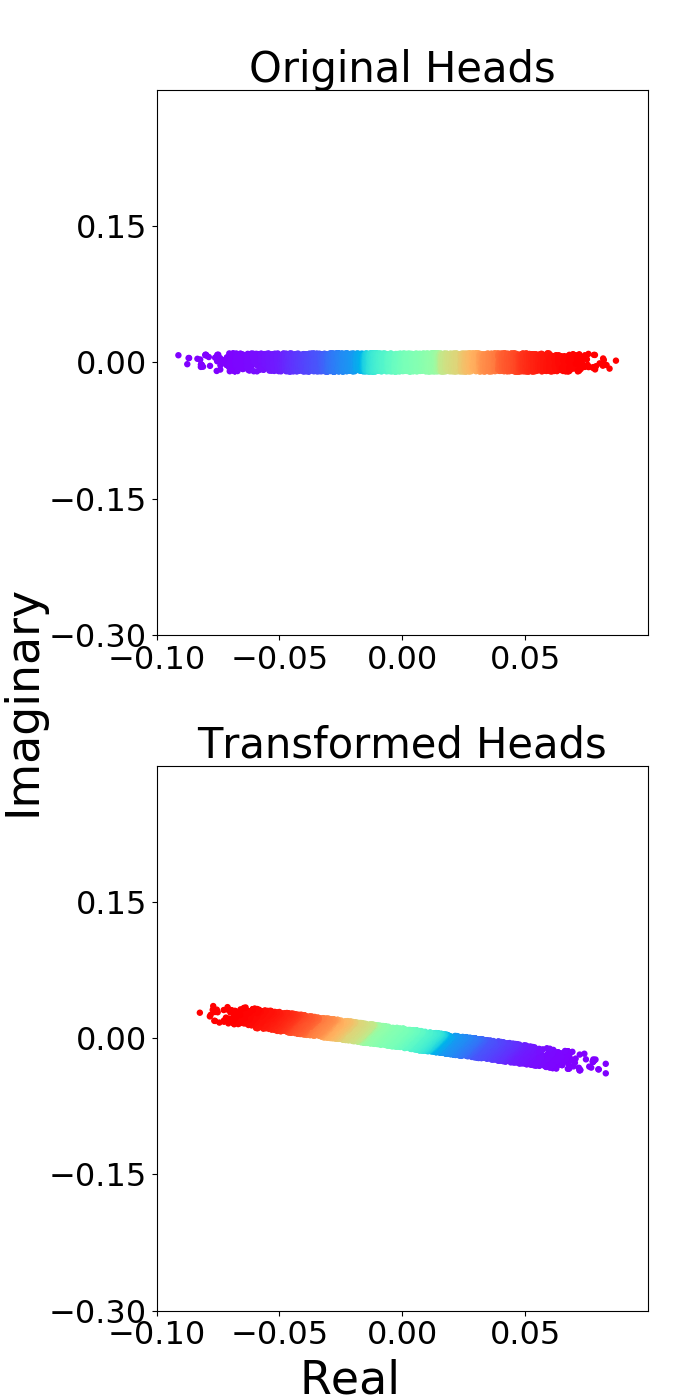}
		\caption{{\scriptsize line-to-line}}
		\label{fig:4d:sym}
	\end{subfigure}%
	\begin{subfigure}[b]{0.18\textwidth}
		\includegraphics[width=\linewidth]{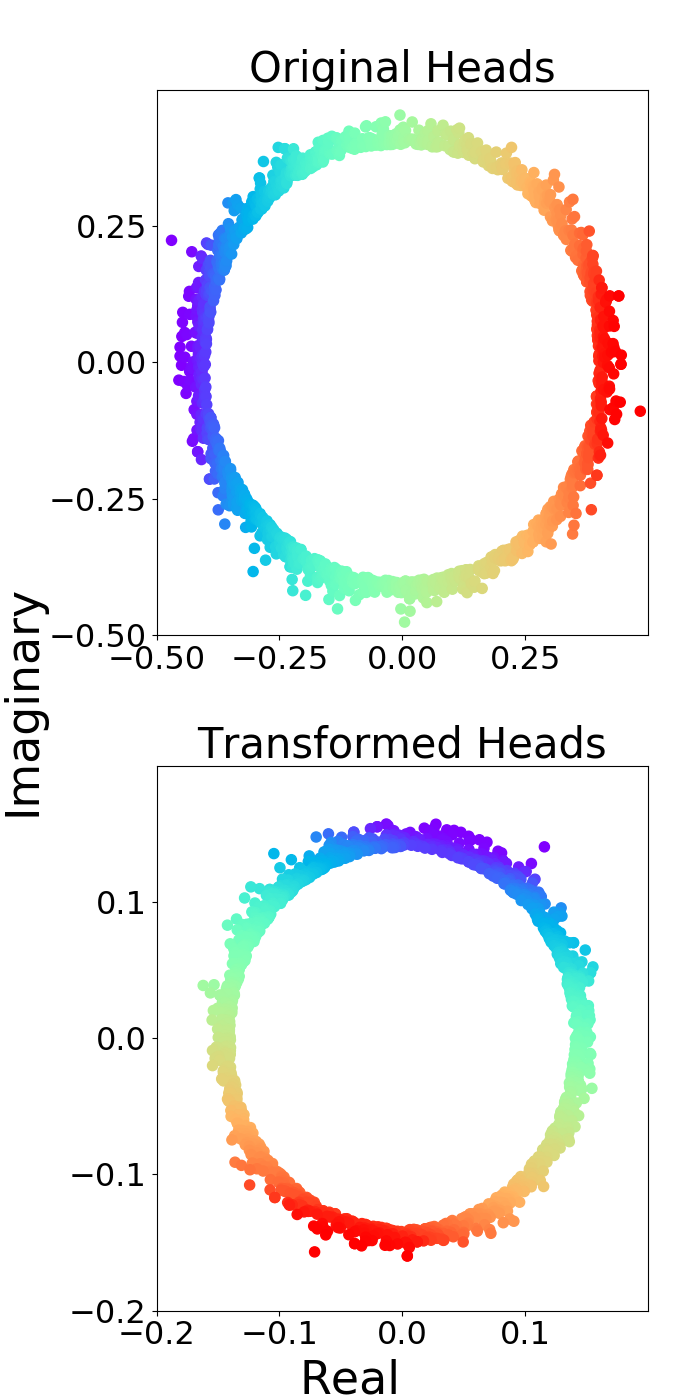}
		\caption{{\scriptsize Circle-to-circle}}
		\label{fig:fb:noInj}
	\end{subfigure}%
	\begin{subfigure}[b]{0.18\textwidth}
		\includegraphics[width=\linewidth]{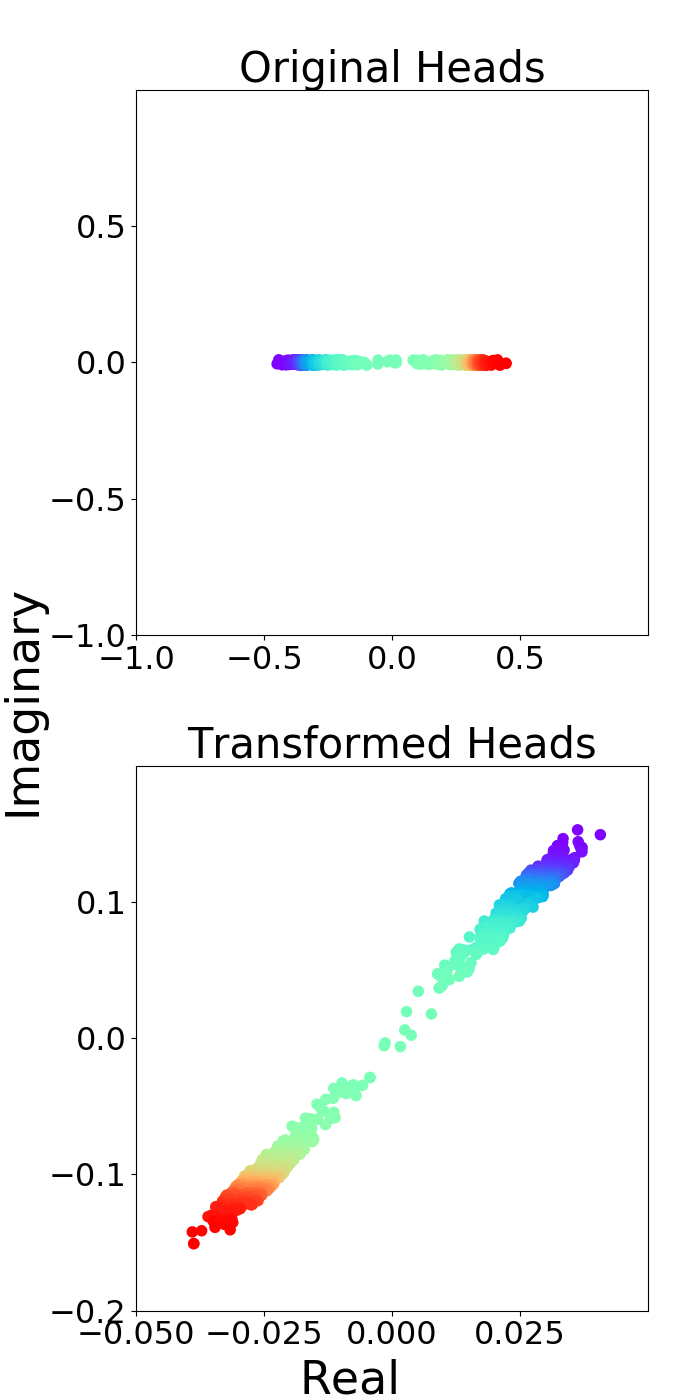}
		\caption{{\scriptsize line-to-line}}
		\label{fig:4d:sym}
	\end{subfigure}%
	\caption{Types of transformations that RotatE (a,b) and ComplEx (c,d) learned on relation "hypernym" in WordNet. Each pair of images visualises one dimension of the relation embedding. The top images show the embeddings of head entities of this relation in the KG at this dimension. Each entity embedding is visualised as a colored dot. The bottom images show the results of applying the relation specific transformation (the entity colors are preserved).
	}
	\label{fig:llcc}
\end{figure*}

\begin{figure*}[ht!]
	\centering 
	\vspace{-10pt}
	\begin{subfigure}[b]{0.18\textwidth}
		\includegraphics[width=\linewidth]{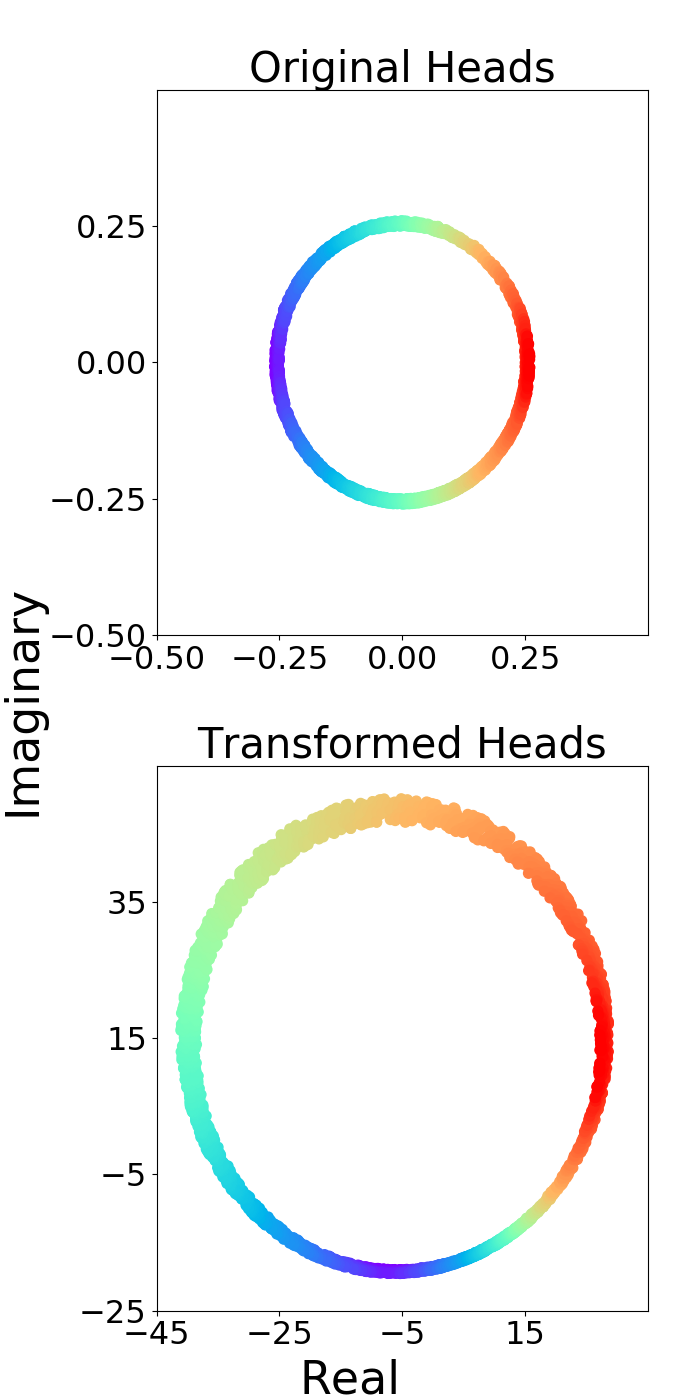}
		\caption{{\scriptsize Circle-to-circle}}
		\label{fig:fb:noInj}
	\end{subfigure}%
	\begin{subfigure}[b]{0.18\textwidth}
		\includegraphics[width=\linewidth]{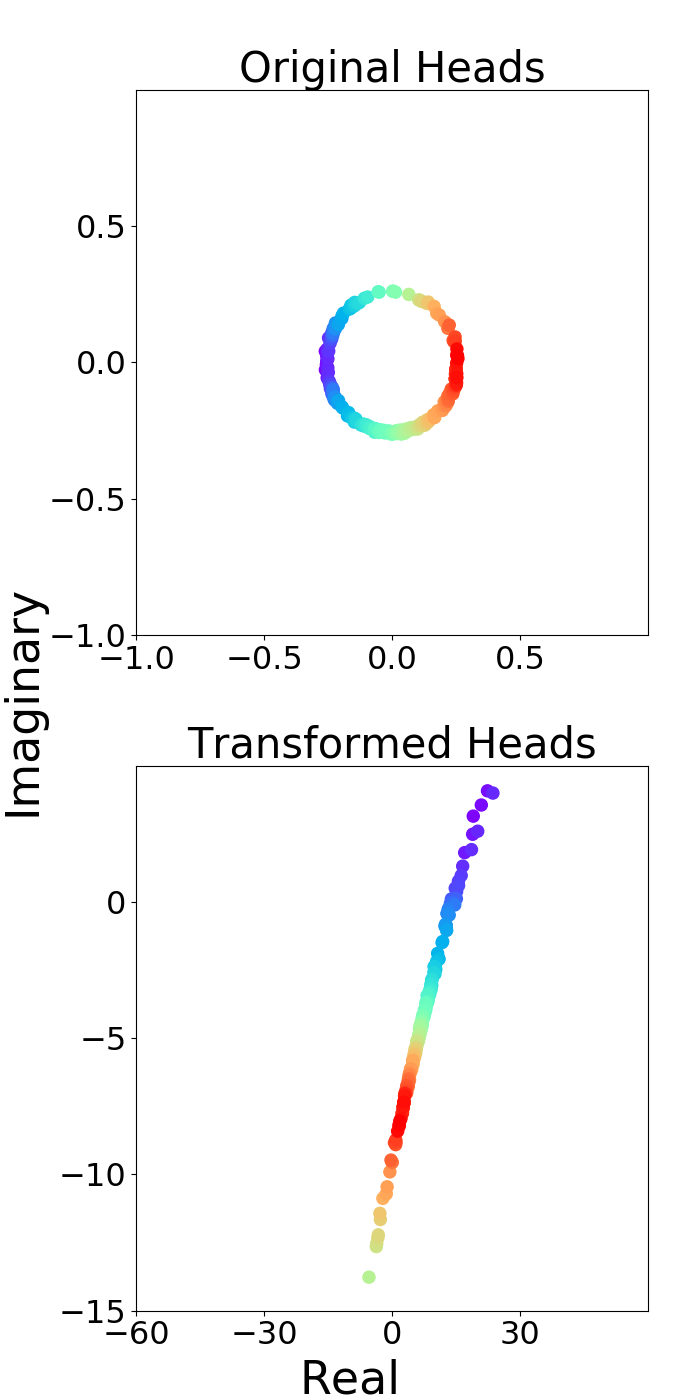}
		\caption{{\scriptsize circle-to-line}}
		\label{fig:fb:noInj}
	\end{subfigure}%
	\begin{subfigure}[b]{0.18\textwidth}
		\includegraphics[width=\linewidth]{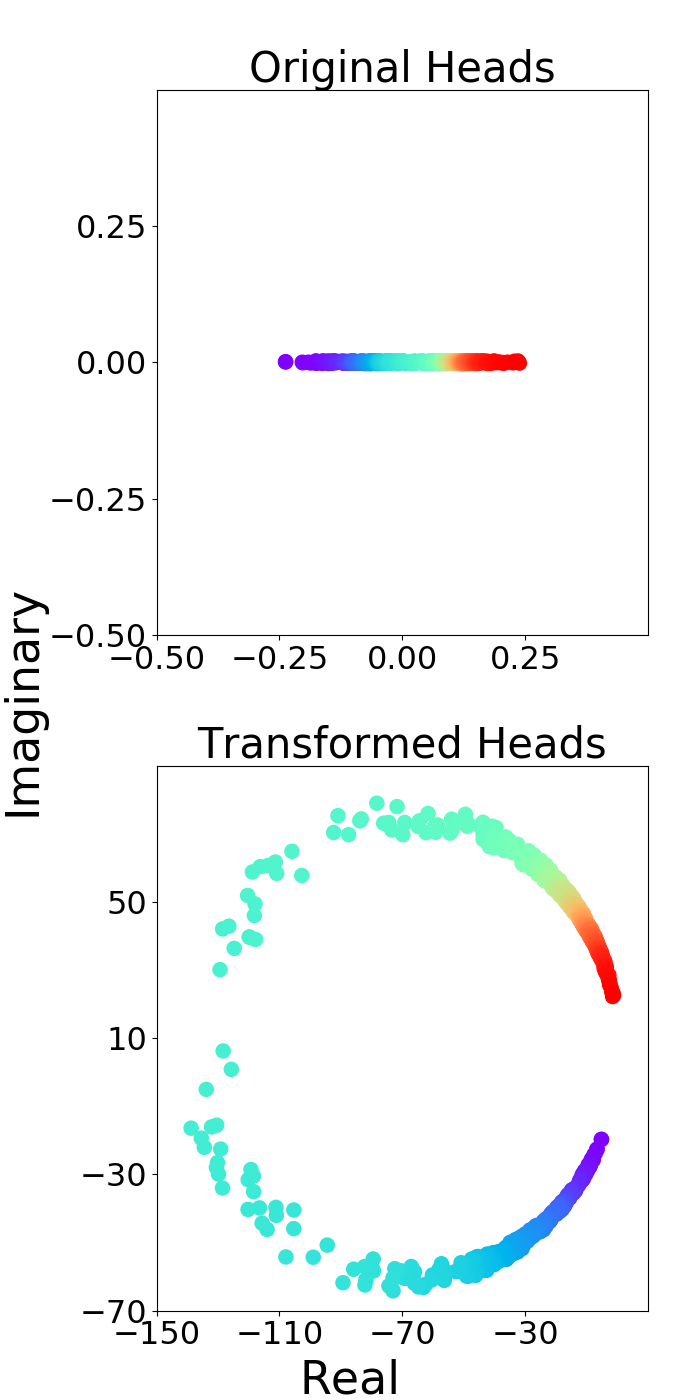}
		\caption{{\scriptsize line-to-circle}}
		\label{fig:4c:eq}
	\end{subfigure}%
	\begin{subfigure}[b]{0.18\textwidth}
		\includegraphics[width=\linewidth]{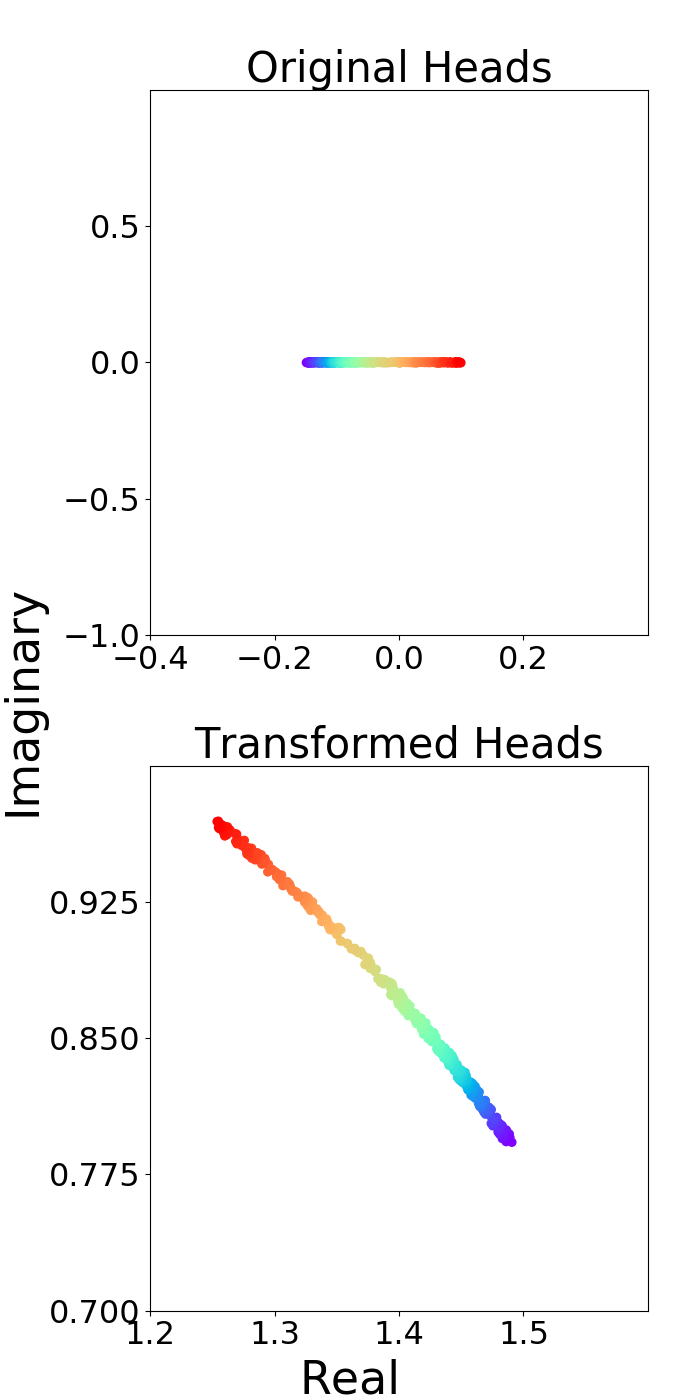}
		\caption{{\scriptsize line-to-line}}
		\label{fig:4d:sym}
	\end{subfigure}%
	\caption{Types of transformations that 5$^\bigstar$E learned on relation "hypernym" in WordNet.
	Each pair of images visualises one dimension of the relation embedding. The top images show the embeddings of head entities of this relation in the KG at this dimension. Each entity embedding is visualised as a colored dot. The bottom images show the results of applying the relation specific transformation (the entity colors are preserved).
	}
	\label{fig:lccl}
\end{figure*}

\medskip
\noindent
\textbf{Results.}
The results of comparing 5$^\bigstar$E to other models on FB15k-237 and WN18RR are shown in Table~\ref{table:result_table_wnfb} ($d =$ 100 and 500) and on NELL in Table~\ref{table:result_table_nell} ($d =$ 100 and 200). 
Our model outperforms all other models across all metrics on WN18RR, which is a dataset with many hierarchical relations as well as relations forming loops such as $similar-to$. 
Although MuRP is specifically designed for hierarchical data, 5$^\bigstar$E still achieves a better performance. 
Generally, we can observe that 5$^\bigstar$E obtains competitive results with a low dimension ($d =$ 100) on WN18RR for the Hits@3 and Hits@10 metrics.

The evaluation shows that rotation-based models (RotatE, QuatE, and ComplEx) obtain state-of-the-art results on the FB15k-237 dataset (with fewer hierarchical paths than WN18RR). Our model, which covers rotation and transformation, obtains similar results to those models. On this dataset, there is no evident benefit of supporting further transformations. We additionally used different versions of the NELL dataset,  which are specifically designed to have a particular percentage of hierarchical relations.
5$^\bigstar$E outperforms other models in all NELL dataset versions.

Overall, the competitive results of 5$^\bigstar$E show that additional transformations have a positive effect for the link prediction task. They also indicate that the additional transformations do not lead to over-fitting problems compared to single-transformation models (or at least the positive effects outweigh potential overfitting).

\begin{figure*}[ht]
	\centering 
	\begin{subfigure}[b]{0.32\textwidth}
		\includegraphics[width=\linewidth]{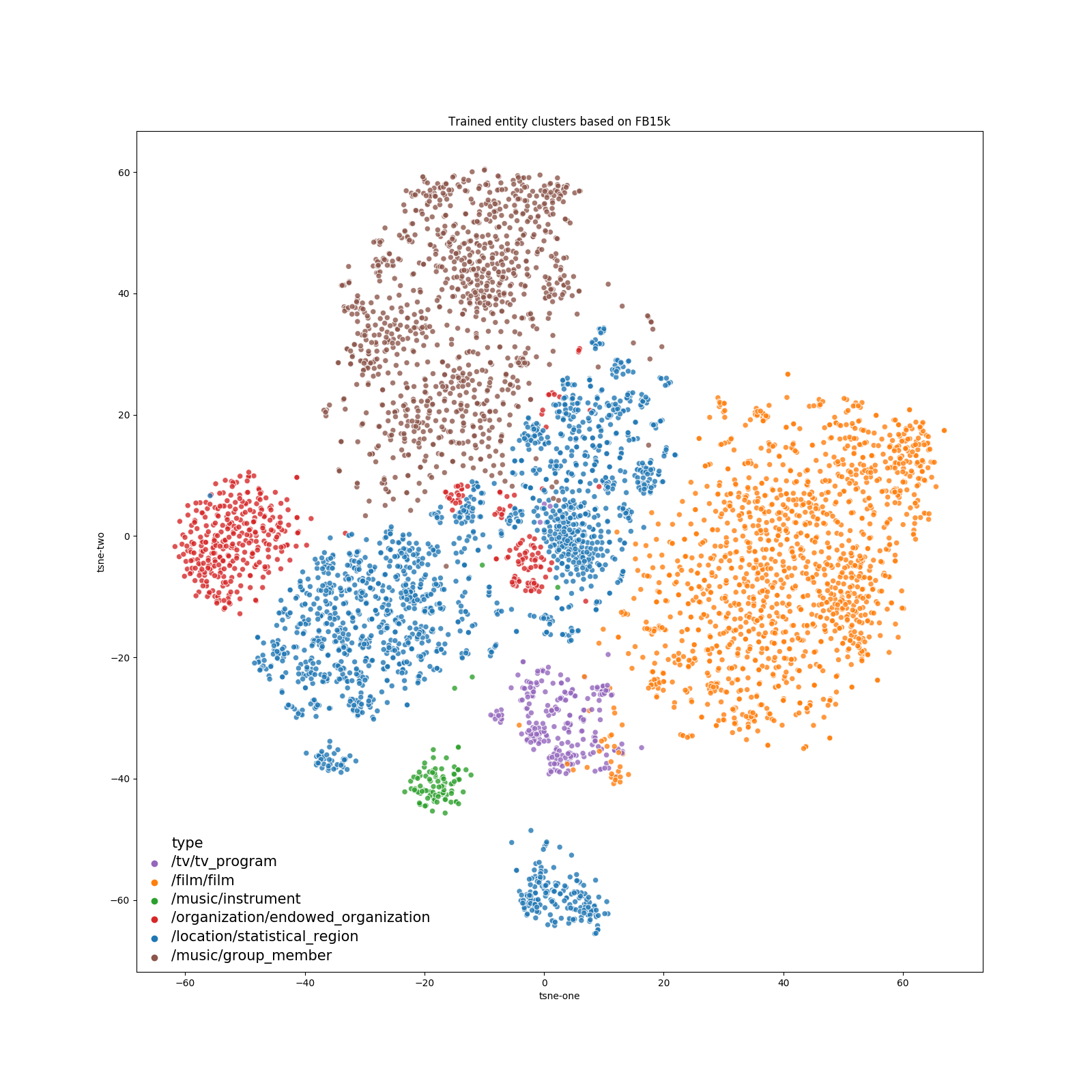}
		\caption{{\scriptsize Clustering by ComplEx}}
		\label{fig:clusteringcomplex}
	\end{subfigure}%
	\begin{subfigure}[b]{0.32\textwidth}
		\includegraphics[width=\linewidth]{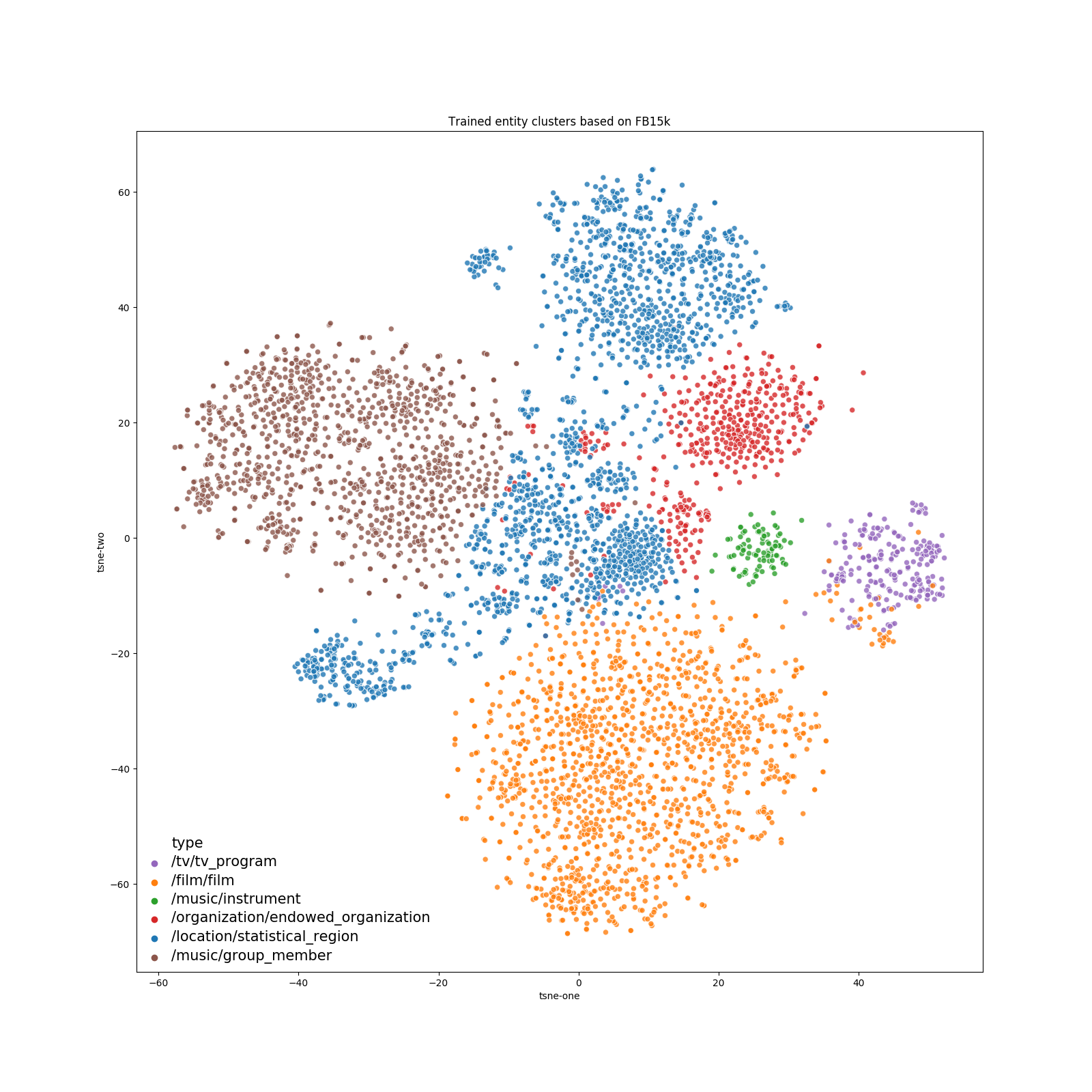}
		\caption{{\scriptsize Clustering by QuatE}}
		\label{fig:clusteringmobius}
	\end{subfigure}%
		\begin{subfigure}[b]{0.32\textwidth}
		\includegraphics[width=\linewidth]{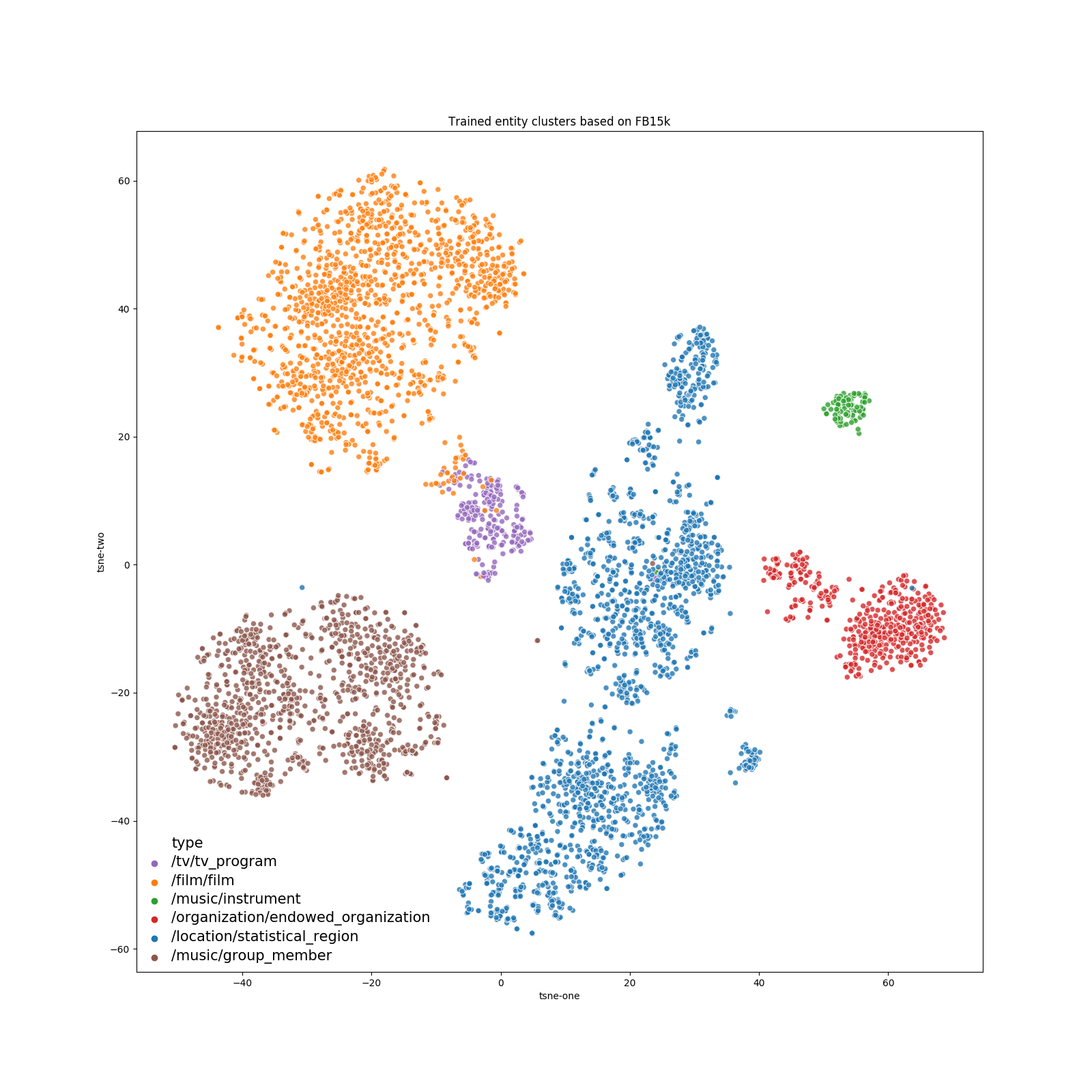}
		\caption{{\scriptsize Clustering by 5$\bigstar$E}}
		\label{fig:clusteringquate}
	\end{subfigure}%
	\caption{Comparison of clustering results between ComplEx, QuatE, and 5$\bigstar$E (left to right). Each color shows a class of entities and each point is one instance entity of the corresponding class.}
	\label{fig:clust}
\end{figure*}

\textbf{Learned Transformation Types}
Each relation in the KG is represented as $d$ projective transformations in 5$^\bigstar$E (one projective transformation per dimension). 
Figure~\ref{fig:grid} shows the transformation types learned by 5$^\bigstar$E for the relations of WordNet.
The original plain view of the grid is given in sub-graph (a) for comparisons of the changes after the transformations, and (b) to (e) show specific relations in WordNet.
The mapping of the \emph{lines} (same-color points) in the original grid to \emph{circle or curve} in sub-graph (b)-(e) indicates an application of an inversion transformation for relation-specific transformations (\emph{hasPart},
\emph{partOf}, \emph{hypernym} and \emph{hyponym}).
By comparing the direction of the lines with the same color (e.g., red) in the original grid and in all examples of the transformed grids, we can observe that the learned transformations cover rotation (\emph{hypernym}, and \emph{hyponym}). 
We can also interpret the results for the \emph{hasPart} relation as counter-clockwise rotation and then reflection w.r.t.~the real axis.
In sub-graph (b), there is a movement in the real and imaginary axis of the grid towards down and slightly right for the \emph{hasPart} relation, which represents translation. 
However, this is not the case for the \emph{hypernym} relation.
Semantically, the pairs (\emph{hypernym}, \emph{hyponym}) \color{black} and (\emph{hasPart} \emph{partOf}) form inverse patterns (see Corollary~\ref{prop:col13}). 
We see that the transformed grids of \emph{hypernym} and \emph{hyponym} are different only w.r.t.~rotation. 
The scale is not changed, so the determinants of the two projective matrices are 1 (no homothety). 
For the \emph{hasPart} and \emph{partOf} grids, we can observe that the scale is changed, so the determinant of those two projection matrices should not be equal to one. This shows that both of those transformations cover \emph{homothety}.

Moreover, each of the five transformation functions performed in Figure~\ref{fig:tranfun} are also learned by 5$^\bigstar$E which confirms the flexibility of the model as well as diversity in density/sparsity of flows.
Figure~\ref{fig:sameD} shows the analysis on the example of \textit{hyponym} and \textit{hypernym} as well as \textit{hasPart} and \textit{partOf} relations which are mutually inverse of each other. 
Based on their inverse characteristic, we have 
$\Im_{\textit{Hypernym}} = \bar{\Im}^{-1}_{\textit{Hyponym}}.$ 
As the representing matrices ($\Im$) are normalized, their determinant is equal to one. 
Consequently, we have $tr(\Im_{\textit{Hypernym}}) = tr(\bar{\Im}^{-1}_{\textit{Hyponym}})$ when the learned transformation function is elliptic. 
We conclude that  in our experiments for a pair of inverse relations, 
the learned transformation functions are in the same category (elliptic) for the $i$-th element of the relation embedding. 
In Figure~\ref{fig:sameD}, sub-figures (a) and (b) illustrate that the learned functions fall into the elliptic category for the same embedding dimension ($i = 12$) of \textit{hyponym} and \textit{hypernym}.  
The difference between the embeddings for this pair of inverse relations is their rotation. 
The same pattern is notable for the \textit{hasPart} and \textit{partOf} relations in sub-figures (c) and (d) for $i=39$. 
Figure~\ref{fig:llcc} shows the mapping of lines and circles by other KGEs. 
When observing each dimension of each relation, there was not a single case where a shape has been mapped to a different one, which empirically confirms our theoretical finding that existing models can only perform homogeneous transformations. 
In contrast, Figure~\ref{fig:lccl} shows a relation-specific mapping of line to circle and circle to line performed by 5$^\bigstar$E model. 

\medskip
\noindent
\textbf{Entity Clustering} As mentioned in theoretical analysis, our model uses a bijective conformal mapping in the projective geometry which consequently preserves angle locally.
In Figure~\ref{fig:clust}, we provide an evaluation for the performance of the models in terms of clustering.  
More precisely, Figures~\ref{fig:clusteringcomplex}, \ref{fig:clusteringmobius} and \ref{fig:clusteringquate} show the clustering of nodes in Freebase KG \cite{moon2017learning} and illustrate comparisons to QuatE, ComplEx and 5$\bigstar$E. 
In this visualization, we can see that entities of the same type are closer in 5$\bigstar$E as compared to the competitors. Moreover, the distance between cluster centers in 5$\bigstar$E is higher than in the other models. Therefore, it is visible that 5$\bigstar$E provides a more suitable clustering for this dataset than other competitors.

\section{Conclusion}

In this paper, we introduce a new KGE model which operates on the complete set of projective transformations. 
We build the model on well researched generic mathematical foundations and showed that it subsumes other state-of-the-art embedding models.
Furthermore, we prove that the model is fully expressive. 
By supporting a wider range of transformations than previous models, it can embed KGs with more complex structures and supports a wide range of relational patterns. 
We empirically studied and visualised the effects using the example of loop and path combinations. 
Our experimental evaluation on six benchmark datasets using established metrics shows that the model outperforms previous approaches of knowledge graph embedding models.

\section{Acknowledgements}
We acknowledge the support of the following projects: SPEAKER (BMWi FKZ 01MK20011A), JOSEPH (Fraunhofer Zukunftsstiftung), Cleopatra (GA 812997), the excellence clusters ML2R (BmBF FKZ 01 15 18038 A/B/C), MLwin (01IS18050), ScaDS.AI ( IS18026A-F), TAILOR (EU GA 952215), and H2020-EU PLATOON (872592).


\bibliography{main.bbl}

\newpage
\section{Appendix}

\subsection{Capturing Path/loop}

\begin{figure*}[ht!]
	\centering 
	\includegraphics[width=\textwidth]{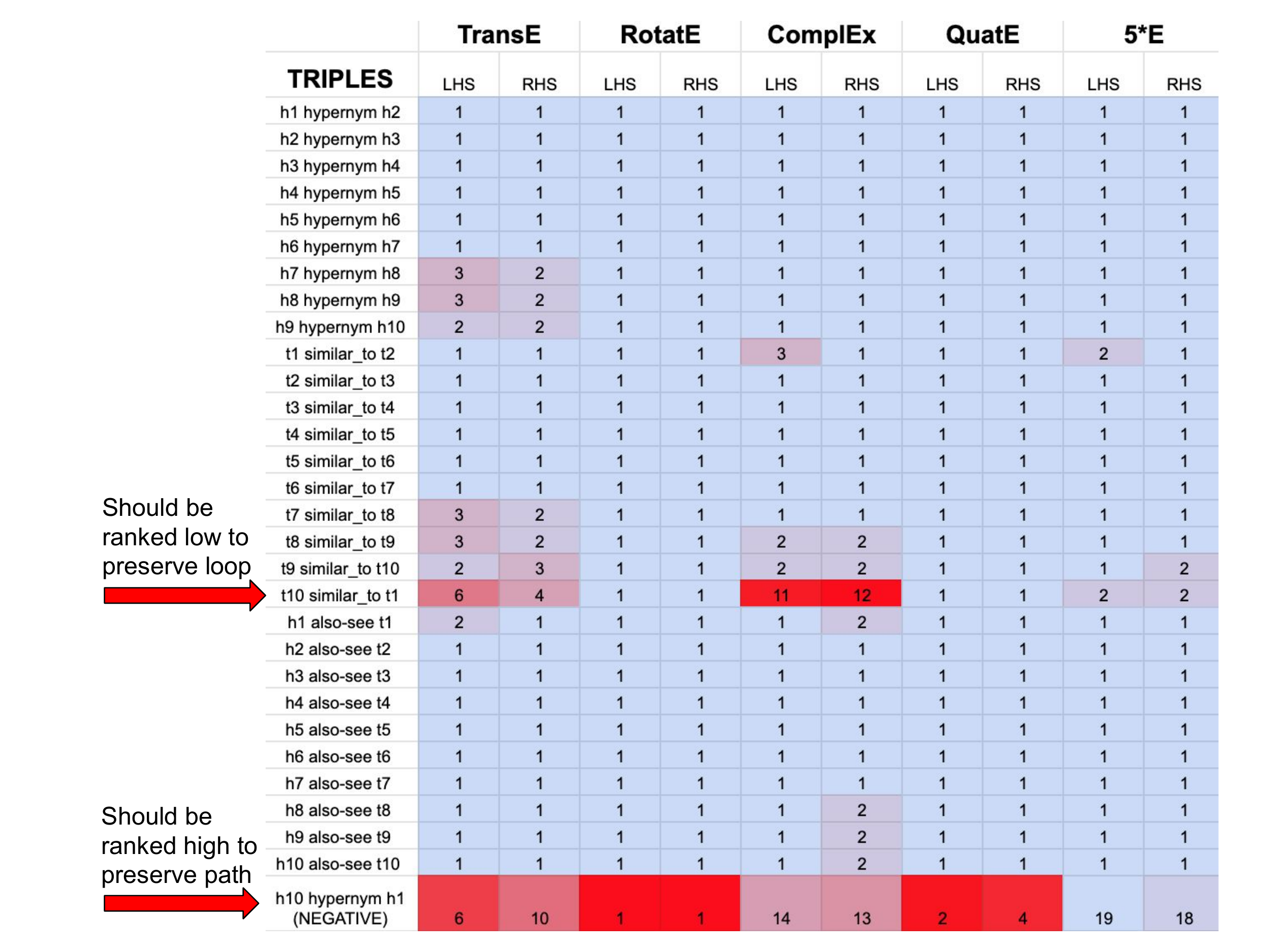}
	\caption{Rankings of sample triples in loop ($t_i$ where $i \dots 10 $) and path ($h_i$ where $i \dots 10 $)) shapes by different models.}
	\label{fig:heatmap}
\end{figure*}

Here, we demonstrate an example for the cases presented in Figure 1 in the main body of the paper. 
We choose path/loop relation to build up our example.
Assume that, we have a set of 10 different nodes which are positioned as a path ($h_i$ where $i \dots 10 $) of a specific relation (e.g. hypernym).
A different set of another 10 nodes ($t_i$ where $i \dots 10 $) are connected through a relation (e.g. similar\_to) in a loop structure. 
Our assumption is that the path structure is positioned beside a loop.
The nodes in the head position are collected (e.g. through the also-see relation) to the ones in the loop position.
In Figure \ref{fig:heatmap}, the given ranking by each embedding model for each triple is shown.
Low number means a better ranking for higher plausibility of each triple, and high ranking means the chance of triple incorrectness is high. 
Column one shows all the possible triples which are involved in these three relations. 
We consider bi-directional ranking as ``right hand side ranking''(RHS) and ``left hand side ranking''(LHS).

The highlights to be noted are that:

\begin{itemize}
    \item In order to preserve the shape of the loop, the triple $(t_{10}, \, similar-to, \, t_1)$ should be ranked as low as possible. 
    The problem occurs when a model gives a high rank to this triple. 
    If this happens, the shape of the loop is transferred (wrongly) to a path. 
    \item In order to preserve the shape of the path, the $(h_{10}, \, similar-to, \, h_1)$ triple should be ranked as high as possible. 
    The problem occurs when a model gives a low rank to this triple. 
    If this happens, the shape of the path is transferred (wrongly) to a loop. 
\end{itemize}

\begin{figure*}[ht!]
	\centering 
	\begin{subfigure}[b]{0.19\textwidth}
		\includegraphics[width=\linewidth]{circle-flow.png}
		\label{fig:fb:noInj}
	\end{subfigure}%
	\begin{subfigure}[b]{0.19\textwidth}
		\includegraphics[width=\linewidth]{elliptic-flow.png}
		\label{fig:4c:eq}
	\end{subfigure}%
	\begin{subfigure}[b]{0.19\textwidth}
		\includegraphics[width=\linewidth]{hyperbolic-flow.png}
		\label{fig:4d:sym}
	\end{subfigure}%
	\begin{subfigure}[b]{.19\textwidth}
		\includegraphics[width=\linewidth]{loxodromic-flow.png}
		\label{fig:4a:noInj}%
	\end{subfigure}%
	\begin{subfigure}[b]{.19\textwidth}
		\includegraphics[width=\linewidth]{parabolic-flow.png}
		\label{fig:4a:noInj}%
	\end{subfigure}%
\\\vspace{-5mm}
	\begin{subfigure}[b]{0.19\textwidth}
		\includegraphics[width=\linewidth]{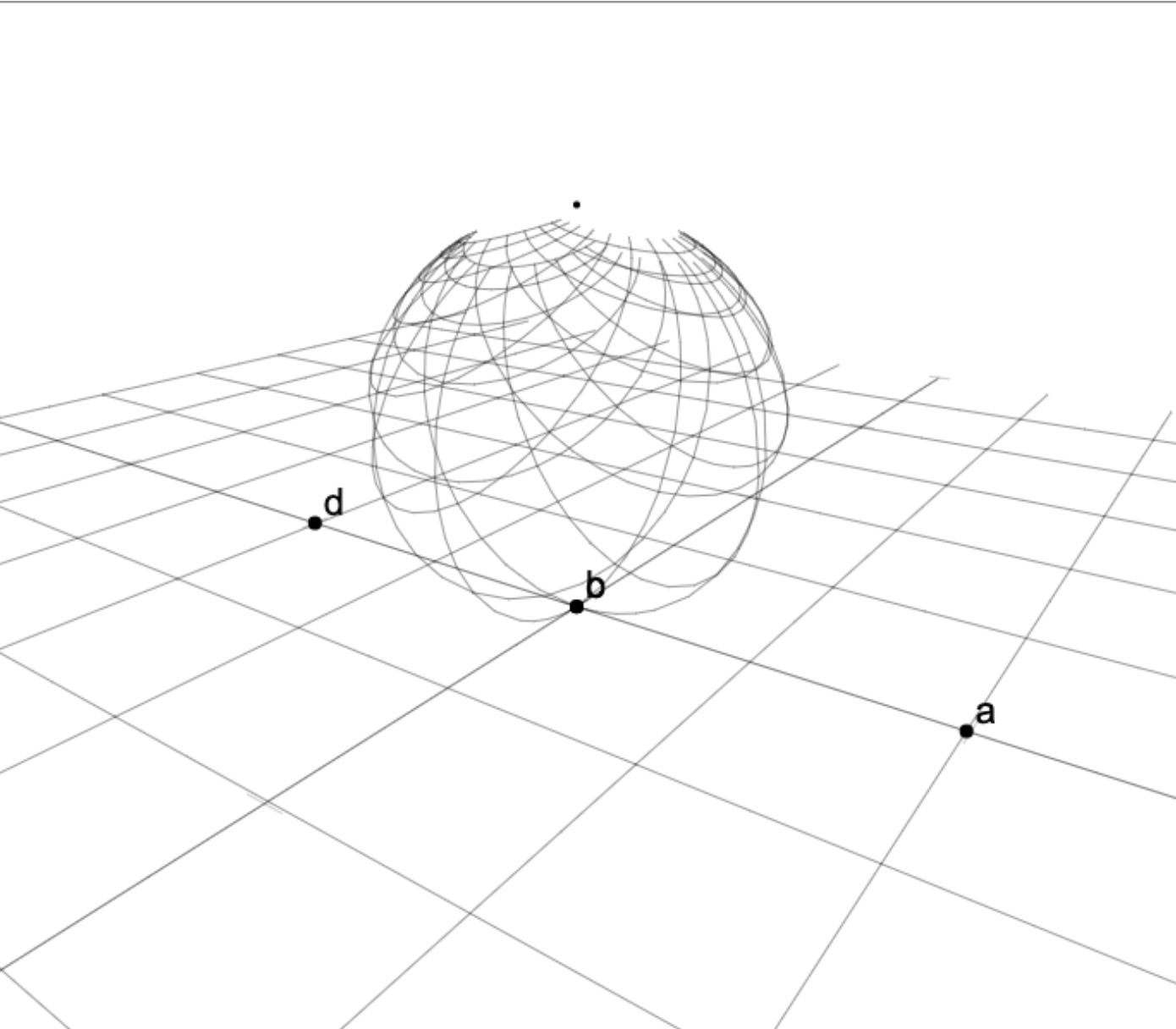}
		\caption{{\scriptsize circular}}
		\label{fig:fb:noInj}
	\end{subfigure}%
	\begin{subfigure}[b]{0.19\textwidth}
		\includegraphics[width=\linewidth]{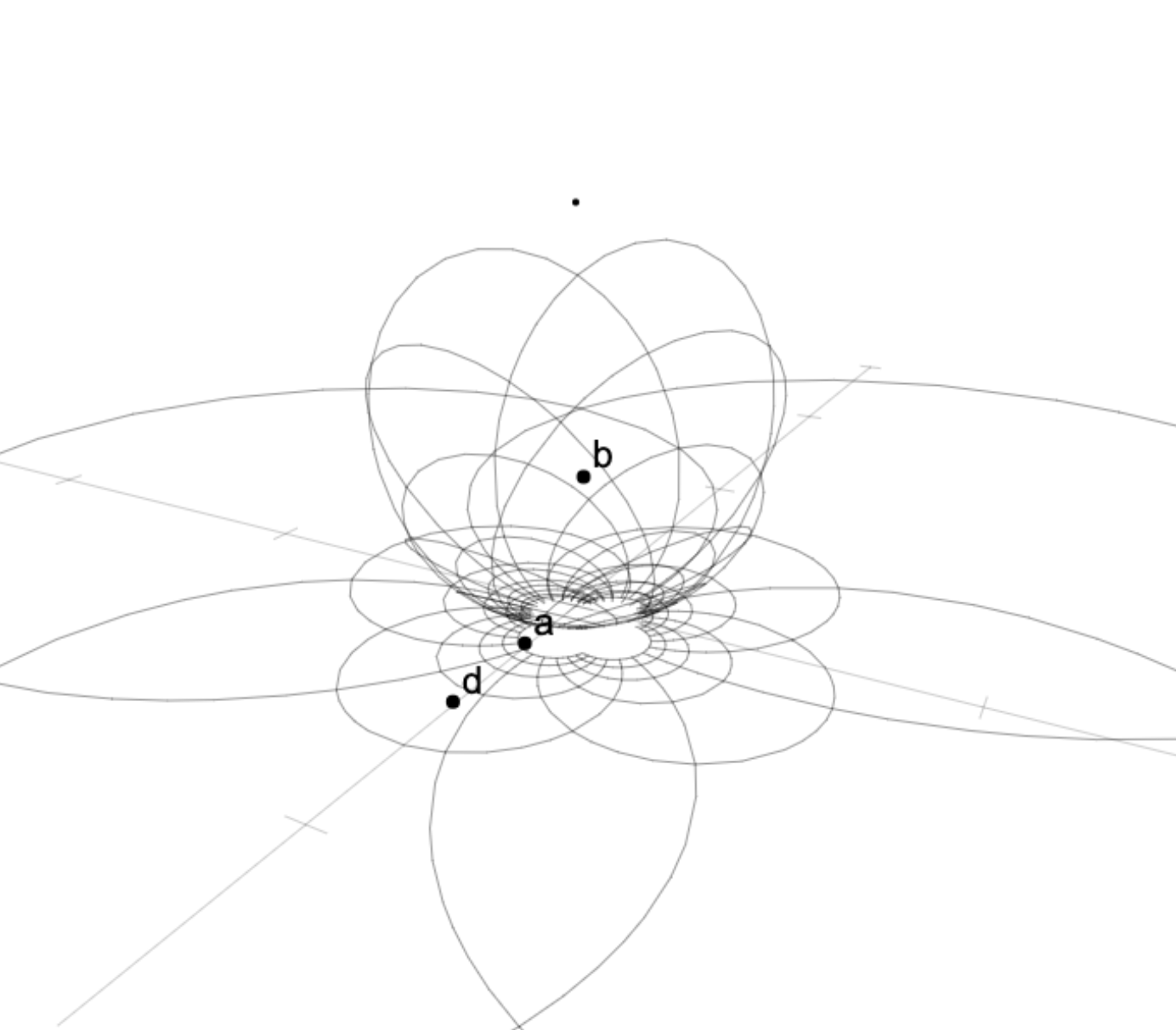}
		\caption{{\scriptsize elliptic}}
		\label{fig:4c:eq}
	\end{subfigure}%
	\begin{subfigure}[b]{0.19\textwidth}
		\includegraphics[width=\linewidth]{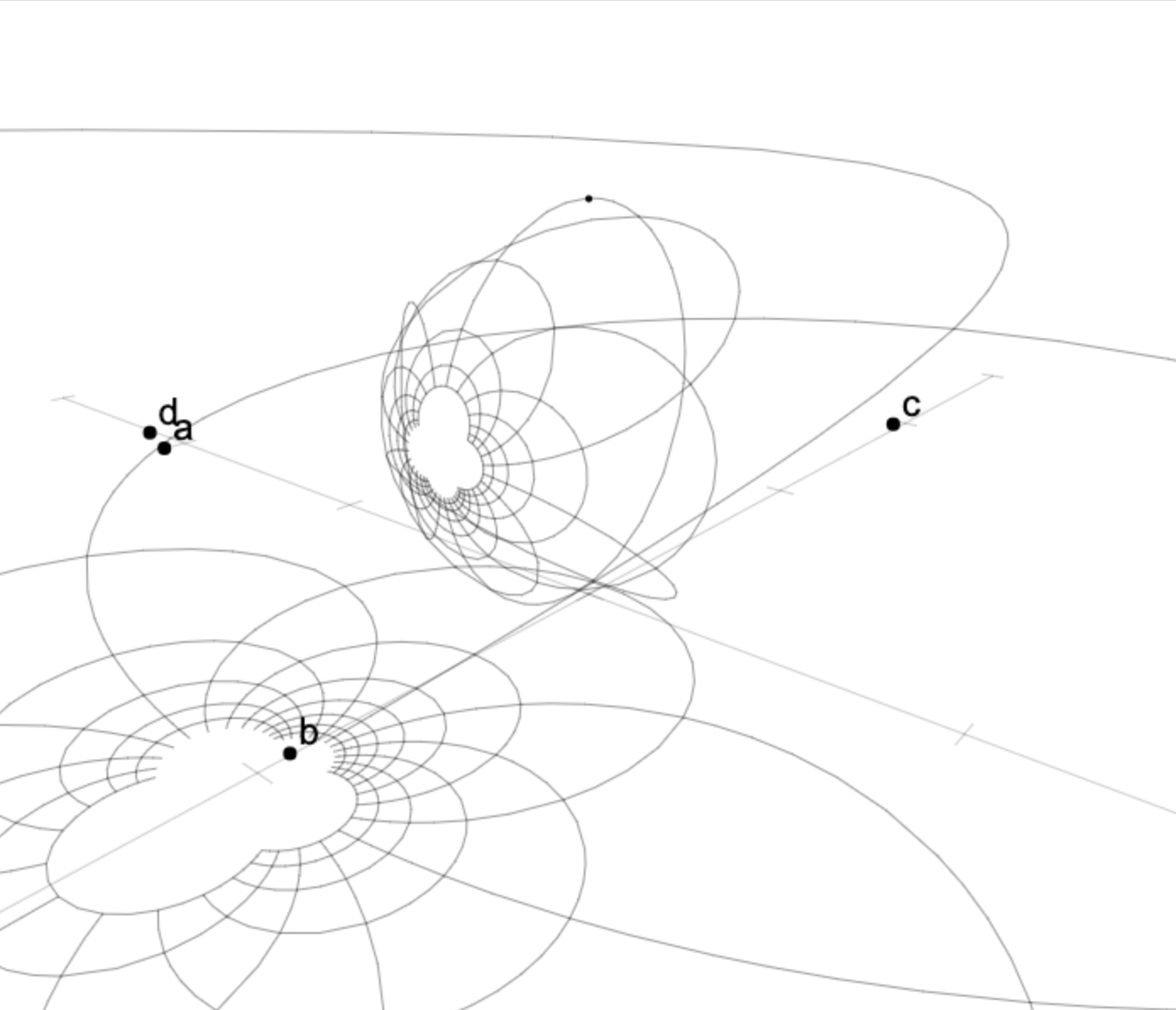}
		\caption{{\scriptsize hyperbolic}}
		\label{fig:4d:sym}
	\end{subfigure}%
	\begin{subfigure}[b]{.19\textwidth}
		\includegraphics[width=\linewidth]{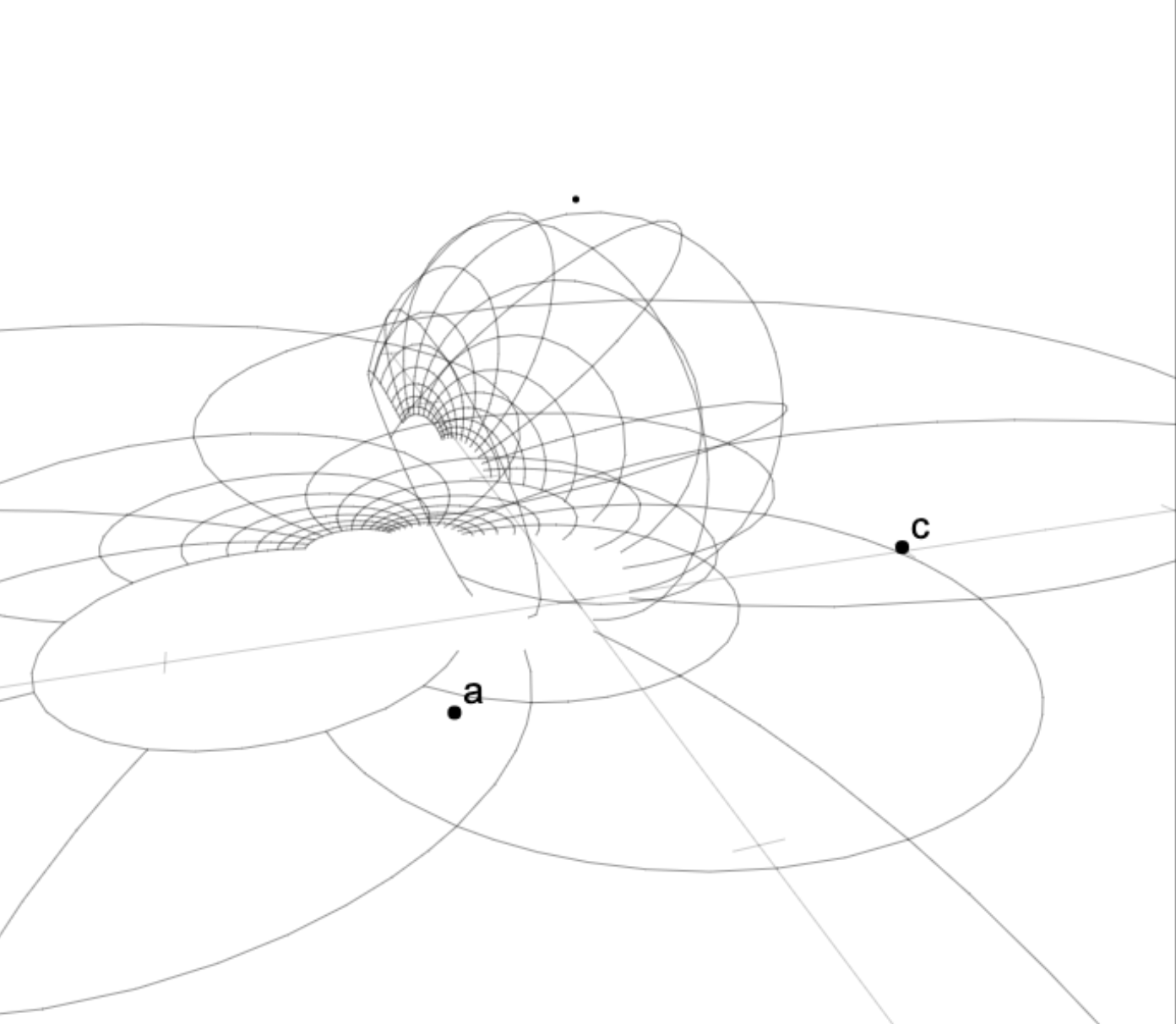}
		\caption{\scriptsize {loxodromic}}
		\label{fig:4a:noInj}%
	\end{subfigure}%
	\begin{subfigure}[b]{.18\textwidth}
		\includegraphics[width=\linewidth]{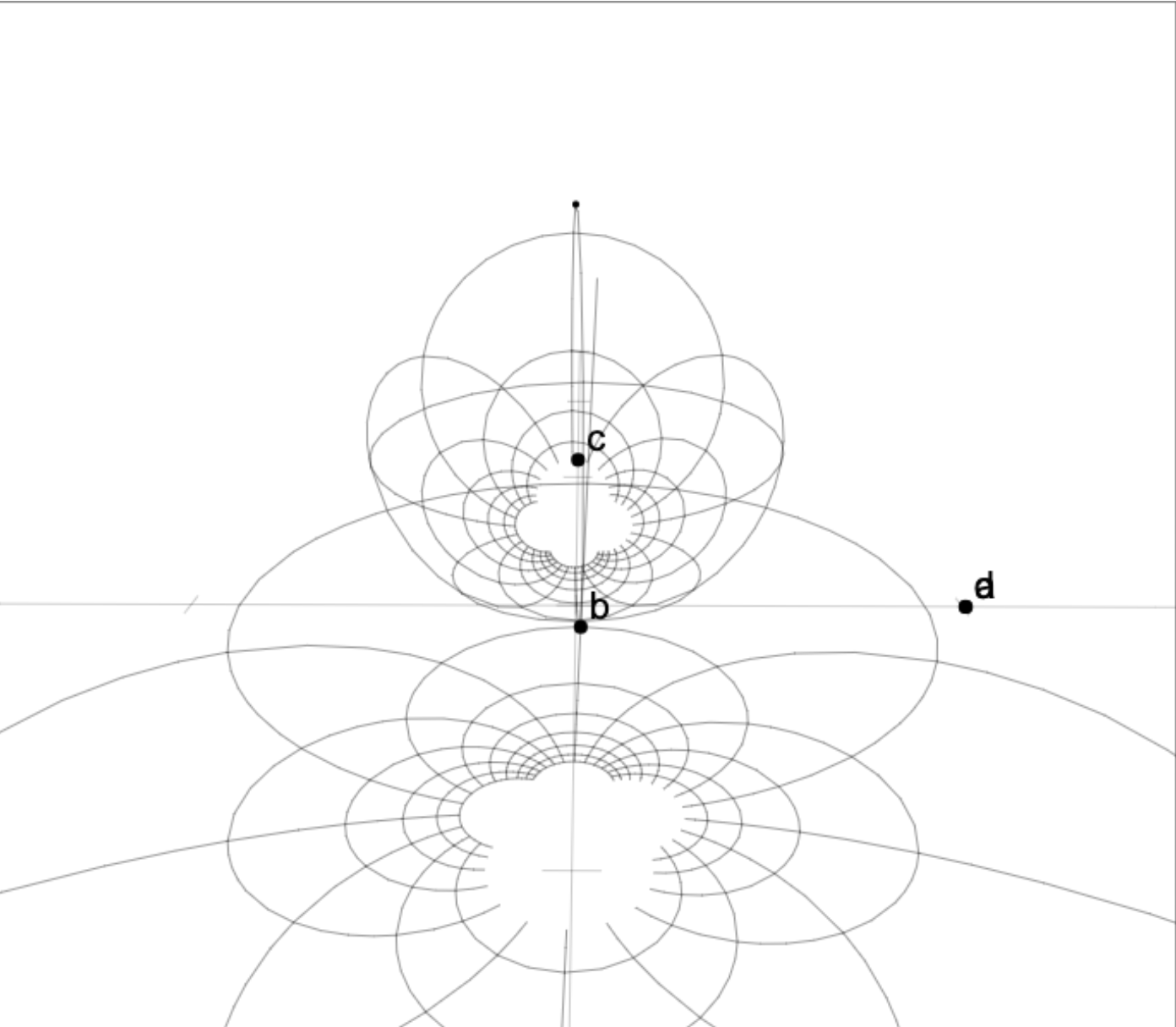}
		\caption{\scriptsize {parabolic}}
		\label{fig:4a:noInj}%
	\end{subfigure}%
	\caption{Default transformation functions with Riemann Sphere (first row) and the M\"obius shape for each transformation after projection on a complex plane (second row) are shown.}
	\label{fig:tranfun}
\end{figure*}

A perfect model should specifically learn $(t_{10}, \, similar-to, \, t_1)$ as positive (rank low) in order to preserve the loop. 
Moreover, the model should learn $(h_{10}, \, similar-to, \, h_1)$ as a negative triple with a high rank which means the path shape is preserved. 

For $(t_{10}, \, similar-to, \, t_1)$ triple which is expected to be ranked very low by a good KGE model, $5^\bigstar$E ranks this triple as low as possible (2 both for LHS and RHS), without problematically affecting the ranking of $(h_{10}, \, similar-to, \, h_1)$ triple  (ranked high as 19, 18 ).  
This concludes that the shape of loop and path are preserved by $5^\bigstar$E.
However, other models completely miss-preserve the head triple of $(h_{10}, \, similar-to, \, h_1)$ when they take a low rank for $(t_{10}, \, similar-to, \, t_1)$. 
This is more visible for QuatE and RotatE. 
RotatE learns $(t_{10}, \, similar-to, \, t_1)$ with lowest rank (1), however it also ranks $(h_{10}, \, similar-to, \, h_1)$ with 1. 
Relatively similar behaviours is observed from QuatE. 
This concludes that the shape of loop is preserved by RotatE and QuatE, however it was with the cost of (wrongly) transferring path to loop as well.
For TransE, a relatively high ranking is assigned to $(h_{10}, \, similar-to, \, h_1)$ triple (which means the model expects this triple to be incorrect) and preserves the path, however this causes a high rank for $(t_{10}, \, similar-to, \, t_1)$ as well.
We see in that by using the ComplEx model, it shows a similar behaviour to TransE model. 
This concludes that the shape of path is preserved, however the loop part is also transferred to a path shape. 
This observation in connection to Figure 1 clarifies the difficulty of other KGE models when different motifs come close in subgraphs of a KG. 
Our proposed model has the potential to correctly preserve each structure separately. 
This power comes from the possible transformations covered in projective geometry, illustrated in Figure \ref{fig:tranfun}. 
In Table \ref{table:transTypes}, the theoretical description and the validity conditions of each transformation is discussed.
All transformations in each group form a subgroup which is isomorphic to the group of all matrices mentioned in the row \textit{Iso}. Note that $tr$ in the table denotes the trance of a matrix.

\subsection{Vector Representation of Path/Loop by RotatE}
As mentioned in the main paper, the RotatE model is not able to preserve a loop structure connected with a path from the graph representation in the vector space. To show this, without loss of generality, we investigate the example given in Figure one with a concrete scenario. 
Let us consider three entities ($h_1, h_2 , h_3$) connected by the relation $r_1$ in a path structure i.e.~$(h_1,r_1,h_2), (h_2,r_1,h_3)$, and three other entities $(t_1, t_2, t_3)$ connected by relation $r_3$ in a loop structure i.e.~$(t_1,r_1,t_2), (t_2,r_1,t_3), (t_3,r_1,t_1)$. 
RotatE encodes path as

\begin{equation}
    \begin{split}
        \mathbf{h}_1 \circ \mathbf{r_1} \approx \mathbf{h}_2,\\
        \mathbf{h}_2 \circ \mathbf{r_1} \approx \mathbf{h}_3, \\
        \mathbf{h}_3 \circ \mathbf{r_1} \neq \mathbf{h}_1,
    \end{split}
    \label{rotatepath}
\end{equation}
 and loop as
 
\begin{equation}
    \begin{split}
        \mathbf{t}_1 \circ \mathbf{r_3} \approx \mathbf{t}_2,\\
        \mathbf{t}_2 \circ \mathbf{r_3} \approx \mathbf{t}_3, \\
        \mathbf{t}_3 \circ \mathbf{r_3} \approx \mathbf{h}_1.
    \end{split}
    \label{rotateloop}
\end{equation}

Additionally, entities of the path structure are connected to the entities of the loop structure i.e.~$(h_1,r_2,t_1), (h_2,r_2,t_2), (h_3,r_2,t_3)$, which are encoded by RotatE as

 \begin{equation}
    \begin{split}
        \mathbf{h}_1 \circ \mathbf{r_2} \approx \mathbf{t}_1,\\
        \mathbf{h}_2 \circ \mathbf{r_2} \approx \mathbf{t}_2, \\
        \mathbf{h}_3 \circ \mathbf{r_2} \approx \mathbf{t}_3.
    \end{split}
    \label{rotatepathloop}
\end{equation}

Note that the representation of relations between entities is actually a rotation in complex space  when using RotatE model i.e.~$\mathbf{r} = e^{i\theta}$.

After combining Equation \ref{rotatepath} and \ref{rotateloop}, we have

\begin{equation}
    \begin{split}
        \mathbf{h}_1 \circ \mathbf{r_2} \circ \mathbf{r_3} \approx \mathbf{t}_2,\\
        \mathbf{h}_2 \circ \mathbf{r_2} \circ \mathbf{r_3} \approx \mathbf{t}_3, \\
        \mathbf{h}_3 \circ \mathbf{r_2} \circ \mathbf{r_3} \approx \mathbf{t}_1.
    \end{split}
    \label{combinedrotatepathloop}
\end{equation}

After replacing $\mathbf{t}_2$ by $\mathbf{h}_2 \circ \mathbf{r}_2$ (according to equation \ref{rotatepathloop}) in equation \ref{combinedrotatepathloop}, we have

\begin{equation}
    \begin{split}
        \mathbf{h}_1 \circ \mathbf{r_3} \approx \mathbf{h}_2,\\
        \mathbf{h}_2 \circ \mathbf{r_3} \approx \mathbf{h}_3, \\
        \mathbf{h}_3\circ \mathbf{r_3} \approx \mathbf{h}_1.
    \end{split}
    \label{rotatefinal}
\end{equation}

Comparing equation \ref{rotatefinal} and \ref{rotatepath}, we have $\mathbf{r}_1 = \mathbf{r}_3.$ Therefore, $\mathbf{h}_3\circ \mathbf{r_3} \approx \mathbf{h}_1$ and $\mathbf{h}_3\circ \mathbf{r_1} \neq \mathbf{h}_1$ should be held simultaneously which is in contradiction with $\mathbf{r}_1 = \mathbf{r}_3.$ Based on the above-mentioned results, RotatE cannot encode loop connected with path in the vector space and either loop becomes path or path becomes loop. 
Similarly, we prove that other models such as QuatE (rotation in Quaternion space) and ComplEx (rotation in complex space together with homothety) also face issues when transforming a path beside a loop structure. 
In Table \ref{table:trans}, we show the capability of KGEs in preserving different transformations.  
Correctly learning of one of these two triples, scarifies the other.

\begin{table}[h!]
\centering
\begin{tabular}{lccccc}
\hline
            & Tra. & Rot. & Hom. & Inv. & Ref. \\ 
TransE           &  $\bigstar$     &   $\openbigstar$  & $\openbigstar$ &  $\openbigstar$ &  $\openbigstar$\\ 
RotatE            &   $\openbigstar$    & $\bigstar$ &  $\openbigstar$& $\openbigstar$ & $\openbigstar$ \\ 
ComplEx                &$\openbigstar$  & $\bigstar$  & $\bigstar$   & $\openbigstar$ & $\openbigstar$\\ 
QuatE            &   $\openbigstar$   & $\bigstar$  &  $\openbigstar$ &  $\openbigstar$   & $\openbigstar$ \\ 
$5^\bigstar$E            &      $\bigstar$  & $\bigstar$  &  $\bigstar$ &   $\bigstar$ &   $\bigstar$    \\ \hline
\end{tabular}
\caption{Supported Transformations by KGE Models.}
\label{table:trans}
\end{table}


\begin{figure}[ht!]
	\centering 
	\includegraphics[trim = 0cm 2cm 0cm 2cm, width=\linewidth]{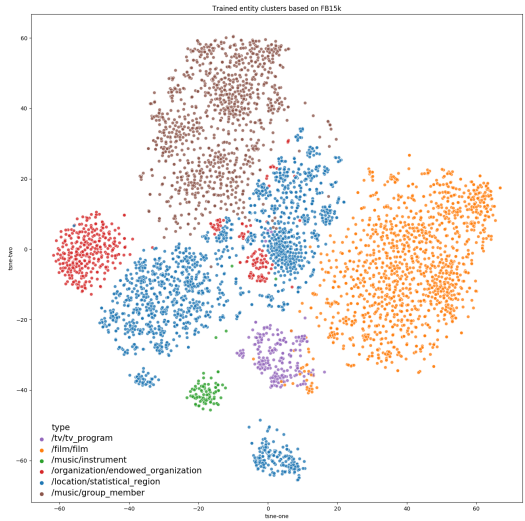}
	\caption{Clustering by ComplEx}
	\label{fig:clusteringcomplex}
\end{figure}

\begin{figure}[ht!]
	\centering 
	\includegraphics[trim = 0cm 2cm 0cm 2cm,width=\linewidth]{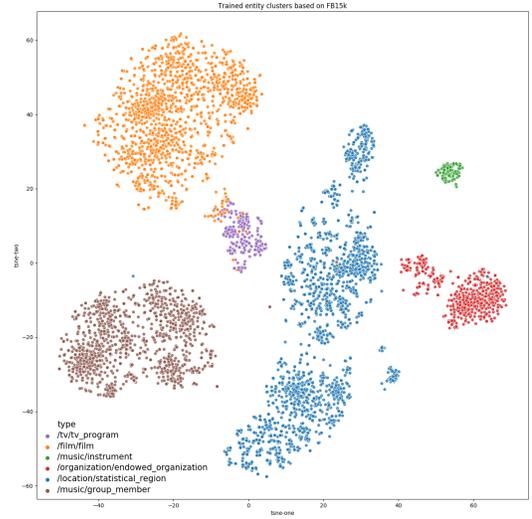}
	\caption{Clustering by 5$\bigstar$E}
	\label{fig:clusteringmobius}
\end{figure}

\begin{figure}[ht!]
	\centering 
	\includegraphics[trim = 0cm 2cm 0cm 2cm,width=\linewidth]{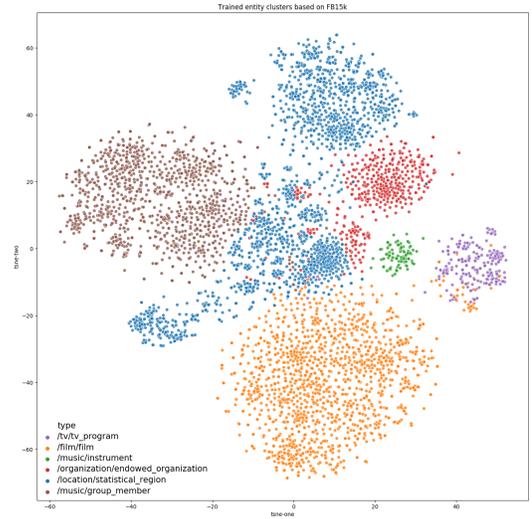}
	\caption{Clustering }
	\label{fig:clusteringquate}
\end{figure}

\begin{table*}[t!]
\centering
\begin{tabular}{llllll}
\toprule 
 Function   & Parabolic   & Circular & Elliptic &  Hyperbolic   & Loxodromic \\ \midrule 
Condition     &   \pbox{15cm}{$tr \Im^2 = 4 $  \\($\Delta = 0$)}     &  \pbox{15cm}{$tr\Im^2 = 0$ \\$(\Delta=0)$}  &  
 \pbox{15cm}{$0 < tr\Im^2 < 4$ \\$(\Delta<0)$} &  
 \pbox{15cm}{$tr\Im^2 > 4$ \\$(\Delta>0)$}  &  
 \pbox{15cm}{$tr\Im^2 \notin [0,4]$}\\ 
 \rule{0pt}{4ex} 
 Isomorphic   & $\begin{bmatrix}
    1&a\\
    0&1
    \end{bmatrix}$  & $\begin{bmatrix}
    i&0\\
    0&-i
    \end{bmatrix}$& $\begin{bmatrix}
    e^{i\theta/2}&0\\
    0&e^{-i\theta/2}
    \end{bmatrix}$  & $\begin{bmatrix}
    e^{\theta/2}&0\\
    0&e^{-\theta/2}
    \end{bmatrix}$  & $\begin{bmatrix}
    k&0\\
    0&\frac{1}{k}
    \end{bmatrix}$ \\ \bottomrule 
\end{tabular}
\caption{Types of M\"obius transformations and their conditions.}
\label{table:transTypes}
\end{table*}

\begin{table*}[ht!]
\caption{Link prediction results on WN18 and WN18RR as well as FB15k and FB15k-237.}
\begin{adjustbox}{width=.8\textwidth,center}
\begin{tabular}{lllllllll}
 \toprule 
   \multirow{1}{*}{Model} & \multicolumn{4}{c}{WN18}         & \multicolumn{4}{c}{FB15k}      \\ \cline{1-1} \cline{2-9} 
                          & MRR & Hits@1 & Hits@3 & Hits@10  & MRR & Hits@1 & Hits@3 & Hits@10 \\ \cline{2-9} 
                 TransE   &0.49& 0.11 & 0.89 & 0.94  &0.46& 0.30 & 0.58 & 0.75    \\ \cline{2-9} 
                 RotatE   &0.95 &0.94 &0.95 &\cellcolor{blue!8}0.96  &0.70& 0.58 & 0.79 & 0.87 \\   \cline{2-9}
                TuckEr    &0.95&0.95  &0.95 &0.96&0.79& 0.74 & 0.83 & 0.89 \\ \cline{2-9} 
                ComplEx   &0.94& 0.94 & 0.94 & 0.94   &0.69& 0.60 & 0.76 &  0.84   \\ \cline{2-9} 
                QuatE     &0.95&0.94 &0.95 & \cellcolor{blue!6}0.96&0.83 & \cellcolor{blue!12}0.80&  \cellcolor{blue!6}0.86 &  \cellcolor{blue!6}0.90  \\ \cline{2-9} 
                SimplE    &0.94& 0.94 & 0.94 & 0.95   &0.73& 0.66 &  0.77& 0.84      \\ \cline{2-9} 
                ConvE     &0.94& 0.93 & 0.95 & \cellcolor{blue!6}0.96  &0.66 & 0.56 &0.72&  0.83    \\ \cline{2-9}
\hline
                $5^\bigstar$E \scriptsize{d = 500} &  \cellcolor{blue!12}0.96&  \cellcolor{blue!12}0.96&  \cellcolor{blue!12}0.96&  \cellcolor{blue!6}0.96&  \cellcolor{blue!12}0.84 &0.78  &\cellcolor{blue!6}0.86  &\cellcolor{blue!6}0.90  \\ 
                $5^\bigstar$E \scriptsize{d = 100} & 0.95  & 0.95 & 0.95 & \cellcolor{blue!6}0.96& 0.73& 0.66&  0.78&  0.86  \\ 
\bottomrule 
\end{tabular}
\end{adjustbox}
\label{table:result_table_wn}
\end{table*}

\subsection{Entity Clustering}
As mentioned in the paper, our model uses bijective conformal mapping in the projective geometry which consequently preserves angle locally but not necessarily the length of motifs. Such characteristic is consistent with the nature of KGs where similarity of nodes in the graph is local i.e.~nodes within a close neighborhood are more likely to be semantically similar than nodes at a higher
distance. 
Figures \ref{fig:clusteringcomplex}, \ref{fig:clusteringmobius} and \ref{fig:clusteringquate} show the clustering results of each model of QuatE, ComplEx and 5$\bigstar$E. According to this visualizations, 5$\bigstar$E puts entities of the same type more closer (lower variance) than other competitors. Moreover, the distance between center of clusters in 5$\bigstar$E is bigger than other models. Therefore, it is visible that 5$\bigstar$E provides a more accurate clustering than other competitors.

\subsection{Experiments On FB15K and WN18}
In Table \ref{table:result_table_wn}, the results of experimenting 5$^\bigstar$E on FB15K and WN18 are compared to other models. 
In WN18, 5$^\bigstar$E outperforms all the models in all the metrics expect Hits@10 where 0.96 is achieved by RotatE, TuckEr, QuatE, and ConvE as well as 5$^\bigstar$E.
On the FB15k dataset, we observe that
5$^\bigstar$E outperforms TransE, RotatE, TuckEr, ComplEx, SimplE, and ConvE on FB15K. 
5$^\bigstar$E and QuatE achieve a very competitive results in most of the cases.
In MRR, our model performs slightly better, and in Hits@1, QuatE is better. 


\subsection{Learned Embedding: Entity Level}
\begin{figure*}[h!]
	\centering 
	\begin{subfigure}[b]{0.4\textwidth}
		\includegraphics[width=\linewidth,height=10 cm]{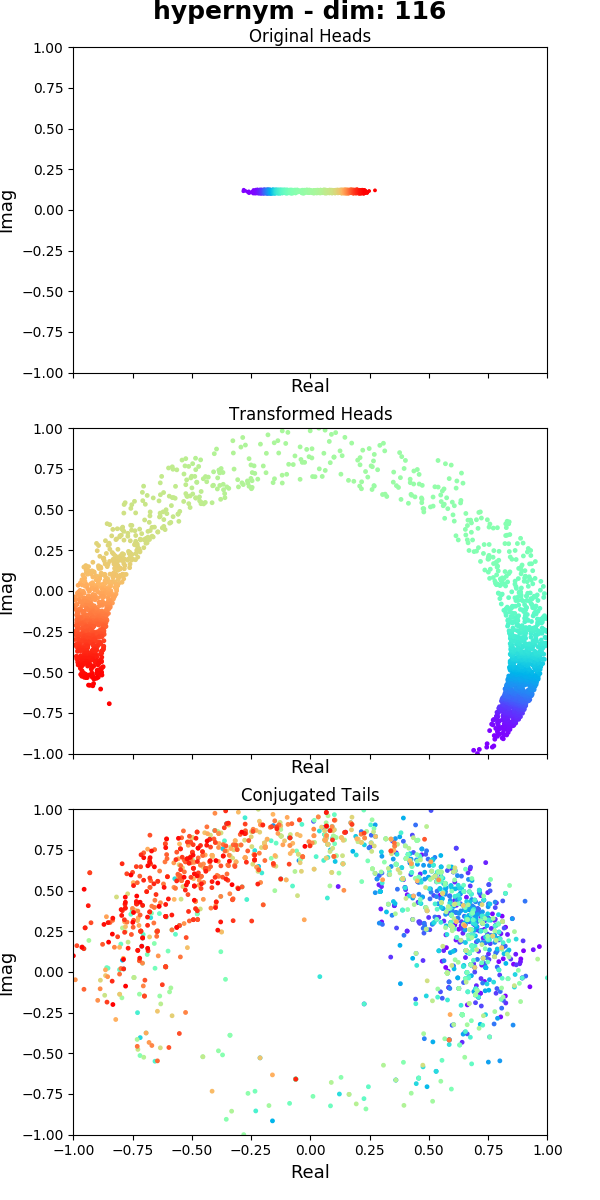}
		\caption{}
		\label{fig:3}
	\end{subfigure}\hspace{2mm}
	\begin{subfigure}[b]{0.4\textwidth}
		\includegraphics[width=\linewidth,height=10 cm]{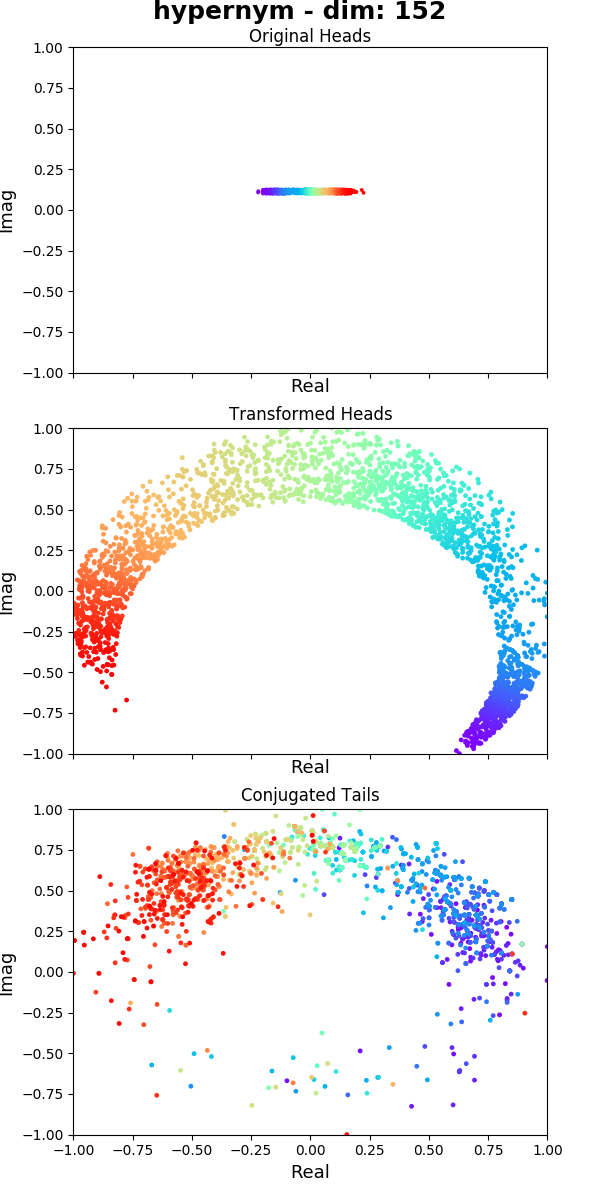}
		\caption{}
		\label{fig:4}
	\end{subfigure}

	\begin{subfigure}[b]{0.4\textwidth}
		\includegraphics[width=\linewidth,height=10 cm]{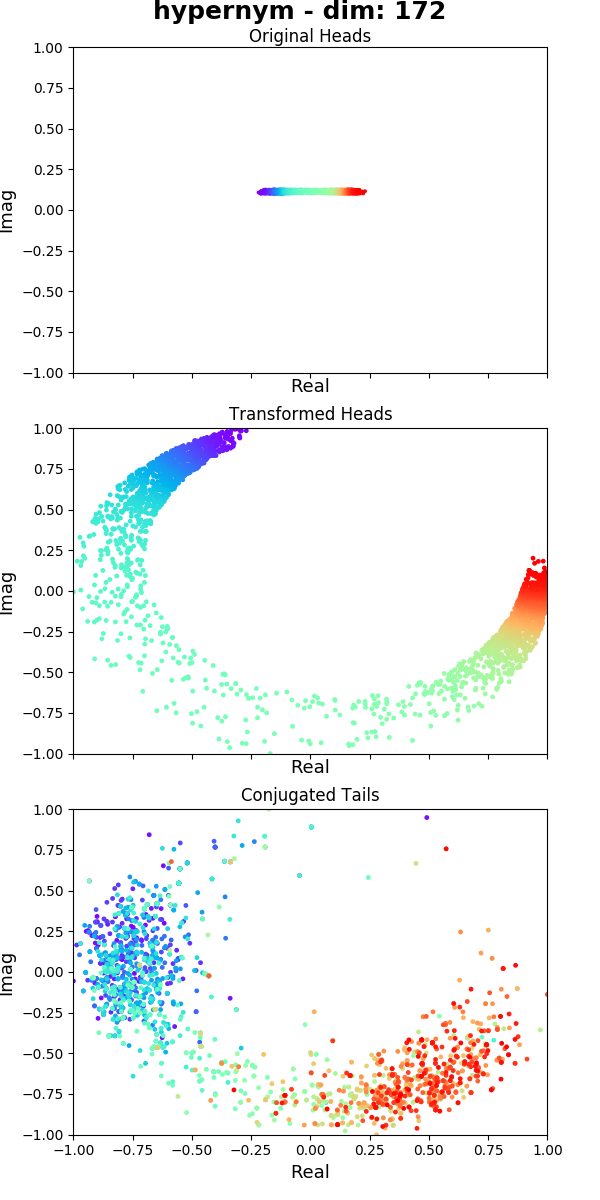}
			\caption{}
		\label{fig:5}
	\end{subfigure}\hspace{2mm}
	\begin{subfigure}[b]{0.4\textwidth}
		\includegraphics[width=\linewidth,height=10 cm]{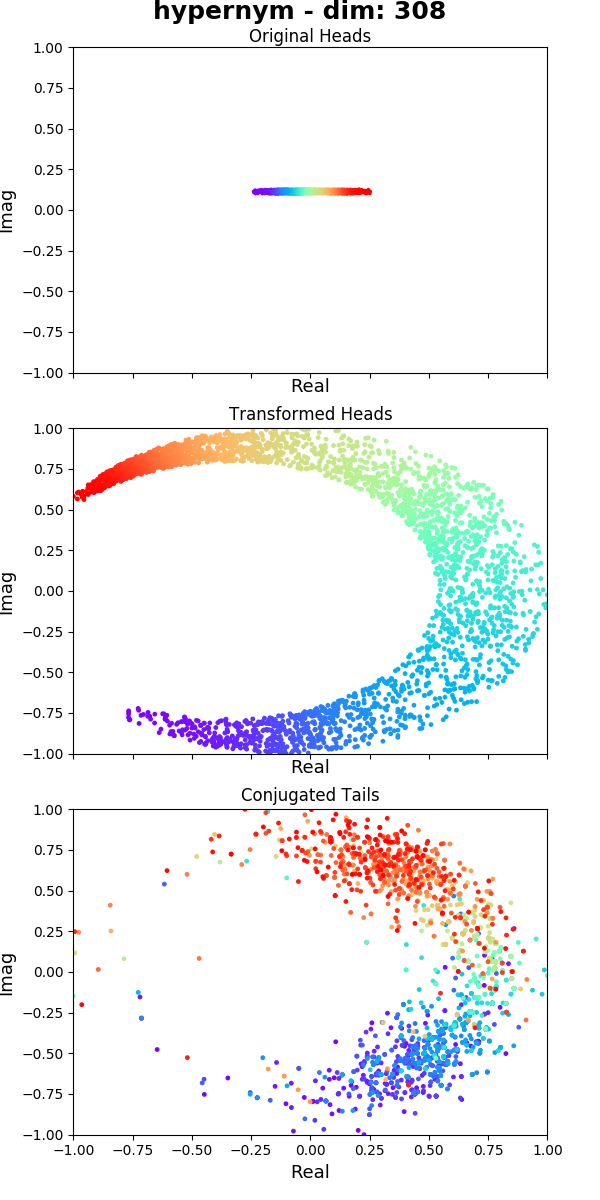}
		\caption{}
		\label{fig:6}
	\end{subfigure}
	\caption{Learned embeddings of entities connected by the hypernym relation.}
	\label{fig:spec2}
\end{figure*}

\begin{figure*}[h!]
	\centering 
	\begin{subfigure}[b]{0.4\textwidth}
		\includegraphics[width=\linewidth,height=10 cm]{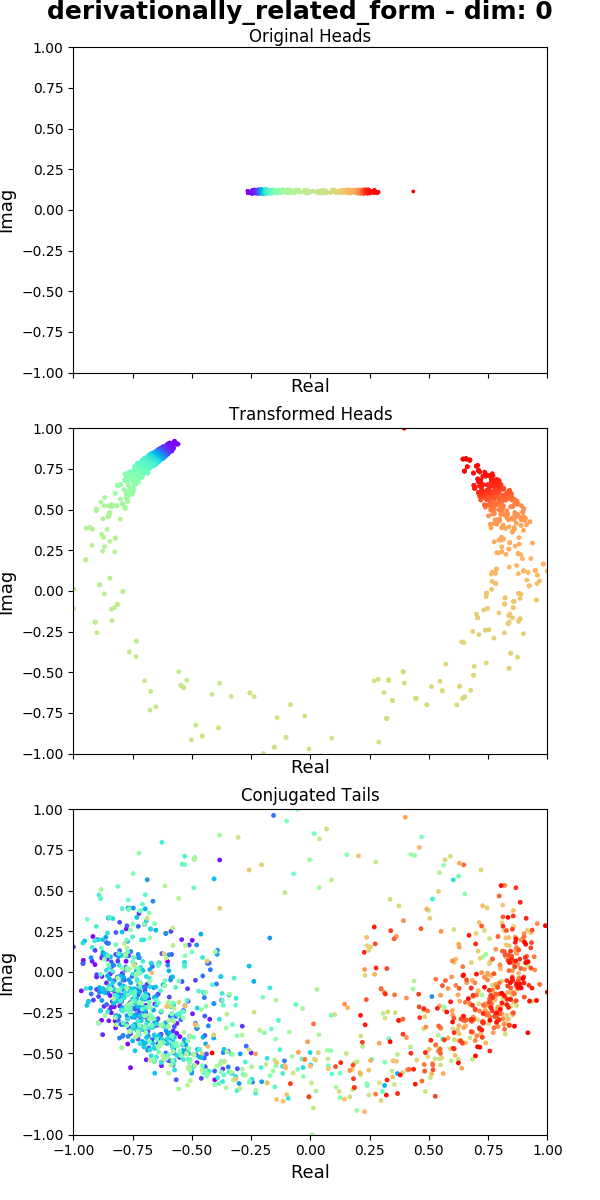}
		\caption{}
		\label{fig:fb:noInj}
	\end{subfigure}\hspace{2mm}
	\begin{subfigure}[b]{0.4\textwidth}
		\includegraphics[width=\linewidth,height=10 cm]{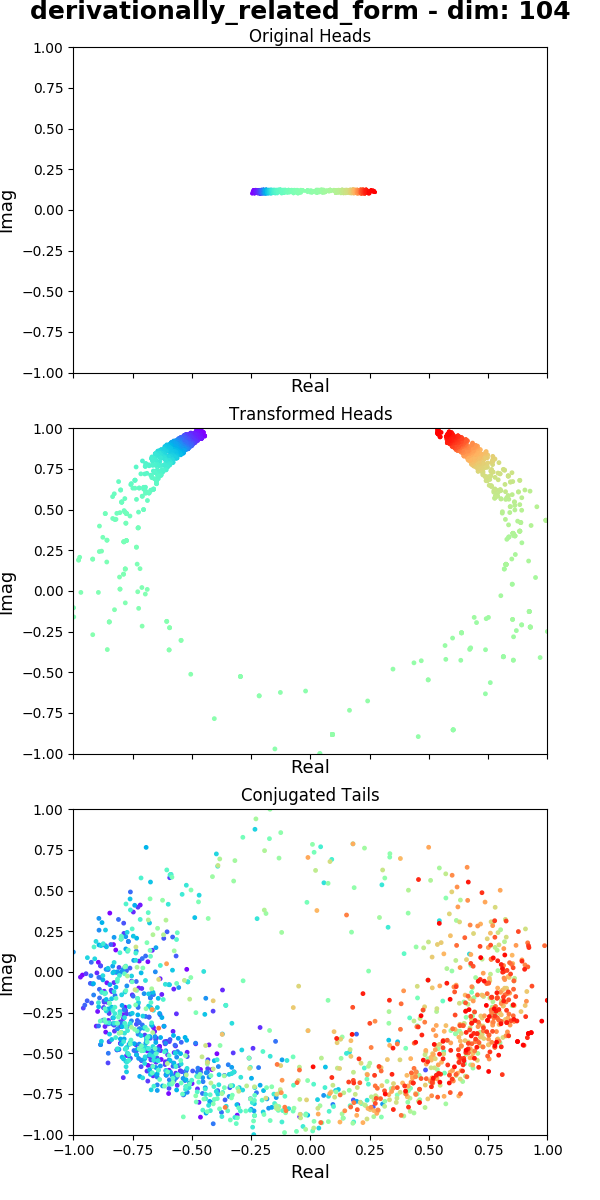}
			\caption{}
		\label{fig:fb:noInj}
	\end{subfigure}
\\
	\begin{subfigure}[b]{0.4\textwidth}
		\includegraphics[width=\linewidth,height=10 cm]{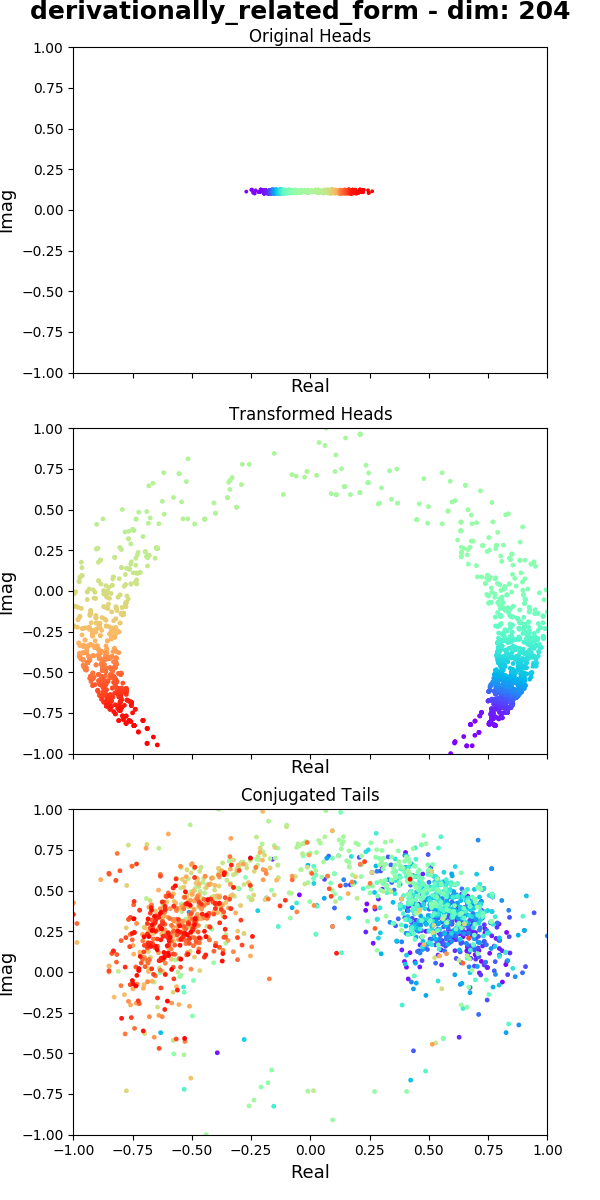}
			\caption{}
		\label{fig:1}
	\end{subfigure}\hspace{2mm}
	\begin{subfigure}[b]{0.4\textwidth}
		\includegraphics[width=\linewidth,height=10 cm]{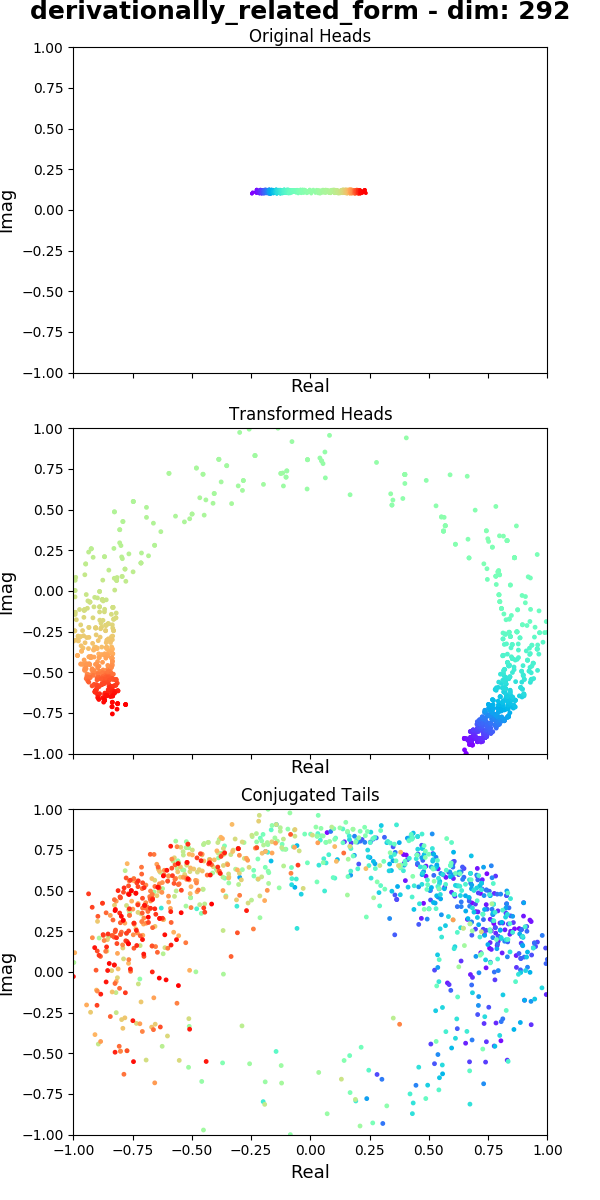}
			\caption{}
		\label{fig:2}
	\end{subfigure}
	\caption{Learned embeddings of entities connected by the derivationally-related-form relation.}
	\label{fig:spec1}
\end{figure*}

One of the main advantage of 5$^\bigstar$E lies in its ability to map a line to a circle after a relation-specific transformation of the head. 
This means that for a given relation, e.g., \textit{hypernym}, if the embeddings of the entities in the head position form a line, each of the transformed heads are either forming a line or a circle. 
Consequently, according to the formula $\mathbf{h}_{ri} = g_r(\mathbf{h}_i) = \bar{\mathbf{t}}_i, i=1, \ldots, d$,
the conjugated tails should also represent a shape which is close to the corresponding transformed head.
In this way, 5$^\bigstar$E captures different structural motifs of different groups of nodes. 
The Figures~\ref{fig:spec1} and \ref{fig:spec2} illustrate the state of each relation in different dimensions with regard to their original head, transformed head, and conjugated tail.
Figure~\ref{fig:spec1} depicts transformations in four different dimensions for the \textit{hypernym} relation where the line shows 
head entities which are located on a line and 
the transformed heads start to develop in the shape of a circle.
Given a fixed relation $r$ (e.g., hypernym), the similarity between all the transformed heads (which form a line or a circle) and the conjugated tails (similarly form a line or a circle) shows that the triples are considered positive in the vector space i.e.~ $ g_r(\mathbf{h}_{ij}) = \bar{\mathbf{t}}_{ij}, i=1, \ldots, d, j = 1, \ldots, n$ where $n$ refers to the number of entities in the head/tail position. $h_{ij}$ refers to the $i$th dimension of embeddings for $j$th head entity. 
In one dimension, all the heads in Figure~\ref{fig:spec2} form a line which are after transformation form a circle. 
Note that in each dimension ($i$th dimension), 5$^\bigstar$E learns different transformation functions. Each entity in head is shown as a point
Therefore, by taking different dimensions, the shape of the transformations changes per relation. 

\subsection{Learned Embedding: Relation Level}

The two relations \textit{hyponym} and \textit{hypernym} are inverse of each other. 
Therefore, we have
$\Im_{\textit{Hypernym}} = \bar{\Im}^{-1}_{\textit{Hyponym}}.$ 
As these representing matrices ($\Im$) are normalized, their determinant is equal to one. 
Consequently, we have $tr(\Im_{\textit{Hypernym}}) = tr(\bar{\Im}^{-1}_{\textit{Hyponym}})$ when the learned transformation function is elliptic. 
We conclude that for a pair of inverse relations, 
the learned transformation functions are in the same category (elliptic) for the $i$th element of the relation embedding. 
Figure~\ref{fig:sameD}, sub-figure (a) and (b) illustrate that the learned functions fall into the elliptic category for the same embedding dimension ($i = 12$) of \textit{hyponym} and \textit{hypernym}.  
The difference between the embeddings for this pair of inverse relations is their rotation. 
The same pattern is notable for \textit{hasPart} and \textit{partOf} relations in sub-figure (c) and (d) on $i=39$. 

\begin{figure*}[ht!]
\centering
	\begin{subfigure}[b]{0.3\textwidth}
		\includegraphics[width=\linewidth]{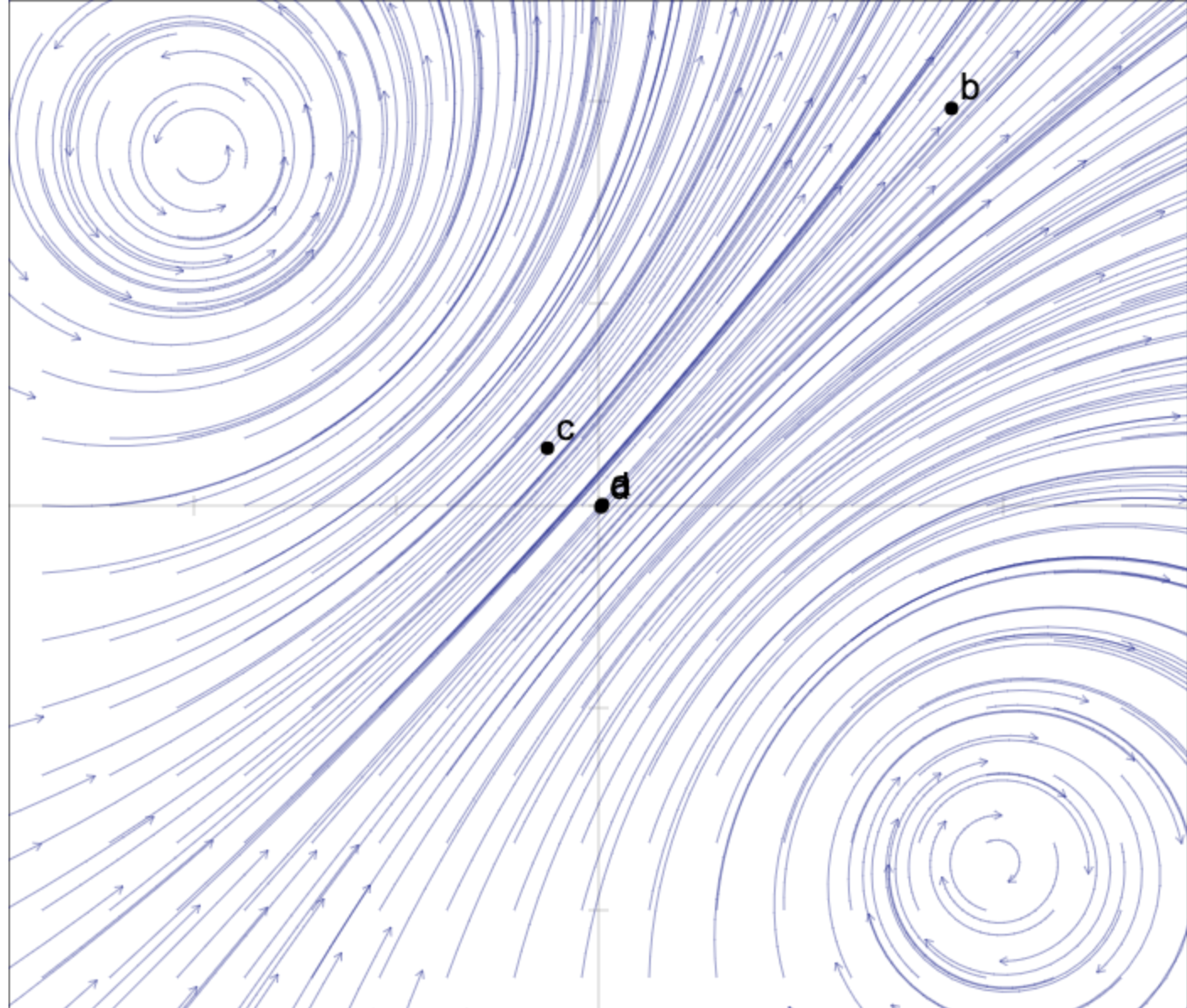}
		\caption{{\scriptsize 12-dim-hypernym}}
		\label{fig:fb:noInj}
	\end{subfigure}\hspace{5mm}
	\begin{subfigure}[b]{0.3\textwidth}
		\includegraphics[width=\linewidth]{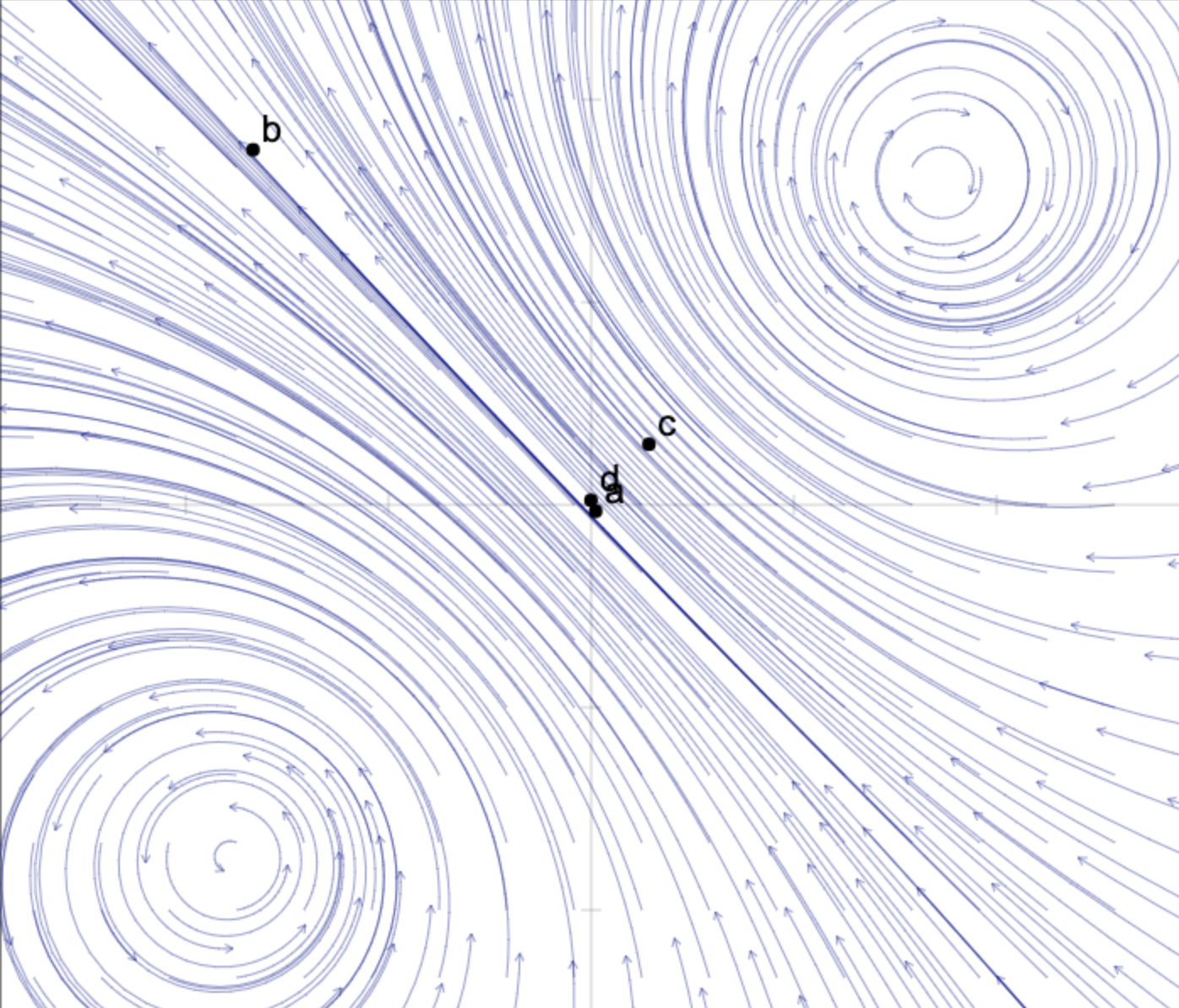}
		\caption{{\scriptsize 12-dim-hyponym}}
		\label{fig:fb:noInj}
	\end{subfigure}\\
	\begin{subfigure}[b]{0.3\textwidth}
		\includegraphics[width=\linewidth]{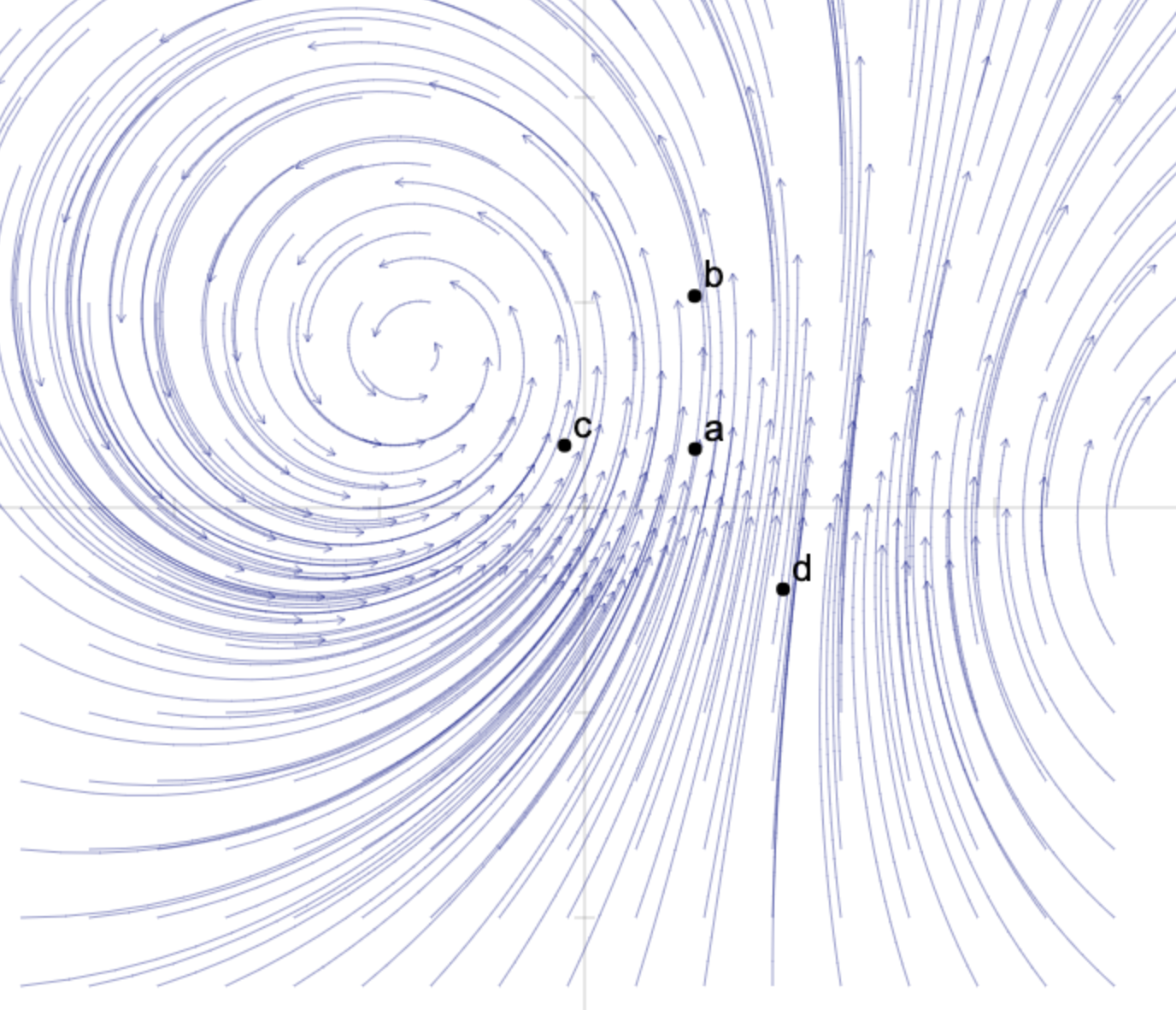}
		\caption{{\scriptsize 39-dim-haspart}}
		\label{fig:fb:noInj}
	\end{subfigure}\hspace{5mm}
	\begin{subfigure}[b]{0.3\textwidth}
		\includegraphics[width=\linewidth]{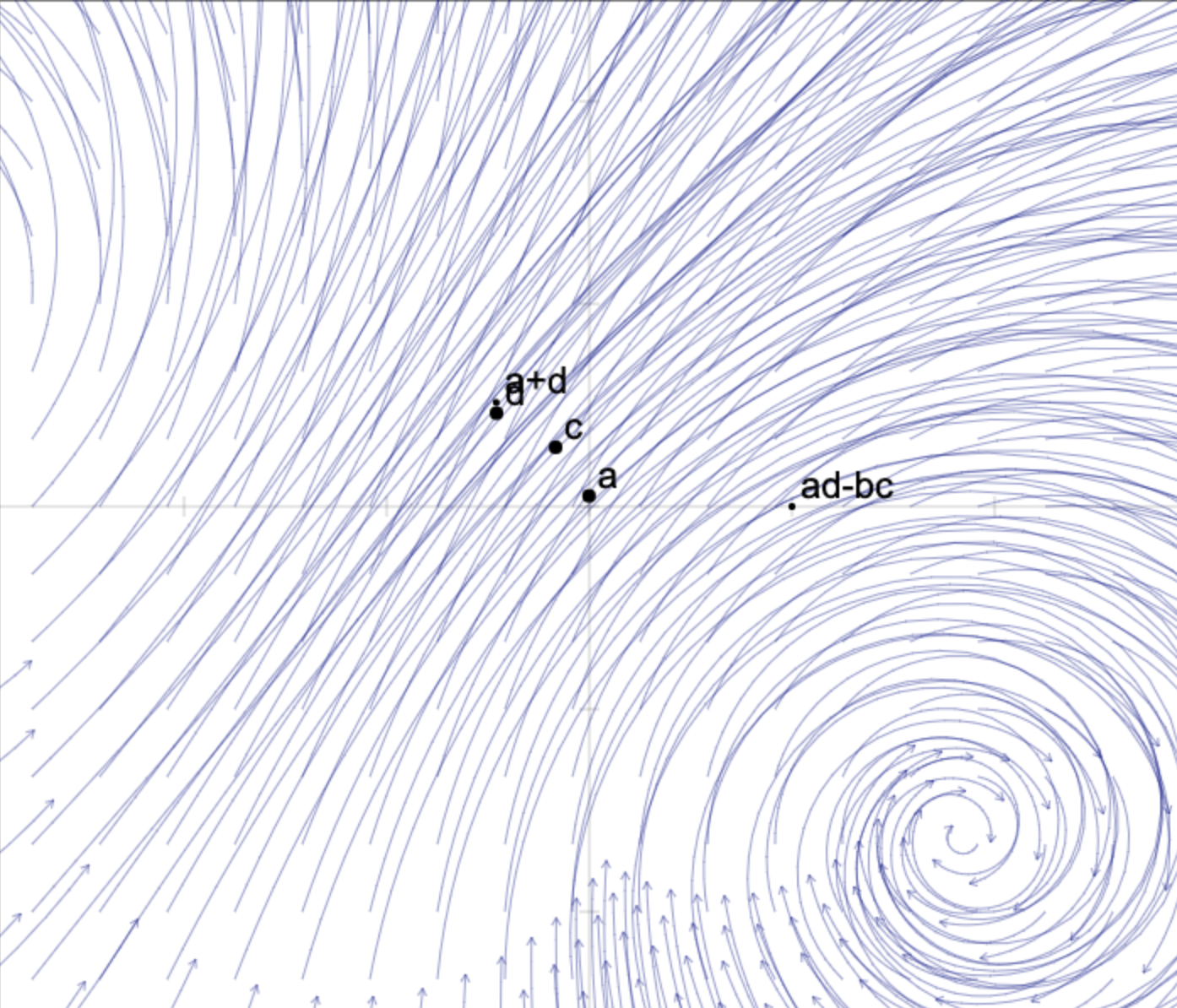}
		\caption{{\scriptsize 39-dim-partof}}
		\label{fig:fb:noInj}
	\end{subfigure}
	\caption{Embeddings for different relations usign the same dimension.}
	\label{fig:sameD}
\end{figure*}

\textbf{Construction Steps of Transformations}
\label{sub:transteps}

\begin{figure*}[ht!]
	\centering 
	\begin{subfigure}[b]{0.2\textwidth}
		\includegraphics[width=\linewidth]{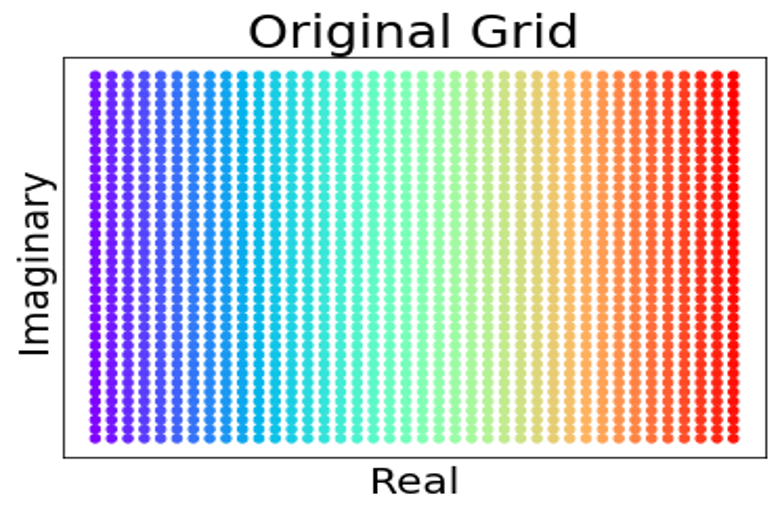}
		\caption{{\scriptsize Original Grid}}
		\label{fig:fb:noInj}
	\end{subfigure}%
	\begin{subfigure}[b]{0.2\textwidth}
		\includegraphics[width=\linewidth]{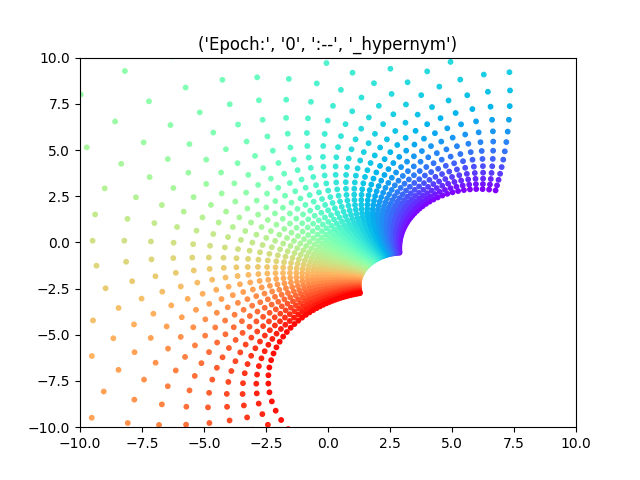}
		\caption{{\scriptsize epoch 0}}
		\label{fig:fb:noInj}
	\end{subfigure}%
	\begin{subfigure}[b]{0.2\textwidth}
		\includegraphics[width=\linewidth]{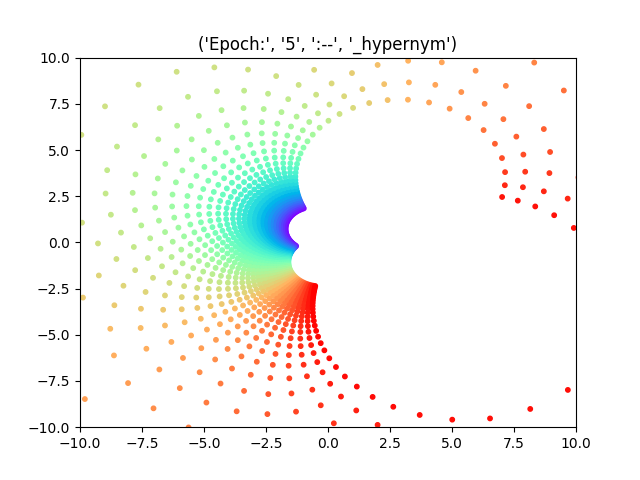}
		\caption{{\scriptsize epoch 5}}
		\label{fig:4c:eq}
	\end{subfigure}%
	\begin{subfigure}[b]{0.2\textwidth}
		\includegraphics[width=\linewidth]{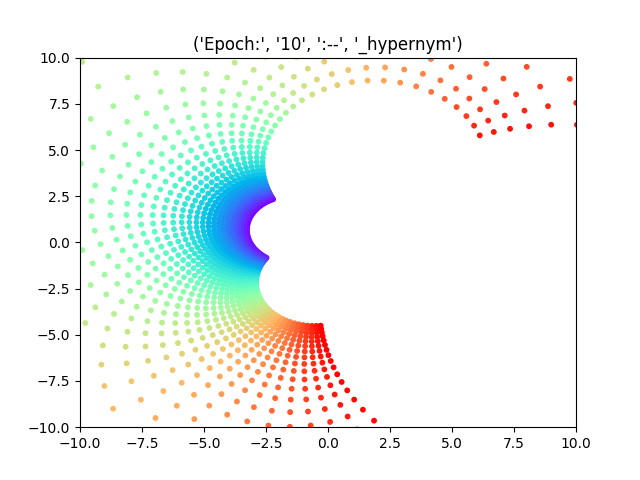}
		\caption{{\scriptsize epoch 10}}
		\label{fig:4d:sym}
	\end{subfigure}%
	\begin{subfigure}[b]{.2\textwidth}
		\includegraphics[width=\linewidth]{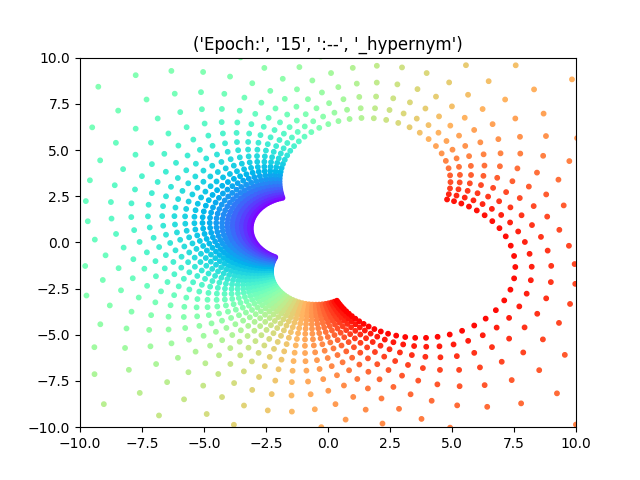}
		\caption{\scriptsize {epoch 15}}
		\label{fig:4a:noInj}%
	\end{subfigure}%
\\
	\begin{subfigure}[b]{.2\textwidth}
		\includegraphics[width=\linewidth]{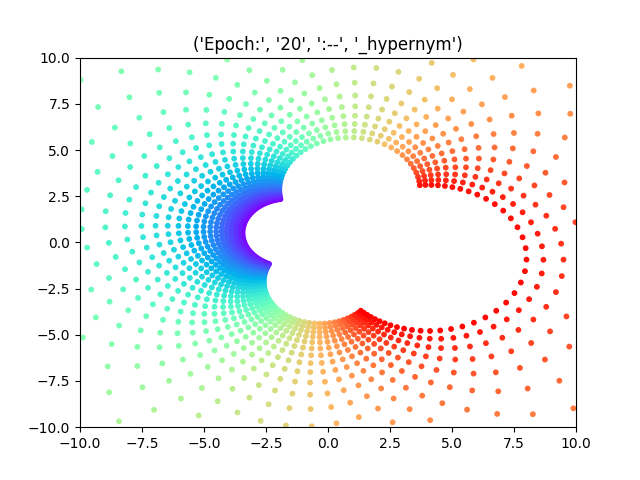}
		\caption{\scriptsize {epoch 20}}
		\label{fig:4a:noInj}%
	\end{subfigure}%
	\begin{subfigure}[b]{.2\textwidth}
		\includegraphics[width=\linewidth]{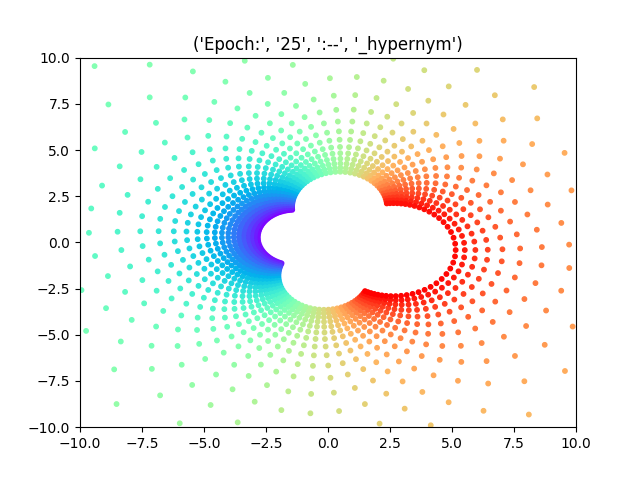}
		\caption{\scriptsize {epoch 25}}
		\label{fig:4a:noInj}%
	\end{subfigure}%
	\begin{subfigure}[b]{.2\textwidth}
		\includegraphics[width=\linewidth]{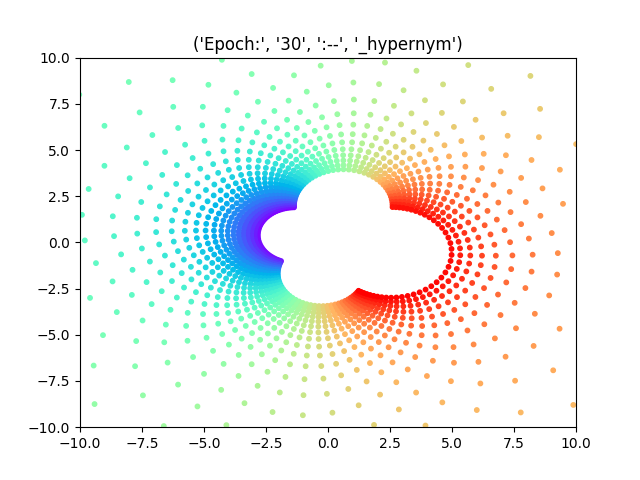}
		\caption{\scriptsize {epoch 35}}
		\label{fig:4a:noInj}%
	\end{subfigure}%
	\begin{subfigure}[b]{.2\textwidth}
		\includegraphics[width=\linewidth]{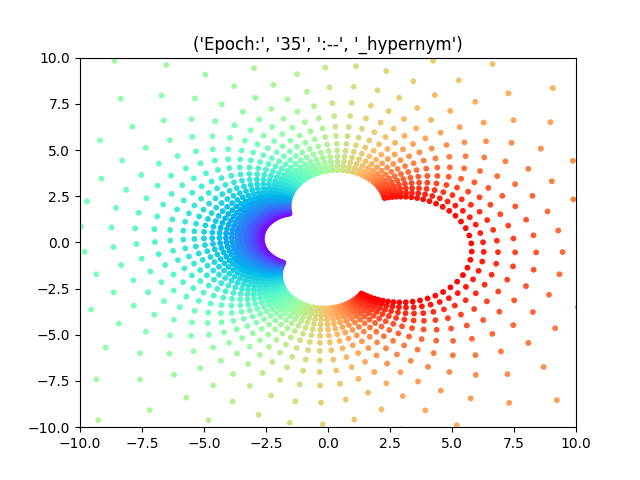}
		\caption{\scriptsize {epoch 40}}
		\label{fig:4a:noInj}%
	\end{subfigure}%
	\begin{subfigure}[b]{.2\textwidth}
		\includegraphics[width=\linewidth]{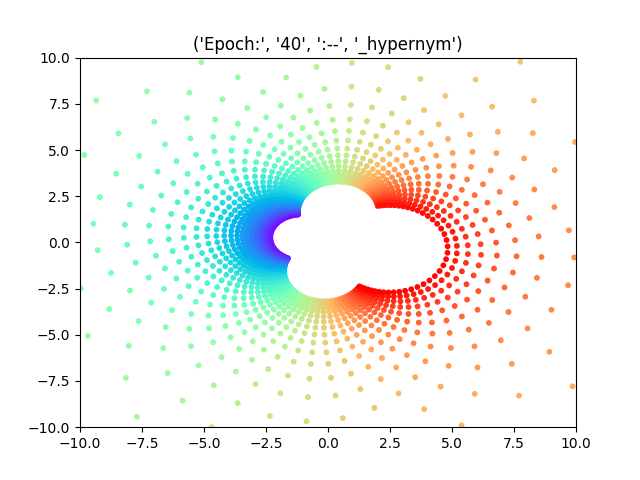}
		\caption{\scriptsize {epoch 49}}
		\label{fig:4a:noInj}%
	\end{subfigure}%
	\caption{Learned 5$^\bigstar$E embeddings for hasType relations in different epochs.}
	\label{fig:epochs}
\end{figure*}

In Figure~\ref{fig:epochs} we show the learned 5$^\bigstar$E embeddings for hasType relations in different epochs ranging from 0 to 49 (the stabilized epoch for this relation). The figure illustrates the gradual development of the M\"obious shape over different epochs.


\subsection{Experimental Setup}

\begin{table}[ht!]
	\centering  
	\begin{tabular}{@{}l |rrr@{}}
		\toprule 
		\textbf{Dataset} & \textbf{\#training} & \textbf{\#validation}  & \textbf{\#test} \\\midrule
		FB15k & 483,142 & 50,000 & 59,071 \\
		WN18  & 141,442 & 5,000 & 5,000 \\
		FB15k-237 & 272,115& 17,535& 20,466 \\
		WN18RR & 86,835 & 3,034 & 3,134 \\
		NELL-995-h25 & 122,618 & 9,194 & 9,187 \\
		NELL-995-h50 & 72,767 & 5,440 & 5,393 \\
		NELL-995-h75 & 59,135 & 4,441 & 4,389 \\
		NELL-995-h100 & 50,314 & 3,763 & 3,746 \\
		\bottomrule
	\end{tabular}
	\caption{\textbf{Dataset Statistics.} Split of datasets in terms of number of triples.} 
	\label{tab:dssplit}
\end{table}

\begin{itemize}
\item 
\textbf{Evaluation Metrics and Statistics of Standard Benchmarks}
FB15K is a standard benchmark created from the original FreeBase KG~\cite{bollacker2008freebase}.
FB15K-237~\cite{toutanova2015observed} is a sub-version of FB15K in which the inverse relations have been removed.
WN18~\cite{bordes2013transe} is a lexical database with hierarchical collection for the English language that was derived from the original WordNet dataset~\cite{miller1995wordnet}. 
The main relation patterns in this dataset are symmetry antisymmetry and inversion.
WN18RR is a sub-version of WN18 for which the inverse relations are omitted. 
We also considered four different versions of the NELL dataset in our evaluations. 
In Table~\ref{tab:dssplit}, we represent the statistics of each corresponding detest.


\item \textbf{Hyperparameter Search}
Generally, the range of embedding dimensions were fixed to \{100, 500 \} with learning rates of \{0.01, 0.05, 0.1 \}, except for the versions of NELL datasets, where d = 200 was used. 
This was done to show the effectiveness of the model in different dimensions. 
The considered batch sizes for all datasets are \{100, 500, 1000, 2000 \}. Regularization coefficients are tested among \{0.01, 0.05, 0.1, 0.5 \}.
For all the required values in our evaluation tables, which were not reported in the original papers, we trained the models and reported our findings. 
To train models on NELL-995-h25, NELL-995-h50, NELL-995-h75 and NELL-995-h100, we proceeded as follows: First, the NELL-995-h25 dataset was used to find the best hyperparameter setting among the aforementioned hyperparameter settings. After the best setting is found per model, the same hyperparameters for each model are applied to NELL-995-h50, NELL-995-h75 and NELL-995-h100. The same fixed hyperparameters per model are then also used when training a NELL-995-h25 model with a different embedding dimension.
\end{itemize}

\subsection{Proof of Subsumption (Proposition 1)}
In this part, we prove that 5$^\bigstar$E subsumes five popular KGE models namely DistMult, ComplEx, pRotatE, RotatE and TransE. 
Here, we provide proofs for the following items:
\\
(a) $5^\bigstar$E subsumes DistMult with its original score function $f(h,r,t) = Re(\langle \mathbf{h}_{r}, \bar{\mathbf{t}}\rangle),$
where

\begin{equation}
\begin{split}
    \label{eq:projTra}
    &\mathbf{h}_{r} = [\mathbf{h}_{r1}, \ldots, \mathbf{h}_{rd}], \,\,
    \mathbf{h}_{ri} = g_r(\mathbf{h}_i) \doteq \Im_{ri} [\mathbf{h}_i , 1]^T,~\\
    &i = 1,\ldots,d.
\end{split}
\end{equation}

Note that here we use homogeneous coordinates.
With the symbol $\doteq$, we mean that the left hand side of Equation~\ref{eq:projTra} (i.e.~$g_r(\mathbf{h}_i)$) equals the results of the right hand side after dehomogenization.

(b) $5^\bigstar$E subsumes pRotatE with its original score function $f(h,r,t) = Re(\langle \mathbf{h}_{r}, \bar{\mathbf{t}}\rangle)$.

(c) $5^\bigstar$E subsumes ComplEx with its original score function $f(h,r,t) =  Re(\langle \mathbf{h}_r , \bar{\mathbf{t}} \rangle)$.

(d) $5^\bigstar$E subsumes RotatE with score function $f(h,r,t) = -\| \mathbf{h}_r - \mathbf{t}\|$ (changed inner product to distance).

(e) $5^\bigstar$E subsumes TransE with score function $f(h,r,t) = -\| \mathbf{h}_r - \mathbf{t}\|$ (changed inner product to distance).

\begin{proof} 
(a) From equation \ref{eq:projTra}, we have
\begin{equation*}
    \mathbf{h}_{ri} = g_r(\mathbf{h}_i) \doteq \Im_{ri} [\mathbf{h}_i , 1]^T, 
    i = 1,\ldots,d,
\end{equation*}
where $\Im_{ri} = \begin{bmatrix}
    \mathbf{r}_{ai} \,& \mathbf{r}_{bi}\\
    \mathbf{r}_{ci} \,& \mathbf{r}_{di}
    \end{bmatrix}$.
Let $Im(\mathbf{r}_{ai}) = Im(\mathbf{r}_{bi}) = Im(\mathbf{r}_{ci}) = Im(\mathbf{r}_{di}) = Im(\mathbf{h}_{i}) =  Re(\mathbf{r}_{ci}) = Re(\mathbf{r}_{bi}) = 0,$  and $Re(\mathbf{r}_{di}) = 1.$ Therefore, we have

\begin{equation*}
\begin{split}
    &\mathbf{h}_{ri}= g_r(\mathbf{h}_i) \doteq \Im_{ri} [\mathbf{h}_i , 1]^T = \begin{bmatrix} \mathbf{r}_{ai} & 0 \\
    0 & 1
    \end{bmatrix} [\mathbf{h}_i , 1]^T \\
    &= [\mathbf{r}_{ai} \mathbf{h}_i, 1]^T,
    i = 1,\ldots,d,
\end{split}
\end{equation*}

from which we conclude $\mathbf{h}_{ri} = \mathbf{r}_{ai} \mathbf{h}_i$ after dehomogenization. 
Therefore, the score function is defined as

$ f(h,r,t) = Re(\langle \mathbf{h}_{r}, \bar{\mathbf{t}}\rangle) = \sum_{i=1}^{d} (\mathbf{h}_i \mathbf{r}_{ai} \mathbf{t}_i) = f^{DistMult}(h,r,t).$ 
From these equality in quotations, we conclude that DistMult is a special case of 5$^\bigstar$E and therefore, is subsumed by 5$^\bigstar$E.
\end{proof}

\begin{proof}
(b) From Equation~\ref{eq:projTra}, we can derive
\begin{equation*}
    \mathbf{h}_{ri} = g_r(\mathbf{h}_i) \doteq \Im_{ri} [\mathbf{h}_i , 1]^T, 
    i = 1,\ldots,d,
\end{equation*}
where $\Im_{ri} = \begin{bmatrix}
    \mathbf{r}_{ai} \,& \mathbf{r}_{bi}\\
    \mathbf{r}_{ci} \,& \mathbf{r}_{di}
    \end{bmatrix}$.
Now, if we assume $Im(\mathbf{r}_{bi}) = Im(\mathbf{r}_{ci}) = Im(\mathbf{r}_{di}) = Re(\mathbf{r}_{ci}) = Re(\mathbf{r}_{bi}) = 0,$ and $Re(\mathbf{r}_{di}) = 1,$ and $|\mathbf{r}_{ai}| = |\mathbf{h}_{i}| = |\mathbf{t}_{i}| = 1.$ 
Consequently, we can conclude
\begin{equation*}
\begin{split}
    &\mathbf{h}_{ri} = g_r(\mathbf{h}_i) \doteq \Im_{ri} [\mathbf{h}_i , 1]^T = \begin{bmatrix} \mathbf{r}_{ai} & 0 \\
    0 & 1
    \end{bmatrix} [\mathbf{h}_i , 1]^T \\
    &= [\mathbf{r}_{ai} \mathbf{h}_i,1]^T,
    i = 1,\ldots,d.
\end{split}
\end{equation*}

Therefore, we end up with $\mathbf{h}_{ri} = \mathbf{r}_{ai} \mathbf{h}_i$ after dehomogenization.
This leads us to define the score function as
$ f(h,r,t) = Re(\langle \mathbf{h}_{r}, \bar{\mathbf{t}}\rangle) = \sum_{i=1}^{d} Re(\mathbf{h}_{ri} \bar{\mathbf{t}}_i) = \sum_{i=1}^d Re(\mathbf{h}_{ri}) Re(\mathbf{t}_i) + Im(\mathbf{h}_{ri}) Im(\mathbf{t}_i)$ which is equivalent with the pRotatE score function \cite{sun2019rotate} i.e.~$f(h,r,t) = -\| \mathbf{h} \circ \mathbf{r} - \mathbf{t} \|$ because $|\mathbf{r}_{ai}| = |\mathbf{h}_{i}| = |\mathbf{t}_{i}| = 1$ (the length of modulus of each elements of embedding vector for entities and relations does not participate in computation of score of triples). 
\end{proof}

\begin{proof}
(c) From Equation\ref{eq:projTra}, we have
\begin{equation*}
    \mathbf{h}_{ri} = g_r(\mathbf{h}_i) \doteq \Im_{ri} [\mathbf{h}_i , 1]^T, 
    i = 1,\ldots,d,
\end{equation*}
where $\Im_{ri} = \begin{bmatrix}
    \mathbf{r}_{ai} \,& \mathbf{r}_{bi}\\
    \mathbf{r}_{ci} \,& \mathbf{r}_{di}
    \end{bmatrix}$.
Let $Im(\mathbf{r}_{bi}) = Im(\mathbf{r}_{ci}) = Im(\mathbf{r}_{di}) = Re(\mathbf{r}_{ci}) = Re(\mathbf{r}_{bi}) = 0,$  and $Re(\mathbf{r}_{di}) = 1.$  

From these assumptions, we can conclude 
\begin{equation*}
\begin{split}
    &\mathbf{h}_{ri} = g_r(\mathbf{h}_i) \doteq \Im_{ri} [\mathbf{h}_i , 1]^T = \begin{bmatrix} \mathbf{r}_{ai} & 0 \\
    0 & 1
    \end{bmatrix} [\mathbf{h}_i , 1]^T \\
    &= [\mathbf{r}_{ai} \mathbf{h}_i,1]^T,
    i = 1,\ldots,d,
\end{split}
\end{equation*}

where after dehomogenization, we have $\mathbf{h}_{ri} = \mathbf{r}_{ai} \mathbf{h}_i $. 
From these and the assumption of $f(h,r,t) = Re(\langle \mathbf{h}_r, \bar{\mathbf{t}} \rangle)$ as the base score function, we then prove
$ f(h,r,t) = Re(\langle \mathbf{h}_{r}, \bar{\mathbf{t}}\rangle) = \sum_{i=1}^{d} Re(\langle \mathbf{h}_i, \mathbf{r}_{ai}, \bar{\mathbf{t}_i} \rangle) = f^{ComplEx}(h,r,t).$ 
Therefore, this shows that the ComplEx model is a special case in a variant of 5$^\bigstar$E with score function of $f(h,r,t) = Re(\langle \mathbf{h}_r, \bar{\mathbf{t}} \rangle)$.
This also means that ComplEx is subsumed by 5$^\bigstar$E.
\end{proof}

\begin{proof}
(d) From Equation~\ref{eq:projTra}, we have
\begin{equation*}
    \mathbf{h}_{ri} = g_r(\mathbf{h}_i) \doteq \Im_{ri} [\mathbf{h}_i , 1]^T, 
    i = 1,\ldots,d,
\end{equation*}
where $\Im_{ri} = \begin{bmatrix}
    \mathbf{r}_{ai} \,& \mathbf{r}_{bi}\\
    \mathbf{r}_{ci} \,& \mathbf{r}_{di}
    \end{bmatrix}$.
Let $Im(\mathbf{r}_{bi}) = Im(\mathbf{r}_{ci}) = Im(\mathbf{r}_{di}) = Re(\mathbf{r}_{bi}) = Re(\mathbf{r}_{ci}) = 0,$  and $Re(\mathbf{r}_{di}) = 1, |\mathbf{r}_{ai}| = 1.$ 
We then have
\begin{equation*}
\begin{split}
    &\mathbf{h}_{ri} = g_r(\mathbf{h}_i) \doteq \Im_{ri} [\mathbf{h}_i , 1]^T = \begin{bmatrix} \mathbf{r}_{ai} & 0 \\
    0 & 1
    \end{bmatrix} [\mathbf{h}_i , 1]^T \\
    &= [\mathbf{r}_{ai} \mathbf{h}_i, 1],
    i = 1,\ldots,d.
\end{split}
\end{equation*}

With $f(h,r,t) = -\| \mathbf{h}_r - \mathbf{t}\|$ as score function, we have

$f(h,r,t) = -\| \mathbf{h}_r - \mathbf{t}\| = -\| \mathbf{h} \circ \, \mathbf{r}_a - \mathbf{t}\| = f^{RotatE}(h,r,t).$ 
Therefore, RotatE is a special case of in a variant of 5$^\bigstar$E with score function of $f(h,r,t) = -\|\mathbf{h}_r - \mathbf{t}\|$ and is subsumed by it.
\end{proof}

\begin{proof}
(e)
From Equation~\ref{eq:projTra}, we conclude
\begin{equation*}
    \mathbf{h}_{ri} = g_r(\mathbf{h}_i) \doteq \Im_{ri} [\mathbf{h}_i , 1]^T, 
    i = 1,\ldots,d,
\end{equation*}
where $\Im_{ri} = \begin{bmatrix}
    \mathbf{r}_{ai} \,& \mathbf{r}_{bi}\\
    \mathbf{r}_{ci} \,& \mathbf{r}_{di}
    \end{bmatrix}$.
Here, we assume $Im(\mathbf{r}_{ai}) = Im(\mathbf{r}_{bi}) = Im(\mathbf{r}_{ci}) = Im(\mathbf{r}_{di}) = Re(\mathbf{r}_{ci}) = 0,$  and $Re(\mathbf{r}_{ai}) = Re(\mathbf{r}_{di}) = 1.$ 
This assumption leads us to have
\begin{equation*}
\begin{split}
    &\mathbf{h}_{ri} = g_r(\mathbf{h}_i) \doteq \Im_{ri} [\mathbf{h}_i , 1]^T = \begin{bmatrix} 1 & \mathbf{r}_{bi} \\
    0 & 1
    \end{bmatrix} [\mathbf{h}_i , 1]^T \\
    &= [\mathbf{h}_i + \mathbf{r}_{bi}, 1],
    i = 1,\ldots,d.
\end{split}
\end{equation*}

After dehomogenization, we have $\mathbf{h}_{ri} = \mathbf{h}_i + \mathbf{r}_{bi}$ which consequently gives 
$f(h,r,t) = -\| \mathbf{h}_r - \mathbf{t}\| = -\| \mathbf{h} + \mathbf{r}_b - \mathbf{t}\| = f^{TransE}(h,r,t).$ 
Therefore, it is proven that the TransE model is a special case in a variant of 5$^\bigstar$E with score function of $f(h,r,t) = -\|\mathbf{h}_r - \mathbf{t}\|$.
This means that 5$^\bigstar$E subsumes by TransE.
\end{proof}



\subsection{Proof of Full Expressiveness (Corollary 1)}
Here we prove that the $5^\bigstar$E model is fully expressive.

\begin{proof}
We divide the proof into two argumentation steps:\\

First, we show that 5$^\bigstar$E can express any ranking tensor $\mathcal{A} \in \mathbb{R}^{n_e \times n_e \times n_r},$ where $n_e, n_r$ are the number of entities and relations in a KG respectively. 
$\alpha_{ikj}$ is the $ikj$th element of the tensor $\mathcal{A}$ that corresponds to the triple $(h_i, r_k, t_j)$.\\
For a triple $(h_i, r_k, t_j)$ which is scored higher than a triple $(h'_i, r'_k, t'_j)$ by the model, the ranking tensor gives a lower rank to $(h_i, r_k, t_j)$ than to $(h'_i, r'_k, t'_j)$. 
More details can be found in~\cite{wang2018multi}.

Second, for any boolean tensor $\mathcal{B} \in \{0,1\}^{n_e \times n_e \times n_r}$, there is a ranking matrix obtained by the 5$^\bigstar$E model which is consistent with the boolean tensor.
More precisely, if we assume $\beta_{ikj} = 1$ where ($h_i,r_k,t_j$) is a positive triple, and $\beta_{i'kj'} = 0$ where ($h'_i,r_k,t'_j$) is a negative triple, then we have $\alpha_{i'kj'} < \alpha_{i'kj'}$.  \\

The first two argumentation steps allow to conclude that 5$^\bigstar$E is fully expressive, i.e.~capable of representing any ground truth over triples of a KG.

Here, we provide the proofs for expressiveness by explaining the steps in detail.

For the proof of the first argumentation, let $\mathcal{M}_m \in \mathbb{R}^{n_e \times n_r \times n_e}$ be the tensor corresponding to the score function of the model $m$ obtained by an assignment to embeddings of entities and relations in a $\mathcal{KG}$. 
The $ikj$-th element of $\mathcal{M}_m$ is denoted by $\mu_{ikj}$ that equals to the score of a model $m$ for a triple $(h_i,r_k,t_j)$ i.e.~$\mu_{ikj} = f^m(h_i,r_k,t_j).$
Given a score tensor $\mathcal{M}_m$, the corresponding ranking tensor $\mathcal{A}_m$ is obtained by applying a mapping $\phi: \mathbb{R}^{n_e \times n_r \times n_e} \rightarrow \mathbb{N}^{n_e \times n_r \times n_e}$ in that $\mu_{i'kj'} \leq \mu_{ikj} \leftrightarrow \alpha_{i'kj'} = \phi(\mu_{i'kj'}) \geq \alpha_{ikj} = \phi(\mu_{ikj}).$
The authors of~\cite{wang2018multi} prove that the ComplEx model is universal i.e.~given any ranking matrix $\mathcal{A}$, there are assignments to embedding vectors such that the obtained score tensor fulfills $\mathcal{A} = \phi(\mathcal{M}_{ComplEx})$. 

In the subsumption proof case (c), we proved that 5$^\bigstar$E subsumes ComplEx, therefore, for any given ranking matrix $\mathcal{A}$ there is a vector assignment to embeddings of entities and relations such that the score of triples create a tensor that satisfies $$\mathcal{A} = \phi(\mathcal{M}_{5^\bigstar\text{E}}).$$

The authors of~\cite{wang2018multi} show that for a given boolean matrix $\mathcal{B}$, there is a ranking matrix which is consistent with the boolean matrix. 
Therefore, for any given boolean matrix $\mathcal{B}$, there exists a ranking $\mathcal{A} = \phi(\mathcal{M}_{5^\bigstar\text{E}})$ which is consistent with it. 

From the first and second argument, we conclude that for any given ground truth over a $\mathcal{KG}$, there is an assignment to embeddings of entities and relations in the $\mathcal{KG}$ such that 5$^\bigstar$E separates the correct triples from incorrect ones. 
This means that 5$^\bigstar$E is fully expressive.
\end{proof}

\subsection{Proof of Pattern Inference (Propositions 2, 3, 4, 5)}
Here we prove that our model is able to infer symmetric, inverse, composition and reflexive patterns.

(f) Let $r_1, r_2, r_3 \in \mathcal{R}$ be relations such that $r_3$ is a composition (e.g.~\textit{UncleOf}) of $r_1 (\textit{e.g.~BrotherOf})$ and $r_2 (\textit{e.g.~FatherOf}).$ 
$5^\bigstar$E infers composition with $\Im_{r_1} \Im_{r_2} = \Im_{r_3}.$ 

(g) Let $r_1 \!\in\! \mathcal{R}$ be the inverse of $r_2 \!\in\! \mathcal{R}$.
$5^\bigstar$E infers this pattern with $\Im_{r_1} = \Im_{r_2^{-1}}.$%

(h) Let $r \in \mathcal{R}$ be symmetric. 
$5^\bigstar$E infers the symmetric pattern if $\Im_{r} = \Im_{r}^{-1}.$ 

(i) Let $r \in \mathcal{R}$ be a reflexive relation.
In dimension $d$, $5^\bigstar$E infers reflexive patterns with $O(2^d)$ distinct representations of entities if the fixed points of the involved transformations are non-identical.

Here we provide the proofs of the aforementioned propositions for pattern inference. 
We use the following proposition from~\cite{kisil2012geometry} in our proof:

\begin{proposition}\cite{kisil2012geometry}
Let $\Im_{r_1}, \Im_{r_2}$ be matrices of two projective transformations $\tau_{r_1}, \tau_{r_2}$.
which are respectively associated to two M\"obius transformations $\vartheta_{r_1}, \vartheta_{r_2}$.
The product of the two matrices results in a matrix $\Im_{r_3} = \Im_{r_1} \Im_{r_2}$, which is associated to a projective transformation $\tau_3$ corresponding to the composition of the two M\"obius transformations $\vartheta_{r3} = \vartheta_{r1}\circ \vartheta_{r2}$.
\label{prop:prop3}
\end{proposition}

Given this proposition, we can then continue:
\begin{proof}
(f).
A relation $r_3$ is composed from two relations $r_1, r_2$ if

\begin{equation}
    \forall e_1, e_2 , e_3 \in \mathcal{E}, (e_1, r_1, e_2) \land (e_2, r_2, e_3) \rightarrow (e_1, r_3, e_3).
\end{equation}

A model infers a composition pattern when for given entities $e_1,e_2,e_3$, if the score of the model represents triples $(e_1,r_1,e_2)$ and $(e_2,r_2,e_3)$ as positive, it also represents ($e_1,r_3,e_3$) as positive, that is 

\begin{equation}
    \begin{split}
\label{compconstrain}
    g_{r1}(\mathbf{e}_{1i}) = \mathbf{e}_{2i}, \\
    g_{r2}(\mathbf{e}_{2i}) = \mathbf{e}_{3i}, 
\end{split}
\end{equation}

then $g_{r3}(\mathbf{e}_{1i}) = \mathbf{e}_{3i}, i=1, \ldots, d,$ where 
\begin{equation}
\label{eq:mobTra}
\begin{split}
        &g_{ri}(\mathbf{h}_i) = \vartheta(\mathbf{h}_i,\mathbf{r}_i) =  \frac{\mathbf{r}_{ai} \mathbf{h}_i + \mathbf{r}_{bi}}{\mathbf{r}_{ci} \mathbf{h}_i + \mathbf{r}_{di}}, \, \, \mathbf{r}_{ai}\mathbf{r}_{di} - \mathbf{r}_{bi}\mathbf{r}_{ci} \neq 0,\\
        &i=1,\ldots,d. 
\end{split}
\end{equation}.

From Equations~\ref{compconstrain}, we insert $\mathbf{e}_{2i} = g_{r1}(\mathbf{e}_{1i})$ into $g_{r2}(\mathbf{e}_{2i}) = \mathbf{e}_{3i},$ which gives 
$g_{r2}(g_{r1}(\mathbf{e}_{1i})) = \mathbf{e}_{3i}$. 
Therefore, we have
$$g_{r2}\circ g_{r1}(\mathbf{e}_{1i}) =  \mathbf{e}_{3i}.$$
This means two M\"obius transformations $g_{r2}, g_{r1}$ are composed. 
Considering the Proposition~\ref{prop:prop3}, and assuming $\Im_{r3} = \Im_{r1} \Im_{r2}$, 
we have 
$g_{r2} \circ g_{r1}(\mathbf{e}_{1i}) = g_{r3}(\mathbf{e}_{1i}) =  \mathbf{e}_{3i}.$ This means that the triple $(e_1, r_3, e_3)$ must be positive (= inferred to be positive). 

Note that the above mentioned proof holds for the case where $g_r(\mathbf{h}_{ri}) = \mathbf{t}_i, i=1, \ldots, d$ which concludes that a triple $(h,r,t)$ is positive. 
It also holds for the case that imaginary parts of embeddings become zero. 

With assumption $g_r(\mathbf{h}_{ri}) = \bar{\mathbf{t}}_i, i=1, \ldots, d$, the imaginary part of $\mathbf{e}_{2i}$ must be zero. 
\end{proof}

\begin{proof}
(g). A relation $r_2$ is inverse of relation $r_1$ if

\begin{equation}
    \forall e_1, e_2 \in \mathcal{E}, (e_1, r_1, e_2) \rightarrow (e_2, r_2, e_1).
\end{equation}

A model infers the inverse pattern for given entities $e_1,e_2$, if the following holds: If $(e_1,r_1,e_2)$ are represented as positive, then $(e_2,r_2,e_1)$ as positive are also represented as positive.
This can be formulated as

\begin{align}
\label{invconstrain}
    g_{r1}(\mathbf{e}_{1i}) = \mathbf{e}_{2i}, 
\end{align}

then $g_{r2}(\mathbf{e}_{2i}) = \mathbf{e}_{1i}.$

From Equation~\ref{invconstrain}, we have $\mathbf{e}_{2i} = g_{r1}(\mathbf{e}_{1i})$. 
Since $g_{r1}$ is M\"obius and is invertible, and $\Im_{r1} = \Im^{-1}_{r2}$ (from the assumption), we have 
$$\mathbf{e}_{2i} = g_{r2}^{-1}(\mathbf{e}_{1i}).$$ 
Therefore, we have $$\mathbf{e}_{1i} = g_{r2}(\mathbf{e}_{2i}).$$
This means that the triple $(e_2, r_2, e_1)$ must be positive (inferred as positive). 

When $g_r(\mathbf{h}_{ri}) = \mathbf{t}_i, i=1, \ldots, d$ (denoting triple ($h,r,t$) is positive), the above-mentioned proof holds. 
The proof also holds for the case that the imaginary part of embeddings become zero. 

In the case that the equality $g_r(\mathbf{h}_{ri}) = \bar{\mathbf{t}}_i, i=1, \ldots, d$, the previous assumption is changed to $\Im_2 = \bar{\Im}^{-1}_1.$

\end{proof}

\begin{proof}
(h).
A relation $r$ is symmetric if

\begin{equation}
\label{eq:sym}
    \forall e_1, e_2 \in \mathcal{E}, (e_1, r, e_2) \rightarrow (e_2, r, e_1).
\end{equation}

A model infers a symmetric pattern when for given entities $e_1,e_2,$ the following holds: 
If the model represents triples $(e_1,r,e_2)$ as positive, then it also represents $(e_2,r,e_1)$ as positive.
This can be formulated as 
\begin{align}
\label{symconstrain}
    g_{r}(\mathbf{e}_{1i}) = \mathbf{e}_{2i}. 
\end{align}

From this and taking Equation~\ref{eq:sym} into account, we have $g_{r}(\mathbf{e}_{2i}) = \mathbf{e}_{1i}$.

From Equation~\ref{symconstrain}, we have 

$\mathbf{e}_{2i} = g_{r}(\mathbf{e}_{1i})$. 
Since $g_{r}$ is M\"obius, and $\Im_{r} = \Im^{-1}_{r}$ (from the assumption), we have 
$$\mathbf{e}_{2i} = g_{r}^{-1}(\mathbf{e}_{1i}).$$ 
Therefore, we have $$\mathbf{e}_{1i} = g_{r}(\mathbf{e}_{2i}).$$
This means that the triple $(e_2, r, e_1)$ must be positive (= inferred as positive). 

Note that the above mentioned proof is held when $g_r(\mathbf{h}_{ri}) = \mathbf{t}_i, i=1, \ldots, d$.
From this, we conclude that a triple $(h,r,t)$ is positive. 
It also holds for the case that the imaginary part of embeddings become zero.

When $g_r(\mathbf{h}_{ri}) = \bar{\mathbf{t}}_i, i=1, \ldots, d$ holds, the assumption is changed to $\Im = \bar{\Im}^{-1}.$
\end{proof}

\begin{proof}
(i).
Let $r$ be reflexive. 
We have $\mathbf{e}_{ri} = \mathbf{e}_i, i=1, \ldots,d.$ 
Because $\mathbf{e}_{ri}$ is a M\"obius transformation of the head, the formula $\mathbf{e}_{ri} - \mathbf{e}_i = 0$ gives its fixed points. 
As each M\"obius transformation has at most two fixed points, we can therefore conclude that there are $O(2^d)$ distinct representations for entities in a KG that allow to infer a reflexive pattern.
When the imaginary part of the embeddings is zero, the above mentioned proof holds for $\mathbf{e}_{ri} = \bar{\mathbf{e}}_i$. 
\end{proof}

\textbf{Learned Transformation Functions.}
Figure~\ref{fig:sameD} illustrates the results of learned transformation functions for various relations in WordNet. 
Sub-figure (a) and (b) refer to the \emph{hyponym} relation. However, the depicted shape of transformation function differs for hyperbolic and elliptic transformations.
This confirms the flexibility of the model in embedding various graph structures as well as diversity in density/sparsity of flow (e.g.,  \emph{hyponym} relation). 
We also observed that when two pairs of relations form inverse patterns (in the same dimension), the model mainly learns the same transformation functions but with different directions.


\end{document}